\documentclass[12pt]{article}

\usepackage[top=0.95in,bottom=0.95in,left=0.95in,right=0.95in]{geometry}
\usepackage[utf8]{inputenc} 
\usepackage[T1]{fontenc}   
\usepackage{lmodern}
\usepackage{hyperref}      
\usepackage{url}            
\usepackage{booktabs}       
\usepackage{amsfonts}       
\usepackage{nicefrac}       
\usepackage{microtype}  
\usepackage{natbib}
\usepackage{caption}

\usepackage{amsfonts}       
\usepackage{nicefrac}       
\usepackage{microtype}      
\usepackage{xcolor}         
\usepackage{mathtools} 
\usepackage{tikz}
\usepackage{graphicx}
\usepackage{subfigure}
\usepackage{diagbox}

\usepackage{amsmath}
\usepackage{dsfont}
\usepackage{amssymb,amsthm,commath}
\usepackage{bm,xfrac,xcolor}
\usepackage{nccmath}
\usepackage[toc,page,header]{appendix}

\hypersetup{colorlinks,urlcolor=black, citecolor=blue}

\usepackage{algorithm}
\usepackage[noend]{algorithmic}
\usepackage{bbm}
\usepackage{enumitem}
\usepackage{gensymb}
\usepackage{refcount}
\usepackage{mathtools}
\usepackage{stackengine}
\usepackage{multirow}
\usepackage{makecell}

\usepackage{comment}
\usepackage{mathrsfs}


\newcommand{\INDSTATE}[1][1]{\STATE\hspace{-0.75em}{\algorithmicindent}}

\def\myfnt{\ifx\protect\@typeset@protect\expandafter\footnote\else\expandafter\@gobble\fi}
\makeatother

\makeatletter
\def\BState{\State\hskip-\ALG@thistlm}
\makeatother





\usepackage{placeins,afterpage}

\makeatletter
\newcommand*\bigcdot{\mathpalette\bigcdot@{.5}}
\newcommand*\bigcdot@[2]{\mathbin{\vcenter{\hbox{\scalebox{#2}{$\m@th#1\bullet$}}}}}
\makeatother

\usepackage{tikz}
\tikzset{
  c/.style={every coordinate/.try}
}

\newcommand\numberthis{\addtocounter{equation}{1}\tag{\theequation}}

\mathchardef\mhyphen="2D

\newcommand{\offdiag}{\mathit{off\mhyphen diag}}

\newcommand{\diag}{\mathit{diag}}

\newcommand{\tupr}{\rmbf^{1:m}}
\newcommand{\tuprhat}{\hat{\rmbf}^{1:m}}

\newcommand{\fair}{\textrm{\textup{fair}}}
\newcommand{\quadr}{\textrm{\textup{quad}}}
\newcommand{\qmean}{\textrm{\textup{qmean}}}
\newcommand{\cov}{\textrm{\textup{cov}}}
\newcommand{\eo}{\textrm{\textup{EO}}}

\newcommand{\Hcal}{{\cal H}}

\newcommand{\Lcal}{{\cal L}}
\newcommand{\Mcal}{{\cal M}}

\newcommand{\Rcal}{{\cal R}}
\newcommand{\Scal}{{\cal S}}

\newcommand{\Xcal}{{\cal X}}

\newcommand{\Bmbb}{\mathbb{B}}

\newcommand{\Pmbb}{\mathbb{P}}

\newcommand{\Rmbb}{\mathbb{R}}

\newcommand{\Zmbb}{\mathbb{Z}}

\newcommand{\thetambf}{\bm{\theta}}

\newcommand{\smbfbar}{\oline{\mathbf{s}}}

\newcommand{\rmbfbar}{\oline{\mathbf{r}}}

\newcommand{\rmbfhat}{\hat{\mathbf{r}}}
\newcommand{\ambfhat}{\hat{\mathbf{a}}}
\newcommand{\fmbfhat}{\hat{\mathbf{f}}}

\newcommand{\Bmbfhat}{\hat{\mathbf{B}}}
\newcommand{\lambdahat}{\hat{\lambda}}

\newcommand{\fmbfbar}{\oline{\mathbf{f}}}

\newcommand{\Bmbfbar}{\oline{\mathbf{B}}}

\newcommand{\Ambf}{\mathbf{A}}
\newcommand{\ambf}{\mathbf{a}}
\newcommand{\Bmbf}{\mathbf{B}}
\newcommand{\bmbf}{\mathbf{b}}
\newcommand{\Cmbf}{\mathbf{C}}
\newcommand{\cmbf}{\mathbf{c}}

\newcommand{\dmbf}{\mathbf{d}}

\newcommand{\embf}{\mathbf{e}}

\newcommand{\fmbf}{\mathbf{f}}

\newcommand{\Imbf}{\mathbf{I}}

\newcommand{\ombf}{\mathbf{o}}

\newcommand{\Qmbf}{\mathbf{Q}}

\newcommand{\Rmbf}{\mathbf{R}}
\newcommand{\rmbf}{\mathbf{r}}

\newcommand{\smbf}{\mathbf{s}}

\newcommand{\Wmbf}{\mathbf{W}}

\newcommand{\xmbf}{\mathbf{x}}

\newcommand{\ymbf}{\mathbf{y}}

\newcommand{\zmbf}{\mathbf{z}}

\newcommand{\oline}[1]{\mkern 1.5mu\overline{\mkern-1.5mu#1}}

\renewcommand{\hbar}{\oline{h}}

\newcommand{\Sbar}{\oline{S}}
\newcommand{\sbar}{\oline{s}}

\newcommand{\ahat}{\hat{a}}

\newcommand{\ophi}{\overline{\phi}}

\newcommand\inner[2]{\langle #1, #2 \rangle}

\newcommand{\1}{\mathbbm{1}}

\newcommand{\hphi}{\hat{\phi}}

\newcommand{\bphi}{\oline{\phi}}

\newcommand{\alphambf}{\bm{\alpha}}
\newcommand{\taumbf}{\bm{\tau}}

\newcommand{\gammambf}{\bm{\gamma}}

\DeclareMathOperator*{\argmax}{argmax}

\renewcommand{\P}{\mathbb{P}}

\newcommand{\baligned}     {\begin{aligned}}
	\newcommand{\ealigned}     {\end{aligned}}
\newcommand{\barray}       {\begin{array}}
	\newcommand{\earray}       {\end{array}}
\newcommand{\bbmatrix}     {\begin{bmatrix}}
	\newcommand{\ebmatrix}     {\end{bmatrix}}
\newcommand{\bcases}       {\begin{cases}}
	\newcommand{\ecases}       {\end{cases}}
\newcommand{\bcenter}      {\begin{center}}
	\newcommand{\ecenter}      {\end{center}}
\newcommand{\bcolumn}      {\begin{column}}
	\newcommand{\ecolumn}      {\end{column}}
\newcommand{\bcolumns}     {\begin{columns}}
	\newcommand{\ecolumns}     {\end{columns}}
\newcommand{\benumerate}   {\begin{enumerate}}
	\newcommand{\eenumerate}   {\end{enumerate}}
\newcommand{\bequation}    {\begin{equation}}
	\newcommand{\eequation}    {\end{equation}}
\newcommand{\bequationn}   {\begin{equation*}}
	\newcommand{\eequationn}   {\end{equation*}}
\newcommand{\bfigure}      {\begin{figure}}
	\newcommand{\efigure}      {\end{figure}}
\newcommand{\bflushright}  {\begin{flushright}}
	\newcommand{\eflushright}  {\end{flushright}}
\newcommand{\bitemize}     {\begin{itemize}}
	\newcommand{\eitemize}     {\end{itemize}}
\newcommand{\bpmatrix}     {\begin{pmatrix}}
	\newcommand{\epmatrix}     {\end{pmatrix}}
\newcommand{\bsubequations}{\begin{subequations}}
	\newcommand{\esubequations}{\end{subequations}}
\newcommand{\btable}       {\begin{table}}
	\newcommand{\etable}       {\end{table}}
\newcommand{\btabular}     {\begin{tabular}}
	\newcommand{\etabular}     {\end{tabular}}
\newcommand{\bvmatrix}     {\begin{vmatrix}}
	\newcommand{\evmatrix}     {\end{vmatrix}}

\newcommand{\bequali}      {\bsubequations\begin{align}}
	\newcommand{\eequali}      {\end{align}\esubequations}

\newtheorem{assumption}{Assumption}
\newtheorem{prop}{Proposition}
\newtheorem{example}{Example}
\newtheorem{remark}{Remark}
\newtheorem{theorem}{Theorem}
\newtheorem{corollary}{Corollary}
\newtheorem{definition}{Definition}
\newtheorem{lemma}{Lemma}

\newcommand{\balgorithm}  {\begin{algorithm}}
	\newcommand{\ealgorithm}  {\end{algorithm}}
\newcommand{\balgorithmic}{\begin{algorithmic}}
	\newcommand{\ealgorithmic}{\end{algorithmic}}
\newcommand{\bassumption} {\begin{assumption}}
	\newcommand{\eassumption} {\end{assumption}}
\newcommand{\bcorollary}  {\begin{corollary}}
	\newcommand{\ecorollary}  {\end{corollary}}
\newcommand{\bdefinition} {\begin{definition}}
	\newcommand{\edefinition} {\end{definition}}
\newcommand{\bexample}    {\begin{example}}
	\newcommand{\eexample}    {\end{example}}
\newcommand{\bproposition}    {\begin{prop}}
	\newcommand{\eproposition}    {\end{prop}}
\newcommand{\blemma}      {\begin{lemma}}
	\newcommand{\elemma}      {\end{lemma}}
\newcommand{\bproblem}    {\begin{problem}}
	\newcommand{\eproblem}    {\end{problem}}
\newcommand{\bproof}      {\begin{proof}}
	\newcommand{\eproof}      {\end{proof}}
\newcommand{\bremark}     {\begin{remark}}
	\newcommand{\eremark}     {\end{remark}}
\newcommand{\btheorem}    {\begin{theorem}}
	\newcommand{\etheorem}    {\end{theorem}}
	
\newcommand{\prodRcal}{\Rcal^{1:m}}

\usepackage{chngcntr}
\counterwithout{equation}{section}
\counterwithout{table}{section}
\counterwithout{figure}{section}
\counterwithout{algorithm}{section}
\usepackage[customcolors]{hf-tikz}
\usepackage{adjustbox}
\usepackage[normalem]{ulem}
\usepackage{cancel}
\usetikzlibrary{calc,shapes.geometric}
\renewcommand\cite{\citep}

\usepackage{colortbl}
\newcommand{\highlight}[1]{{\cellcolor{gray!25}#1}}

\begin{document}

\title{Quadratic Metric Elicitation for Fairness and Beyond}

\author{Gaurush Hiranandani \\
UIUC\\
gaurush2@illinois.edu
\and
Jatin Mathur \\
UIUC\\
jatinm2@illinois.edu
\and
Harikrishna Narasimhan \\
Google Research\\
hnarasimhan@google.com
\and
Oluwasanmi Koyejo\\
UIUC \& Google Research Accra\\
sanmi@illinois.edu
}

\date{\today}

\flushbottom
\maketitle

\begin{abstract}
    Metric elicitation is a recent framework for eliciting classification performance metrics that best reflect implicit user preferences based on the task and context. However, available elicitation strategies have been limited to linear (or quasi-linear) functions of predictive rates, which can be practically restrictive for many applications including fairness. This paper develops a strategy for eliciting more flexible multiclass metrics defined by quadratic functions of rates, designed to reflect human preferences better. We show its application in eliciting quadratic violation-based group-fair metrics. Our strategy requires only relative preference feedback, is robust to noise, and achieves near-optimal query complexity. We further extend this strategy to eliciting polynomial metrics -- thus broadening the use cases for metric elicitation.
\end{abstract}

\section{Introduction}
\label{sec:intro}
\emph{Given a classification task, which performance metric should the classifier optimize?} This question is often faced by practitioners while developing machine learning solutions. For example, consider cancer diagnosis where a doctor applies a cost-sensitive predictive model to classify patients into cancer categories~\citep{yang2014multiclass}. The costs may be based on known consequences of misdiagnosis, i.e, side-effects of treating a healthy patient vs. mortality rate for not treating a sick patient. Although it is clear that the chosen costs directly determine the model decisions and dictate patient outcomes, it is not clear how to quantify the expert's intuition into precise quantitative cost trade-offs, i.e., the performance metric. 

Indeed, the above is also true for a variety of other domains including \emph{fair machine learning} where picking the right metric 
is a critical challenge~\citep{Dmitriev2016MeasuringM, zhang2020joint}. The issue is exacerbated when the practitioner’s notion of fairness does not exactly match with any standard fairness criterion. For example, a practitioner may be interested in weighting each group discrepancy differently, but may not be able to provide us with the exact weights or a precise mathematical expression that reflects on the practitioner’s innate fairness notion.

\begin{figure}[t]
    \centering
    \vspace{-5pt}
    \includegraphics[scale = 0.4]{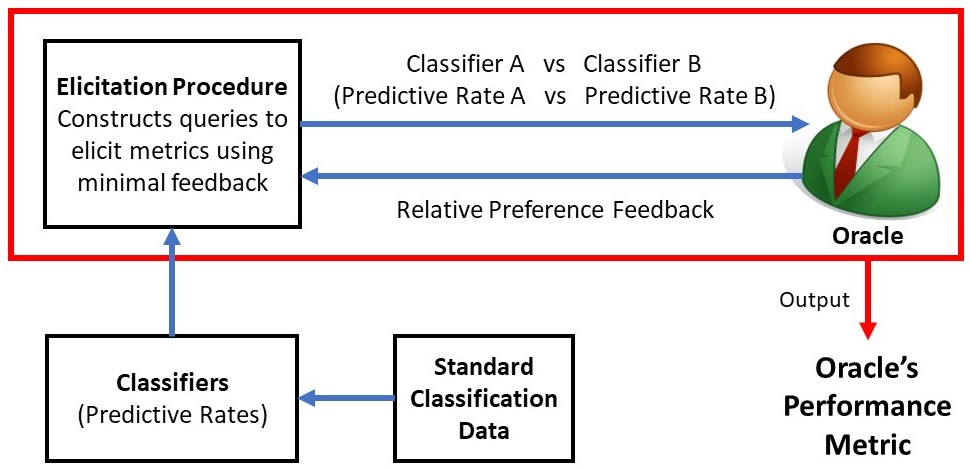}
    \vspace{-0.1cm}
    \caption{Metric Elicitation~\citep{hiranandani2018eliciting}.}
    \label{fig:meframework}
    \vskip -0.6cm
\end{figure}

\cite{hiranandani2018eliciting, hiranandani2019multiclass, hiranandani2020fair} addressed this issue by formalizing the framework of \emph{Metric Elicitation (ME)}, whose goal is to estimate a performance metric using preference feedback from a user. The motivation is that by employing metrics that reflect a user's innate trade-offs given the task, context, and population at hand, one can learn models that best capture the user preferences~\citep{hiranandani2018eliciting}. As humans are often inaccurate in providing absolute quality feedback~\citep{qian2013active}, \cite{hiranandani2018eliciting} propose to use pairwise comparison queries, where the user (oracle) is asked to compare two classifiers and provide a relative preference. Using such pairwise comparison queries, ME aims to recover the oracle's metric. Figure~\ref{fig:meframework} (reproduced from \cite{hiranandani2018eliciting})  depicts the ME framework.

A notable drawback of existing ME strategies is that they only handle linear or quasi-linear function of predictive rates, which can be restrictive for many applications where the metrics are non-linear. For example, in \emph{fair machine learning}, classifiers are often judged by measuring discrepancies between predictive rates for different protected groups~\citep{hardt2016equality}. Similarly, discrepancies among different distributions are measured in \emph{distribution matching} applications~\citep{narasimhan2018learning, Fab1}. A common measure of discrepancy in such applications is the squared difference, which is a quadratic metric that cannot be handled by existing approaches. Quadratic metrics also find use in class-imbalanced learning~\citep{goh2016satisfying, narasimhan2018learning} (see Section~\ref{ssec:metric} for examples). Motivated by these examples, in this paper, we propose strategies for eliciting metrics defined by \emph{quadratic} functions of rates, that encompass linear metrics as special cases. Our approach also generalizes to eliciting polynomial metrics, a universal family of functions~\citep{stone1948generalized},  
allowing one to better capture real-world human preferences. 

Our high-level idea is to
approximate the quadratic metric
with multiple linear functions, employ linear ME to estimate the individual 
 local slopes, 
 and combine the slope estimates to reconstruct the original metric. 
 While natural and elegant, this approach comes with non-trivial challenges. 
 Firstly, we must choose 
 center 
 points for the local-linear approximations,
 and the chosen points must represent
 feasible queries. Secondly, because of the use of pairwise queries, we only receive \emph{slopes} (directions) and not magnitudes for the local-linear functions, requiring intricate analyses to reconstruct the original 
 metric and to deal with multiplicative errors that result.
 Despite the challenges,
 our method requires a query complexity that is only \emph{linear} in the number of unknowns, which we show is {\em near-optimal}. To our knowledge, we are the first to prove such a lower bound for metric elicitation.

We further elaborate on eliciting group-fair metrics. The prior work by \citet{hiranandani2020fair} consider a restricted class of fairness metrics, where the fairness discrepancies are defined to be the \emph{absolute} differences between group-specific rates. Moreover, their approach does not  generalize 
to other families of metrics. In contrast, we are able to handle a more general family of non-linear fairness metrics defined by quadratic functions of group rate differences and 
show how our proposed quadratic ME approach is easily adaptable to elicit such group-fair quadratic metrics. 

In summary, we make the following contributions :
\bitemize[leftmargin=8pt, itemsep=0pt]
\item We propose a novel quadratic ME algorithm for classification problems, which requires only pairwise preference feedback 
either over classifiers or predictive rates. 

\item Specific to group-based fairness tasks, we show how to jointly elicit the predictive performance and fairness metrics, and the trade-off between them.
\item We show that the proposed approach is robust under feedback and 
finite sample noise and requires a  near-optimal number of queries. 
\item We empirically validate the proposal 
for multiple classes and groups on simulated oracles.
\item 
We discuss how our strategy can be generalized to elicit higher-order polynomials by recursively applying the procedure to elicit lower-order approximations. 
\eitemize

\textbf{Paper Organization:} For ease of exposition, we first discuss quadratic metric elicitation in the usual multiclass classification setup without fairness. Section~\ref{sec:background} contains the problem setup and the associated background, and Section~\ref{sec:quadme} describes the proposed quadratic ME procedure. 
We then cover ME under the multiclass-multigroup framework in Section ~\ref{sec:fairme}, where we additionally have protected group information embedded in the problem setup. 
In Section ~\ref{sec:guarantees}, we provide guarantees for our proposed procedures, and in Section~\ref{sec:experiments}, we present our experiments. We discuss related work in Section~\ref{sec:relatedwork} and provide concluding remarks in Section~\ref{sec:discussion}.

\textbf{Notations.} 
For $k \in \Zmbb_+$, we denote $[k] = \{1, \cdots , k\}$ and use $\Delta_k$ to denote the $(k-1)$-dimensional simplex. 
 We denote  inner products 
 by $\inner{\cdot}{\cdot}$ and  Hadamard products by $\odot$. 
$\|\cdot\|_F$ represents the Frobenius norm, 
and $\alphambf_i \in \Rmbb^q$ denotes the $i$-th standard basis vector, where the $i$-th coordinate is 1 and  others are 0. 

\section{Background}
\label{sec:background}

We consider a $k$-class classification setting with $X \in \Xcal$ and $Y \in [k]$ denoting the input and output random variables, respectively. We assume access to an $n$-sized sample $\{(\xmbf, y)_i\}_{i=1}^n$ generated \emph{iid} from a distribution $ \Pmbb(X, Y)$. 
We work with randomized classifiers $h : \Xcal \rightarrow \Delta_k$ that for any $\xmbf$ gives a distribution $h(\xmbf)$ over the $k$ classes and use 
 $\Hcal = \{h : \Xcal \rightarrow \Delta_k\}$
 to denote the set of all classifiers. 

\emph{Predictive rates:} 
We denote the predictive rates for a classifier $h$ by the vector $\rmbf(h, \Pmbb) \in \Rmbb^{k}$, where the $i$-th coordinate is the fraction of label-$i$ examples for which the randomized classifier $h$ also predicts class $i$:
\begin{align}
	r_{i}(h, \Pmbb) \coloneqq \Pmbb(h(X) = i | Y = i)  \quad \text{for} \; i \in [k].
	\label{eq:components}
\end{align}
The probability above is over draw of $(X, Y) \sim \P$ and the randomness in $h$. The proposed setup and solution (discussed later) easily extends to general predictive rates of the form $\Pmbb(h(X) = j | Y = i)$ for $i, j \in [k]$. For simplicity, we defer this extension to Appendix  \ref{append:generalquad}.  

\emph{Metrics:} We consider metrics  that are defined by a general function $\phi : [0, 1]^{k}  \rightarrow \Rmbb$ of rates: 
  $$\phi(\rmbf(h, \Pmbb)).$$
This includes the (weighted) accuracy
$\phi^{\text{acc}}(\rmbf(h, \Pmbb))$ $\,=\, \sum_{i} a_i r_i(h, \Pmbb)$, for weights $a_i \in \mathbb{R}_{+}$, the G-mean, and many more metrics~\citep{sokolova2009systematic}. 
Unless specified, we treat metrics as utilities, i.e., larger values are better. 
Since the metric's scale does not affect the learning problem~\citep{narasimhan2015consistent}, we allow $\phi : [0, 1]^{k}  \rightarrow [0,1]$.

\emph{Feasible rates:} We will restrict our attention to only those rates that are feasible, i.e., can be achieved by some classifier. The set of all feasible rates is given by: 
$$\Rcal = \{\rmbf(h, \Pmbb) \,:\, h \in \Hcal \}.$$ 
To avoid clutter in notations, we will suppress the dependence on $\Pmbb$ and $h$ if it is clear from the context.
\subsection{Metric Elicitation: Problem Setup}
\label{ssec:me}
We now describe the problem of \emph{Metric Elicitation}, 
which follows from~\cite{hiranandani2019multiclass}. There's an \textit{unknown} metric $\phi$, and we seek to elicit its form by posing queries to an \emph{oracle} asking  which of two classifiers is more preferred by it. The oracle has access to the  metric $\phi$ and responds by comparing its value on the two classifiers.
\bdefinition
[Oracle Query] Given two classifiers $h_1, h_2$ (equiv. to rates $\rmbf_1, \rmbf_2$ respectively), a query to the Oracle (with metric $\phi$) is represented by:
\begin{align}
\Gamma(h_1, h_2\,;\, \phi) = \Omega(\rmbf_1, \rmbf_2\,;\,\phi) &= \1[\phi(\rmbf_1) > \phi(\rmbf_2)], 
\end{align}
\noindent where $\Gamma: \Hcal \times \Hcal \rightarrow \{0,1\}$ and $\Omega: \Rcal \times \Rcal \rightarrow \{0, 1\}$. The query asks whether $h_1$ is preferred to $h_2$ (equiv. if $\rmbf_1$ is preferred to $\rmbf_2$), as measured by $\phi$. 
\label{def:query}
\edefinition
\vskip -0.1cm

In practice, the oracle can be an expert, a group of experts, or an entire user population. The ME framework can be applied by posing classifier comparisons directly via interpretable learning techniques~\citep{ ribeiro2016should} 
or via A/B testing~\citep{tamburrelli2014towards}. For example, in an internet-based application 
one may perform the A/B test by deploying two classifiers A and B with two different sub-populations of users and use their level of engagement to decide the preference over the two classifiers. 
For other applications, one may present 
visualizations of rates of the two classifiers 
(e.g.,  \citep{shen2020designing}), 
and have the user provide the  preference (see Appendix~\ref{append:userstudy} for an example). 
Moreover, since the metrics we consider are functions of only the predictive rates, queries comparing classifiers are the same as queries on the associated rates. So for convenience, we will have our algorithms pose queries comparing two (feasible) rates.
Indeed  given a feasible rate, one can efficiently find the associated classifier (see Appendix \ref{append:ssec:sphere} for details). 
We next formally state the ME problem.
\bdefinition [Metric Elicitation with Pairwise Queries (given $\{(\xmbf,y)_i\}_{i=1}^n$)~\citep{hiranandani2018eliciting, hiranandani2019multiclass}] Suppose that the oracle's (unknown) performance metric is $\phi$.  Using oracle queries of the form $\Omega(\rmbfhat_1, \rmbfhat_2\,;\,\phi)$, where $\rmbfhat_1, \rmbfhat_2$ are the estimated rates from samples, recover a metric $\hphi$ such that $\Vert\phi - \hphi\Vert < \kappa$ under a  suitable norm $\Vert \cdot \Vert$ for sufficiently small error tolerance $\kappa > 0$.
\label{def:me}
\edefinition
\vskip -0.2cm

The performance of ME is evaluated both by the query complexity and the quality of the elicited metric~\citep{hiranandani2018eliciting, hiranandani2019multiclass}. As is standard in the decision theory literature~\citep{koyejo2015consistent},
we present our ME approach by first assuming access to population quantities such as the population rates $\rmbf(h, \Pmbb)$, then examine estimation error from finite samples, i.e., with empirical rates $\rmbfhat(h, \{(\xmbf,y)_i\}_{i=1}^n)$. 

\subsection{Linear Metric Elicitation}
\label{ssec:mpme}

As a warm up, we overview the Linear Performance Metric Elicitation (LPME) procedure of \citep{hiranandani2019multiclass}, which we will use as a subroutine.  
Here we assume that the oracle's metric is a linear function of rates $\phi^{\text{lin}}(\rmbf) \coloneqq \inner{\ambf}{\rmbf}$, for some unknown weights $\ambf \in \Rmbb^k$. 
In other words, given two rates $\rmbf_1$ and $\rmbf_2$, the oracle returns $\1[\inner{\ambf}{\rmbf_1} > \inner{\ambf}{\rmbf_2}]$. Since the metrics are scale invariant~\citep{narasimhan2015consistent}, without loss of generality (w.l.o.g.), one may assume $\Vert \ambf \Vert_2=1$. The goal is to elicit (the slope of) $\ambf$ using pairwise comparisons over rates.

When the number of classes $k = 2$, the coefficients $\ambf$ can be elicited using a one-dimensional binary search. When $k > 2$, one can apply a  coordinate-wise procedure, performing a binary search in one coordinate, while keeping the others fixed. The efficacy of this procedure, however, hinges on the geometry of the 
set of rates $\Rcal$. Before discussing the geometry, we make a mild assumption 
that ensures some signal for non-trivial classification.
\bassumption
\label{assump:distribution}
The conditional-class distributions are distinct, i.e., $\Pmbb(Y=i|X) \ne \Pmbb(Y=j|X)$
$\forall \; i \ne j$.
\label{as:sphere}
\eassumption

Let $\embf_i \in \{0,1\}^k$ denote the rates achieved by a trivial classifier that predicts class $i$ for all inputs. 

\begin{figure}[t]
\hspace{3cm}
\begin{tikzpicture}[scale = 1.25]
    \begin{scope}[scale = 0.6]\scriptsize
    
    \def\r{0.12};
    
    \coordinate (a) at (-0.4,1);
    \coordinate (b) at (0.6, 3.25);
    \coordinate (c) at (7, 4);
    \coordinate (d) at (6.5, 2);
    \coordinate (e) at (2.5, -0.75);
    \coordinate (f) at (-0.1, -0.5);
    
    \coordinate (labelleft) at (3, -1.25);
    
    \coordinate (Cent) at (3,1.75);
    \coordinate (Centcent) at (2.85,1.90);
    \coordinate (Cent1) at (4.45,1.75);
    \coordinate (Cent2) at (3,3.2);
    \coordinate (CentL) at (1.55,1.75);
    
    \coordinate (Space1) at (0.2,0.2);
    \coordinate (SpaceR) at (0.25,3.25);
    \coordinate (Spacem) at (-0.1,2.5);
    
    \coordinate (Sphere) at (5,3.25);
    \coordinate (Sphere0) at (3,1);
    \coordinate (Sphere1) at (4.45,1);
    \coordinate (Sphere2) at (3,2.45);
    \coordinate (SphereL) at (1.65,1);
    \coordinate (Sphereminus) at (3.4,0.95);
    
    \coordinate (r) at (3.75,2);
    
    \coordinate (u11) at (-0.25, -0.25);
    \coordinate (uextra121) at (1, 3.8);
    \coordinate (u21) at (4, 4.5);
    \coordinate (uextra2k1) at (6, 1);
    \coordinate (uk1) at (3.5, -0.5);
    
    \coordinate (u12) at (0.3, -0.70);
    \coordinate (uextra122) at (0.65, 2.65);
    \coordinate (u22) at (3.4, 4.2);
    \coordinate (uextra2k2) at (5.15, 1.6);
    \coordinate (uk2) at (3.75, 0.1);
    
    \coordinate (u13) at (-0.30, 0.10);
    \coordinate (uextra123) at (1.25, 3.25);
    \coordinate (u23) at (4, 4.5);
    \coordinate (uextra2k3) at (5.75, 1);
    \coordinate (uk3) at (3.75, -0.5);
    
    \coordinate (u14) at (-0.30, 0.10);
    \coordinate (uextra124) at (1.25, 3.25);
    \coordinate (u24) at (4, 4.5);
    \coordinate (uextra2k4) at (5.75, 1);
    \coordinate (uk4) at (3.75, -0.5);

    \fill[color=black] 
            (Cent) circle (0.08)
            (u11) circle (0.08)
            (u21) circle (0.08)
            (uk1) circle (0.08);
    
    \draw[thick] (u11) .. controls (a) and (b) .. (uextra121) 
    -- (u21) .. controls (c) and (d) .. (uextra2k1) -- (uk1) .. controls  (e) and (f) .. (u11);
    
    
    
    \draw[thick] (Cent) circle (2cm);

    \draw[thick, dotted] (Cent) circle (0.5cm);
    \draw[thick, dotted] (Cent1) circle (0.5cm);
    \draw[thick, dotted] (Cent2) circle (0.5cm);
    \draw[thick, dotted] (CentL) circle (0.5cm);
    \node at (SpaceR) {{$\Rcal$}};
    
    \node at (Sphere) {\large{${\Scal}$}};
    \node at (Sphere0) {\tiny{$\Scal_{\ombf}$}};
    \node at (Sphere1) {\tiny{$\Scal_{\zmbf_1}$}};
    \node at (Sphere2) {\tiny{$\Scal_{\zmbf_2}$}};
    \node at (SphereL) {\tiny{$\Scal_{-\zmbf_1}$}};
    
    \node[below right] at (Centcent) {$\ombf$};
    
    \node[below] at (u11) {{$\embf_1$}};
    \node[above] at (u21) {{$\embf_2$}};
    \node[below right] at (uk1) {{$\embf_k$}};
    
    \node at (labelleft) {{\normalsize{(a)}}};
    
     \end{scope}
     
     \begin{scope}[shift={(5,0)},scale = 0.5]\scriptsize
    
    \def\r{0.12};
    
    \coordinate (a) at (-0.2,1);
    \coordinate (b) at (0.8, 2.75);
    \coordinate (c) at (7, 4);
    \coordinate (d) at (6.5, 2);
    \coordinate (e) at (2.5, -0.75);
    \coordinate (f) at (-0.1, -0.5);
    
    \coordinate (labelright) at (3, -1.5);
    
    \coordinate (Cent) at (3,1.75);
    \coordinate (Centcent) at (2.85,1.90);
    \coordinate (CentR) at (3.6,2.35);
    \coordinate (CentL) at (2.4,1.15);
    
    \coordinate (Space1) at (0.2,0.2);
    \coordinate (Space2) at (-0.3,1.3);
    \coordinate (Spacem) at (-0.1,2.5);
    
    \coordinate (Sphere) at (4.5,3);
    \coordinate (Sphereplus) at (2.6,2.55);
    \coordinate (Sphereminus) at (3.4,0.95);
    
    \coordinate (r) at (3.75,2);
    
    \coordinate (u11) at (-0.25, -0.25);
    \coordinate (uextra121) at (1.25, 3.75);
    \coordinate (u21) at (4, 4.5);
    \coordinate (uextra2k1) at (6, 1);
    \coordinate (uk1) at (3.5, -0.5);
    
    \coordinate (u12) at (0.3, -0.70);
    \coordinate (uextra122) at (0.65, 2.65);
    \coordinate (u22) at (3.4, 4.2);
    \coordinate (uextra2k2) at (5.15, 1.6);
    \coordinate (uk2) at (3.75, 0.1);
    
    \coordinate (u13) at (-0.30, 0.10);
    \coordinate (uextra123) at (1.25, 3.25);
    \coordinate (u23) at (4, 4.5);
    \coordinate (uextra2k3) at (5.75, 1);
    \coordinate (uk3) at (3.75, -0.5);
    
    \coordinate (u14) at (-0.30, 0.10);
    \coordinate (uextra124) at (1.25, 3.25);
    \coordinate (u24) at (4, 4.5);
    \coordinate (uextra2k4) at (5.75, 1);
    \coordinate (uk4) at (3.75, -0.5);

    \fill[color=black] 
            (Cent) circle (0.08)
            (u11) circle (0.08)
            (u21) circle (0.08)
            (uk1) circle (0.08);
    
    \draw[dashed, blue, thick] (u11) .. controls (a) and (b) .. (uextra121) 
    -- (u21) .. controls (c) and (d) .. (uextra2k1) -- (uk1) .. controls  (e) and (f) .. (u11);
    
    \draw[dashed, brown, thick] (u11) .. controls (-2.25, 1.25) and (0.9, 3) .. (u21) .. controls (8.5,4.75) and (7,2) .. (6.5, 1.25) .. controls (6.5, 1) and (5.25, 0.3) .. (uk1) -- (u11);
    
    \draw[dashed, red, thick] (u11) .. controls (-1.5, 1.5) and (-0.5, 3.5) .. (1.5, 4.5) 
    -- (u21) .. controls (6.5, 3.5) and (6, 2) .. (6, 1.75) -- (uk1) .. controls  (3, -1) and (-0.25, -0.75) .. (u11);
    
    \draw[thick] (Cent) circle (1.5cm);
    
    \node at (Space1) {{$\Rcal^1$}};
    \node at (Space2) {{$\Rcal^2$}};
    \node at (Spacem) {{$\Rcal^m$}};
    
    \node at (Sphere) {\large{$\overline{\Scal}$}};
    
    \node[below right] at (Centcent) {$\ombf$};
    
    \node[below left] at (u11) {{$\embf_1$}};
    \node[above] at (u21) {{$\embf_2$}};
    \node[below right] at (uk1) {{$\embf_k$}};
    
    \node at (labelright) {{\normalsize{(b)}}};
    
     \end{scope}
     
    \end{tikzpicture}
    \vskip -0.2cm
      
     \caption{(a) Geometry of the set of predictive rates $\Rcal$: A convex set enclosing a sphere ${\Scal}$ with trivial rates $\embf_i \, \forall \, i \in [k]$ as vertices; (b) Geometry of the product set of group rates $\Rcal^1 \times \dots \times \Rcal^m$ (best seen in color)~\citep{hiranandani2020fair}; $\Rcal^u \, \forall \, u \in [m]$ are convex sets with common vertices $\embf_i \, \forall \, i \in [k]$ and enclose a sphere $\overline{\Scal} \subset \Rcal^1 \cap \dots \cap \Rcal^m$.}
      \label{fig:geometry}
\end{figure}

\bproposition
[Geometry of $\Rcal$; Figure~\ref{fig:geometry}(a)] The set of  rates $\Rcal \subseteq [0, 1]^{k}$ is convex, has vertices $\{\embf_i\}_{i=1}^k$, and  
contains the rate profile $\ombf = \tfrac{1}{k} \tiny{\sum_{i=1}^k \embf_i}$ in the interior. Moreover, $\ombf$ is achieved by the uniform random classifier which for any input predicts each class with equal probability.
\label{prop:C}
\eproposition
\bremark[Existence of sphere ${\Scal}$]
Since $\Rcal$ is convex and contains
the point $\ombf$ in the interior, there exists a 
sphere ${\Scal} \subset \Rcal$ of non-zero radius $\rho$ centered at $\ombf$.  
\label{rem:sphere}
\eremark

By restricting the coordinate-wise binary search procedure to posing queries from within a sphere, LPME can be seen as minimizing a strongly-convex function and shown to converge to a solution $\ambfhat$ close to $\ambf$. 
Specifically, the LPME procedure 
takes any 
sphere $\Scal \subset \Rcal$, binary-search tolerance $\epsilon$, and the oracle $\Omega$ (with metric $\phi^{\text{lin}}$) 
 as input, and by posing 
$O(k\log(1/\epsilon))$
queries recovers coefficients $\ambfhat$ with 
$\Vert \ambf - \ambfhat \Vert_2 \leq O(\sqrt{k}\epsilon)$. 
The details of the algorithm are provided in Appendix~\ref{append:sec:slme} for completeness, but the following remark is the most important for our subsequent discussion.

\bremark[LPME Guarantee]
Given any $k$-dimensional 
sphere $\Scal \subset \Rcal$ and an oracle $\Omega$ with metric $\phi^{\textrm{\textup{lin}}}(\rmbf)\coloneqq\inner{\ambf}{\rmbf}$, 
the LPME algorithm (Algorithm~\ref{alg:slme}, Appendix~\ref{append:sec:slme}) provides an estimate $\ambfhat$ with $\Vert \ambfhat \Vert_2=1$ such that the estimated slope is close to the true slope, i.e.,  $\sfrac{{a}_i}{{a}_j} \approx \sfrac{\hat a_i}{\hat a_j} \; \forall \; i, j\in [k]$.
\label{rm:ratio}
\eremark

Note that the LPME procedure is closely tied to the scale invariance condition and thus only estimates the slope (direction) of the coefficient vector $\ambf$,
and not its magnitude. 
Despite this drawback, we will discuss how we can elicit quadratic metrics using LPME in Section~\ref{sec:quadme}. 
Also note the algorithm takes as input an \emph{arbitrary} sphere $\Scal \subset \Rcal$, and restricts its queries to rate vectors within the sphere. 
 In Appendix~\ref{append:ssec:sphere}, we discuss an efficient procedure~\citep{hiranandani2019multiclass} for identifying a sphere 
 of suitable radius.

\subsection{Quadratic Performance Metrics}
\label{ssec:metric}
Equipped with the LPME subroutine, our aim is to elicit metrics that are quadratic functions of rates.

\bdefinition[Quadratic Metric] For a vector $ \ambf \in \Rmbb^k$  
and a negative semi-definite 
matrix $\Bmbf \in NSD_k$ with $\Vert \ambf \Vert_2^2  + \Vert \Bmbf \Vert_F^2 = 1$ (w.l.o.g.\ due to scale invariance):
\vspace{-0.1cm}
\begin{equation}
    \phi^\quadr(\rmbf \,;\, \ambf, \Bmbf) = \inner{\ambf}{\rmbf} + \frac{1}{2} \rmbf^T \Bmbf \rmbf.
    \label{eq:quadmet}
\end{equation}
\vspace{-0.7cm}
\label{def:quadmet}
\edefinition
This family trivially includes the linear metrics
as well as many modern metrics outlined below: 

\bexample[Class-imbalanced learning]
\emph{In problems with imbalanced class proportions, it is common to use metrics that emphasize equal performance across all classes. One example is Q-mean 
\citep{menon2013statistical},
which is the quadratic mean of rates:
{\small
$\phi^{\qmean}(\rmbf) = 1 -  1/k\sum_{i=1}^k \left(1 - r_i \right)^2.$}
}
\eexample

\bexample[Distribution matching]
\emph{
In certain binary classification applications, one needs the proportion of predictions 
for each class (i.e., the coverage) to match a target distribution $\boldsymbol{\pi} \in \Delta_2$ 
\citep{goh2016satisfying,narasimhan2018learning}. 
A 
measure often used for this task is the squared difference between the per-class coverage and the target distribution: 
{\small$\phi^{\cov}(\rmbf) \,=\, 1 - \frac{1}{2}\sum_{i=1}^2 \left(\cov_i(\rmbf) - \pi_i\right)^2$}, where 
{\small$\cov_i(\rmbf) = r_i + 1 - r_{\neq i}$}. 
Similar metrics can be found in the quantification literature where the target is set to the class prior $\Pmbb(Y=i)$ \citep{Fab1, 
Kar16}. 
We capture more general quadratic distance measures for distributions, e.g.\ {\small$(\bf{\cov}(\rmbf) - \boldsymbol{\pi})^{T}\Qmbf (\bf{\cov}(\rmbf)-\boldsymbol{\pi})$} for $\Qmbf \in NSD_2$ \citep{Lindsay08}. 
}
\label{ex:distmatchbin}
\eexample
\vspace{-0.1cm}

Lastly, we need the following assumption on the metric.

\bassumption
\label{assump:smoothness}
The gradient of  $\phi$ at the trivial rate $\ombf$ is non-zero, i.e., $\nabla \phi^{\quadr}(\rmbf)|_{\rmbf=\ombf} = \ambf + \Bmbf\ombf \neq 0.$
\label{as:smooth}
\eassumption
\vspace{-0.1cm}

The non-zero gradient assumption is reasonable for a concave $\phi^{\text{quad}}$, where it merely implies that the optimal classifier for the metric is not the uniform random classifier. 

\section{Quadratic Metric Elicitation}
\label{sec:quadme}
We now 
present our procedure for Quadratic Performance Metric Elicitation (QPME). We assume that the oracle's unknown metric is quadratic  (Definition~\ref{def:quadmet}) and seek to estimate its parameters $(\ambf, \Bmbf)$ 
by posing  queries to the oracle. 
Unlike LPME, a simple binary search based procedure cannot be directly applied to elicit these parameters. Our approach instead approximates the quadratic metric by a linear function at a few select but \emph{feasible} rate vectors and invokes LPME to estimate the local-linear approximations' slopes. 
One of the key challenges is to pick a small number of \emph{feasible} rates for performing the local approximations and to reconstruct the original metric \emph{just} from the estimated local slopes. 

\vspace{-0.25cm}
\subsection{Local Linear Approximation}
\vskip -0.2cm
We will find it convenient to work with a shifted version of the quadratic metric, centered at the  point $\ombf$, the uniform random rate vector (see Proposition \ref{prop:C}): 
\vspace{-0.1cm}
\begin{align*}
\phi^\quadr(\rmbf;\, \ambf, \Bmbf) &=  
\inner{\dmbf}{\rmbf - \ombf} + \frac{1}{2}(\rmbf - \ombf)^T \Bmbf (\rmbf - \ombf) + c \\
&=\bphi(\rmbf;\, \dmbf, \Bmbf) + c\numberthis \label{eq:quadmetshift},
\end{align*}
\vskip -0.2cm
where $\dmbf= \ambf+\Bmbf\ombf$ and $c$ is a constant independent of $\rmbf$, and so the oracle can be equivalently seen as responding with the shifted metric $\bphi(\rmbf;\, \dmbf, \Bmbf)$.

Note that, due to the scale invariance condition in Definition~\ref{def:quadmet}, the largest singular value of $\Bmbf$ is bounded by 1. This is because $\Vert \Bmbf \Vert_2 \leq \Vert \Bmbf \Vert_F \leq 1$. Thus the metric $\phi^{\quadr}$ is $1$-smooth and implies that it is locally linear around a given rate.
To this end, let $z$ be a fixed point in $\Rcal$, then the metric 
can be closely approximated by its first-order Taylor expansion in a small neighborhood around $\zmbf$, for a constant $c'$ as follows:
\vspace{-0.2cm}
\begin{equation}
\bphi(\rmbf;\, \dmbf, \Bmbf) \approx \inner{\dmbf + \Bmbf (\zmbf - \ombf)}{\rmbf} + c'.
\label{eq:loclinapx}
\end{equation}
\vskip -0.2cm
So if we apply LPME to the metric $\bphi$ with the queries $(\rmbf_1, \rmbf_2)$ to the oracle restricted to a small ball around $\zmbf$, the procedure effectively estimates the  slope of the vector $\dmbf + \Bmbf (\zmbf - \ombf)$ in the above linear function (up to a small approximation error). 

We exploit this idea by applying LPME to small neighborhoods around selected  points to  elicit the coefficients $\ambf$ and $\Bmbf$ for the original metric in~\eqref{eq:quadmet}. For simplicity, we will assume that the oracle is noise-free and later show robustness to noise and the query complexity guarantees in Section~\ref{sec:guarantees}.

\vspace{-0.2cm}
\subsection{Eliciting Metric Coefficients}
\vskip-0.2cm
We outline the main steps of Algorithm~1 below. Please see Appendix~\ref{append:sec:qpme} for the full derivation.  

\textbf{Estimate coefficients $\dmbf$ (Line 1).}\
We first wish to estimate the linear portion $\dmbf$ of the metric $\bphi$ in~\eqref{eq:quadmetshift}. For this, we
apply the LPME subroutine to a small ball $\Scal_\ombf \subset \Scal$ of radius $\varrho < \rho$ around the point $\ombf$ (Fig.\ \ref{fig:geometry}(a) illustrates this). 
Within this ball, the metric $\bphi$ approximately equals the linear function
$\inner{\dmbf}{\rmbf} + c'$ (see \eqref{eq:loclinapx}), and so the LPME gives us an estimate of the slope of $\dmbf$.
 From Remark~\ref{rm:ratio},  
 the estimates $\fmbf_0 =
 (f_{10}, \dots, f_{k0})$  approximately satisfy the following $(k-1)$ equations: 
        \vspace{-0.2cm}
\begin{equation}
    \frac{d_i}{d_1} = \frac{f_{i0}}{f_{10}} \qquad \forall \; i \in \{2, \dots, k\}.
    \label{eq:0col}
\end{equation}

\textbf{Estimate coefficients $\Bmbf$ (Lines 2--4).}
Next, we wish to estimate each column of the matrix $\Bmbf$ of the metric $\bphi$ in~\eqref{eq:quadmetshift}. For this, we apply LPME to small neighborhoods around points in the direction of standard basis vectors $\alphambf_{j} \in \Rmbb^{k}$, $j = 1, \ldots, k$. 
Note that within a small ball around $\ombf + \alphambf_j$, the metric $\ophi$ is approximately  the linear function
$\inner{\dmbf + \Bmbf_{:,j}}{\rmbf} + c'$, and so the LPME procedure when applied to this region will give us an estimate of the slope of $\dmbf + \Bmbf_{:,j}$. However, to ensure that the center point we choose is a feasible rate, we will have to re-scale the standard basis, and apply the subroutine to balls $\Scal_{\zmbf_j}$ of radius $\varrho < \rho$ centered at $\zmbf_j = \ombf + (\rho - \varrho)\alphambf_j$. See Figure~\ref{fig:geometry}(a) for the visual intuition. The returned estimates $\fmbf_j = (f_{1j}, \dots, f_{kj})$ approximately satisfy:
        \vspace{-0.1cm}
\begin{equation}
\frac{d_i + (\rho-\varrho)B_{ij}}{d_1 + (\rho-\varrho)B_{1j}} = \frac{f_{ij}}{f_{1j}} \quad \forall \; i \in \{2, \ldots, k\},\; j \leq i.
\label{eq:jcol}
\end{equation}
Now note that since we are only eliciting slopes using LPME, we always lose out on one degree of freedom. However, the matrix $\Bmbf$ is symmetric, thus we have $k(k+1)/2 - 1$ equations. There are $k(k+1)/2 + k$ unknown entities in $\ambf$ and $\Bmbf$, and to estimate them we need $1$ more equation besides the normalization condition. 
For this, we apply LPME to a sphere $\Scal_{-\zmbf_1}$ of radius $\varrho$ around rate $-\zmbf_1$ as shown in Figure~\ref{fig:geometry}(a). The returned slopes $\fmbf_1^- = (f_{11}^-, \dots, f_{k1}^-)$ approximately satisfy:
\vspace{-0.1cm}
\begin{equation}
    \frac{d_2-(\rho - \varrho)B_{21}}{d_1-(\rho - \varrho)B_{11}} = \frac{f_{21}^-}{f_{11}^-}.
    \label{eq:negativegrad}
\end{equation}
\textbf{Putting it together (Line 5).}\ By combining~\eqref{eq:0col},~\eqref{eq:jcol} and~\eqref{eq:negativegrad}, and 
denoting $F_{i,j,l} = f_{il} / f_{jl}$ and $F^-_{i,j,l} = f^-_{il}/f^-_{jl}$,
we  express each entry of $\Bmbf$ in terms of $d_1$ as follows:
\begin{align*}
    B_{ij} &= \Big(F_{i,1,j} (1 + F_{j,1,1}) - F_{i,1,j} F_{j,1,0} d_{1} - F_{i,1,0}
    \\&\hspace{1cm}+ 
    F_{i,1,j}\textstyle\frac{F^-_{2,1,1} + F_{2,1,1} - 2F_{2,1,0}}{F^-_{2,1,1} - F_{2,1,1}}\Big)d_1.
    \numberthis \label{eq:poly2elicitamatfinal}
\end{align*}
Using $\dmbf= \ambf+\Bmbf\ombf$ and the fact that the coefficients are normalized, i.e., $\Vert \ambf \Vert_2^2  + \Vert \Bmbf \Vert_F^2 = 1$, we can obtain estimates for $\Bmbf$ and $\ambf$ independent of $d_1$. 
Note that the derivation so far assumes $d_1 \ne 0$. This is based on Assumption  \ref{assump:smoothness} that at least one coordinate of $\dmbf$ is non-zero, which w.l.o.g.\  we  take to be $d_1$.
In practice, we can identify a non-zero coordinate using $q$ trivial queries of the form $(\varrho\alphambf_i + \ombf, \ombf), \forall i \in [k]$.

\textbf{Technical novelty.}\ We emphasize that a key difference from~\cite{hiranandani2018eliciting, hiranandani2019multiclass}  is that they rely on a boundary point characterization which may not hold for general nonlinear metrics. 
Instead, we use structural properties of the metric to estimate local-linear approximations. While this may be a convenient approach (given LPME), as discussed in Section~\ref{sec:intro},  implementing it involves non-trivial challenges, such as: (a) working with \emph{only} slopes for the local-linear functions, (b) ensuring that the center points for the approximations are feasible, and (c) handling multiplicative errors in the analysis
(see Section \ref{sec:guarantees}). 

\begin{figure}[t]
\centering
\fbox{\parbox[t]{0.65\textwidth}{\small{\underline{\bf Algorithm~1: QPM Elicitation}\normalsize}    \\
\small
\textbf{Input:} 
${\Scal}$, 
Search tolerance $\epsilon > 0$, Oracle $\Omega$ with 
metric $\bphi$\\
1: \text{ \ }$\fmbf_0 \leftarrow$ LPME$\left(\Scal_\ombf, \epsilon, \Omega\right)$ with $\Scal_\ombf \subset {\Scal}$ and obtain~\eqref{eq:0col}\\
2: \text{ \ }\textbf{For} \, $j \in \{1,2,\dots,k\}$ \textbf{do}\\
3: \text{ \ \ \ } $\fmbf_j\leftarrow$LPME$\left(\Scal_{\zmbf_j}, \epsilon, \Omega\right)$ with $\Scal_{\zmbf_j} \subset {\Scal}$ and obtain~\eqref{eq:jcol}\\
4: \text{ \ }$\fmbf^-_{1} \leftarrow$ LPME$\left(\Scal_{-\zmbf_1}, \epsilon, \Omega\right)$ with $\Scal_{-\zmbf_1}\hspace{-2pt}\subset\hspace{-1pt} {\Scal}$ and obtain~\eqref{eq:negativegrad}\\
5: \text{ \ }$\ambfhat, \Bmbfhat \leftarrow $ normalized solution dervied from
~\eqref{eq:poly2elicitamatfinal}\\
\textbf{Output:} $\ambfhat, \Bmbfhat$ 
\normalsize \vspace{-0.25em}
}}
\label{alg:q-me}
\end{figure}

\vspace{-0.1cm}
\section{Eliciting Fairness Metrics}
\label{sec:fairme}
\vskip -0.1cm
Having understood the QPME procedure, we now discuss how our proposal can be applied to \emph{quadratic metric elicitation for algorithmic fairness}. 
Like~\cite{hiranandani2020fair}, we consider eliciting a metric that trades-off between predictive performance and fairness violation \citep{kamishima2012fairness, chouldechova2017fair, menon2018cost}. 
However, unlike~\cite{hiranandani2020fair}, 
we handle general quadratic fairness violations and show how QPME can be easily employed to elicit group-fair metrics. 

\vspace{-0.2cm}
\subsection{Fairness Preliminaries}
\label{ssec:fpmebackground}
\vskip -0.1cm
The fairness setting is the same as the one in Section~\ref{sec:quadme} except that we additionally have $m$ groups in the data and use $g \in [m]$ to denote the group membership. The groups are assumed to be disjoint, fixed, and known apriori 
\citep{agarwal2018reductions}.
We will work with a separate (randomized) classifiers $h^g : \Xcal \rightarrow \Delta_k$ for each group $g$, and use 
 $\Hcal^g = \{h^g : \Xcal \rightarrow \Delta_k\}$
 to denote the set of all classifiers for 
 $g$. 
 
\emph{Group predictive rates:} Similar to~\eqref{eq:components}, we denote the group-conditional rates for  $h^g$ 
by $\rmbf^g(h^g, \Pmbb) \in \Rmbb^{k}$, where the $i$-th entry is additionally conditioned on group $g$: 
\begin{align}
	r^g_{i}(h^g, \Pmbb) \coloneqq \Pmbb(h^g = i | Y = i, G= g )\,\forall \,i \in [k].
	\label{eq:f-components}
\end{align}
Analogous to the general setup, 
we denote the set of feasible rates for group $g$  by $\Rcal^g = \{\rmbf^g(h^g, \Pmbb) \,:\, h^g \in \Hcal^g \}$. 

\bexample[Fairness violation]
\emph{
A popular criterion for group fairness is the equal opportunity criterion of \cite{hardt2016equality}, which for a binary classification setup with $m$ protected groups, would require that $r_1^u = r_1^v$ for each pair of groups $(u,v)$. This can be formulated as constraints $|r_1^u - r_1^v|\leq \epsilon$, for some slack $\epsilon$ for all pairs $(u,v)$~\citep{agarwal2018reductions}, or more generally as a regularization term in the learning objective~\citep{bechavod2017learning, hardt2016equality},  by measuring the squared difference between the group rates: $\phi^{\text{EOpp}}((\rmbf^{1},\dots,\rmbf^{m}))
 = {{m\choose 2}}^{-1}\sum_{v>u}(r_1^u - r_1^v)^2$. 
Another popular criterion is {equalized odds}, which requires equal rates across different protected groups 
\citep{bechavod2017learning}.
This again can be specified as a quadratic objective:
$\phi^{\eo}((\rmbf^{1},\dots,\rmbf^{m})) \,=\, {[k{m\choose 2}]}^{-1}\sum_{ v>u}\sum_{i=1}^k \left(r^u_i - r^v_i\right)^2$. 
Other fairness criteria that can be expressed as quadratic metrics 
include 
 {balance for the negative class}, which for a binary classification problem is given by $\phi^{\text{BN}}((\rmbf^{1},\dots,\rmbf^{m}))
  = {{m\choose 2}}^{-1}\sum_{v>u}(r_2^u - r_2^v)^2$~\citep{kleinberg2017inherent}, and the {error-rate balance} $\phi^{\text{EB}}((\rmbf^{1},\dots,\rmbf^{m})) =  {{m\choose 2}}^{-1}\frac{1}{2}\sum_{v>u}(r_1^u - r_1^v)^2 + (r_2^u - r_2^v)^2$~\citep{chouldechova2017fair}  and their weighted variants. 
}
\eexample
\vskip -0.1cm

In the next section, we introduce a general family of metrics that trades-off between an 
error term and a quadratic fairness violation term,
for which we will need to define the rates for the overall classifier.

\emph{Rates for overall classifier:} We construct the overall classifier $h : (\Xcal, [m]) \rightarrow \Delta_k$ by predicting with classifier $h^g$ for group $g$, i.e.\ $h(\xmbf, g) \coloneqq h^g(\xmbf)$.  We will be interested in both the fairness violation and predictive performance of the overall classifier. 
For the former, we will need the $m$ group-specific rates, represented together as a tuple: 
$$\rmbf^{1:m} \coloneqq  (\rmbf^1, \dots, \rmbf^m) \in \Rcal^1 \times \dots \times \Rcal^m =: \prodRcal.$$ 
For the latter, we will measure the overall rates for $h$ as described in~\eqref{eq:components}. The overall rates can also be written in terms of group-specific rates as:
$\rmbf = \sum_{g=1}^m \bm{\tau}^g \odot \rmbf^g,$ where $\bm{\tau}^g$ is just a constant vector whose $i$-th entry denote the prevalence of group $g$ within class $i$, i.e., $\Pmbb(G=g|Y=i)$.

\vspace{-0.2cm}
\subsection{Fair Quadratic Metric Elicitation}
\label{ssec:f-metric}
\vskip -0.1cm

We seek to elicit a metric 
that trades-off between predictive performance (a linear function of overall rates $\rmbf$) and fairness violation (a quadratic function of group rates $\rmbf^{1:m}$). For simplicity, we will denote the fairness metric in cost form, i.e., lower values are better. 

\begin{figure}[t]
    \centering
    \hspace{-0.1cm}
    \includegraphics[scale=0.8]{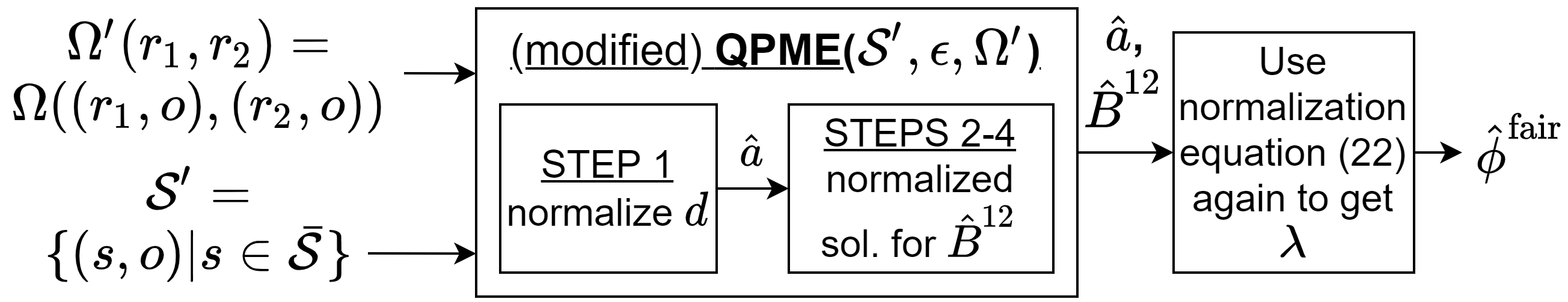}
    \vskip -0.1cm
     \caption{Eliciting Fair Quadratic Metrics 
     for two groups. We formulate a $k$-dimensional elicitation problem and use a variant of QPME (Algorithm~1).
    }
    \label{fig:fairness-workflow}
    \vskip -0.45cm
\end{figure}

\bdefinition \emph{(Fair Quadratic Performance Metric)} For misclassification costs  $\ambf \in \Rmbb^k$, $\ambf \geq 0$, 
fairness violation costs $\mathbb{B} \,=\, \{\Bmbf^{uv} \in PSD_k\}_{u, v=1, v>u}^m$, 
and a trade-off parameter $\lambda \in [0,1]$, we define:
\begin{align*}
&\phi^\fair(\tupr; \ambf, \mathbb{B}, \lambda) \,\coloneqq\, (1-\lambda)\inner{\ambf}{\bm{1} - \rmbf} ~+~
\lambda \frac{1}{2} \left(\sum\nolimits_{v>u} (\rmbf^u - \rmbf^v)^T\mathbbm{\Bmbf}^{uv}(\rmbf^{u} - \rmbf^v)\right)
    \numberthis \label{eq:f-linmetric},
\end{align*}
where w.l.o.g.\
the parameters $\ambf$ and $\Bmbf^{uv}$'s are normalized:
    $\Vert \ambf \Vert_2 = 1, \, \frac{1}{2}\sum_{v>u}^{m} \Vert \Bmbf^{uv} \Vert_F = 1.$
\label{def:f-linmetric}
\edefinition

The coefficients $\ambf, \Bmbf^{uv}$'s are separately normalized so that the predictive performance and fairness violation are in the same scale, and we can additionally elicit the trade-off parameter $\lambda$. Analogous to Definitions \ref{def:query}--\ref{def:me}, the problem of \emph{Fair Quadratic Metric Elicitation} is as follows: given access to pairwise oracle queries of the form $\Omega(\tuprhat_1, \tuprhat_2)$, recover a metric $\hphi^\fair = (\ambfhat, \hat{\mathbb{B}}, \lambdahat)$ such that $\Vert\phi^\fair - \hphi^\fair\Vert < \kappa$ under a  suitable norm $\Vert \cdot \Vert$ for small  $\kappa > 0$. 

Similar to Section \ref{ssec:mpme}, we study the space of feasible rates $\Rcal^{1:m}$ under the following mild assumption. 

\bassumption
For all $g\in[m]$, the conditional distributions $\Pmbb(Y=j|X, G=g), \, j \in [k],$ are distinct, i.e., there is some signal for non-trivial classification for each group.
\label{as:f-sphere}
\eassumption
\bproposition
[Geometry of $\prodRcal$; Figure~\ref{fig:geometry}(b)] For each group $g$, a classifier that predicts class $i$ on all inputs results in the same rate vector $\embf_i$. The rate space $\Rcal^g$ for each group $g$ is convex and so is the intersection 
$\Rcal^1 \cap \dots \cap \Rcal^m$, which also contains the rate profile $\ombf = \tfrac{1}{k} \tiny{\sum_{i=1}^k \embf_i}$ (achieved by the uniform random classifier) in the interior. 
\label{prop:f-C}
\eproposition

\bremark[Existence of sphere $\overline{\Scal}$]
There exists a 
sphere $\overline{\Scal} \subset \Rcal^1 \cap \dots \cap \Rcal^m$ of radius $\rho$ centered at $\ombf$. Thus, a rate $\smbf \in\overline{\Scal}$ is feasible for each of the $m$ groups, i.e.,\ $\smbf$ is achievable by some classifier $h^g$ for each group $g \in [m]$.
\label{as:f-sphere}
\eremark
\vskip -0.1cm

Because we allow a separate classifier for each group, Remark~\ref{as:f-sphere} implies that any rate  $\rmbf^{1:m} = (\smbf^1, \ldots, \smbf^m)$ for arbitrary points $\smbf^1, \ldots, \smbf^m \in \overline{\Scal}$ is achievable for some choice of group-specific classifiers $h^1, \ldots, h^m$. This observation will be key to the elicitation algorithm we describe next.

\vspace{-0.2cm}
\subsection{Eliciting Metric Parameters $({\ambf}, \mathbb{B}, \lambda)$}
\vskip -0.1cm
We present a strategy for eliciting fair metrics 
by adapting the QPME algorithm. For simplicity, we  focus on the $m=2$ case and extend our approach for $m>2$ 
in Appendix \ref{append:sec:fpme}.

Observe that for a rate profile $\rmbf^{1:2} = (\smbf, \ombf)$, where the first group is assigned an arbitrary point in $\overline{\Scal}$ and the second group  is assigned the uniform random classifier's rate $\ombf$, the fair metric~\eqref{eq:f-linmetric} becomes: 
\begin{align*}
\phi^\fair((\smbf, \ombf); \ambf,\, \Bmbf^{12}, \lambda) &\coloneqq (1-\lambda)\inner{\ambf}{\bm{1} - (\bm{\tau}^1 \odot \smbf + \bm{\tau}^2 \odot \ombf)} + 
\frac{\lambda}{2} (\smbf - \ombf)^T\Bmbf^{12}(\smbf - \ombf)\vspace{-0.2cm}\\
& \coloneqq \inner{\dmbf}{\smbf - \ombf} +
\frac{1}{2} (\smbf - \ombf)^T\Bmbf(\smbf - \ombf) \\ 
& \coloneqq \overline{\phi}(\smbf; \dmbf, \Bmbf),
    \numberthis \label{eq:f-linmetricshift}
\end{align*}
where $\dmbf = -(1-\lambda)\taumbf^1\odot\ambf$ and $\Bmbf = \lambda \Bmbf^{12}$, and we use $\taumbf^1 + \taumbf^2 =\bm{1}$ (the vector of ones) for the second step. The metric $\overline{\phi}$ above is a particular instance of the quadratic metric in~\eqref{eq:quadmetshift}.  
We can thus apply a slight variant of the QPME procedure in Algorithm~1 to solve the quadratic metric elicitation problem over the sphere $\Scal' = \{(\smbf, \ombf) \,|\, \smbf \in \overline{\Scal}\}$ with the modified oracle $\Omega'(\rmbf_1, \rmbf_2) = \Omega((\rmbf_1, \ombf), (\rmbf_2, \ombf))$. 

The only change needed for the algorithm is in line 5, where 
we need to account for the changed relationship between $\dmbf$ and $\ambf$ and need to separately (not jointly) normalize the linear and quadratic coefficients. With this change, the output of the algorithm directly gives us the required estimates. 
Specifically, from step 1 of Algorithm~1 and \eqref{eq:0col}, we have $\hat{d}_i = -(1-\lambda)\tau^1_i \hat{a}_i$. By normalizing $\dmbf$, we 
get $\ambfhat = \frac{\dmbf}{\|\dmbf\|}$ for the linear coefficients. 
Similarly, steps 2-4 of Algorithm~1 and \eqref{eq:poly2elicitamatfinal} 
allow us to express $\hat{B}_{ij} =\lambda\hat{B}^{12}_{ij}$ in terms of $\hat{a}_1$. After normalizing we directly get estimates 
$\Bmbfhat^{12} = {\Bmbfhat}/{\|\Bmbfhat\|_F}$ for the quadratic coefficients.

Finally, because the linear and quadratic coefficients are separately  normalized, the estimates $\ambfhat,\, \Bmbfhat^{12}$ are independent of the trade-off parameter $\lambda$.  
Given estimates {\small$\hat{B}^{12}_{ij}$} and $\ahat_1$,  we can now additionally estimate the trade-off parameter {\small$\hat{\lambda}$}. See Appendix \ref{append:sec:fpme} for details 
and Figure \ref{fig:fairness-workflow} for an illustration. 

\begin{figure*}[t]
	\centering 
	\vskip -0.1cm
	\subfigure[]{
        \hspace{-0.3cm}
		{\includegraphics[width=4.5cm]{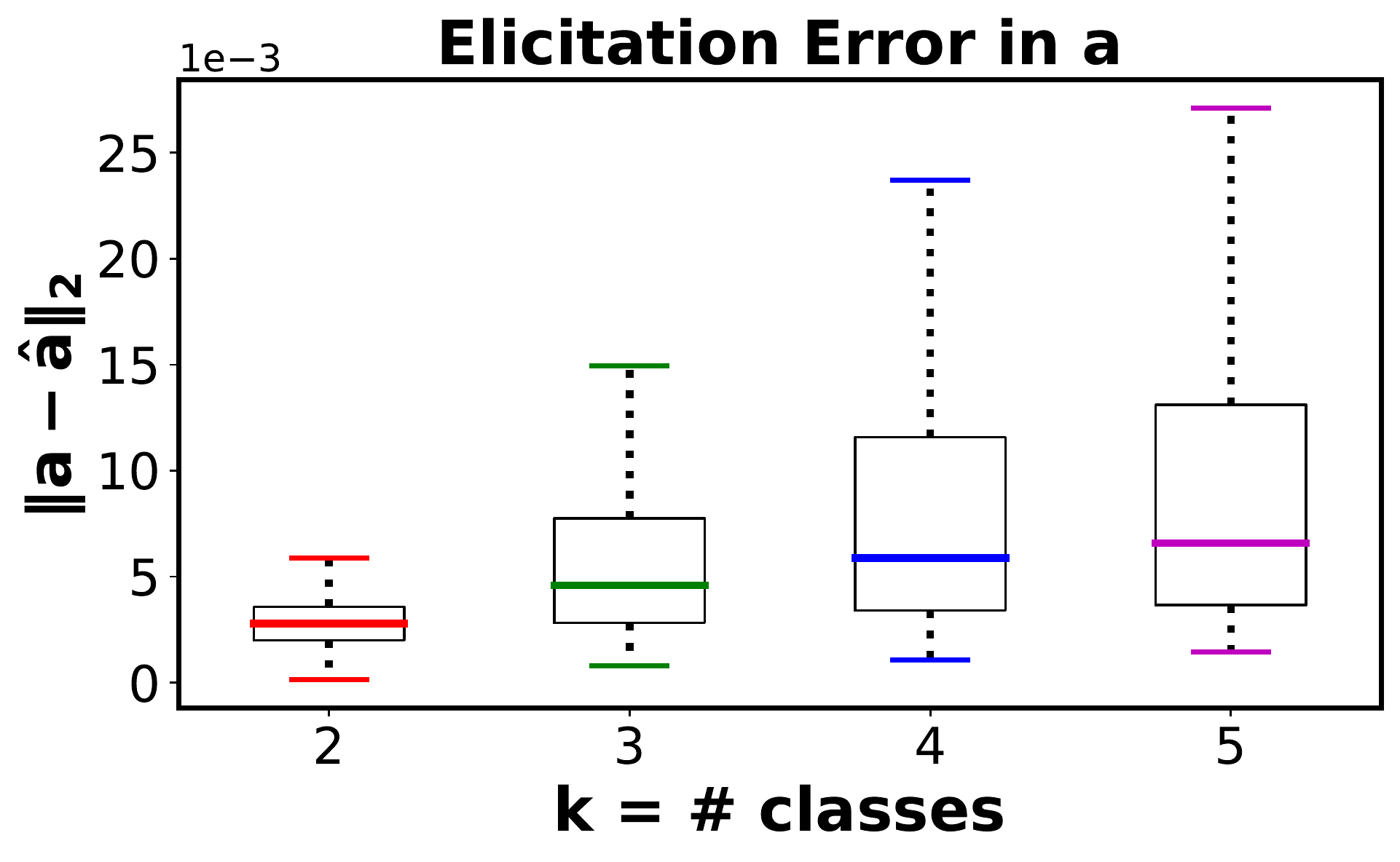}}
		\label{fig:q_rec_a}
	}
	\subfigure[]{
        \hspace{-0.3cm}
		{\includegraphics[width=4.5cm]{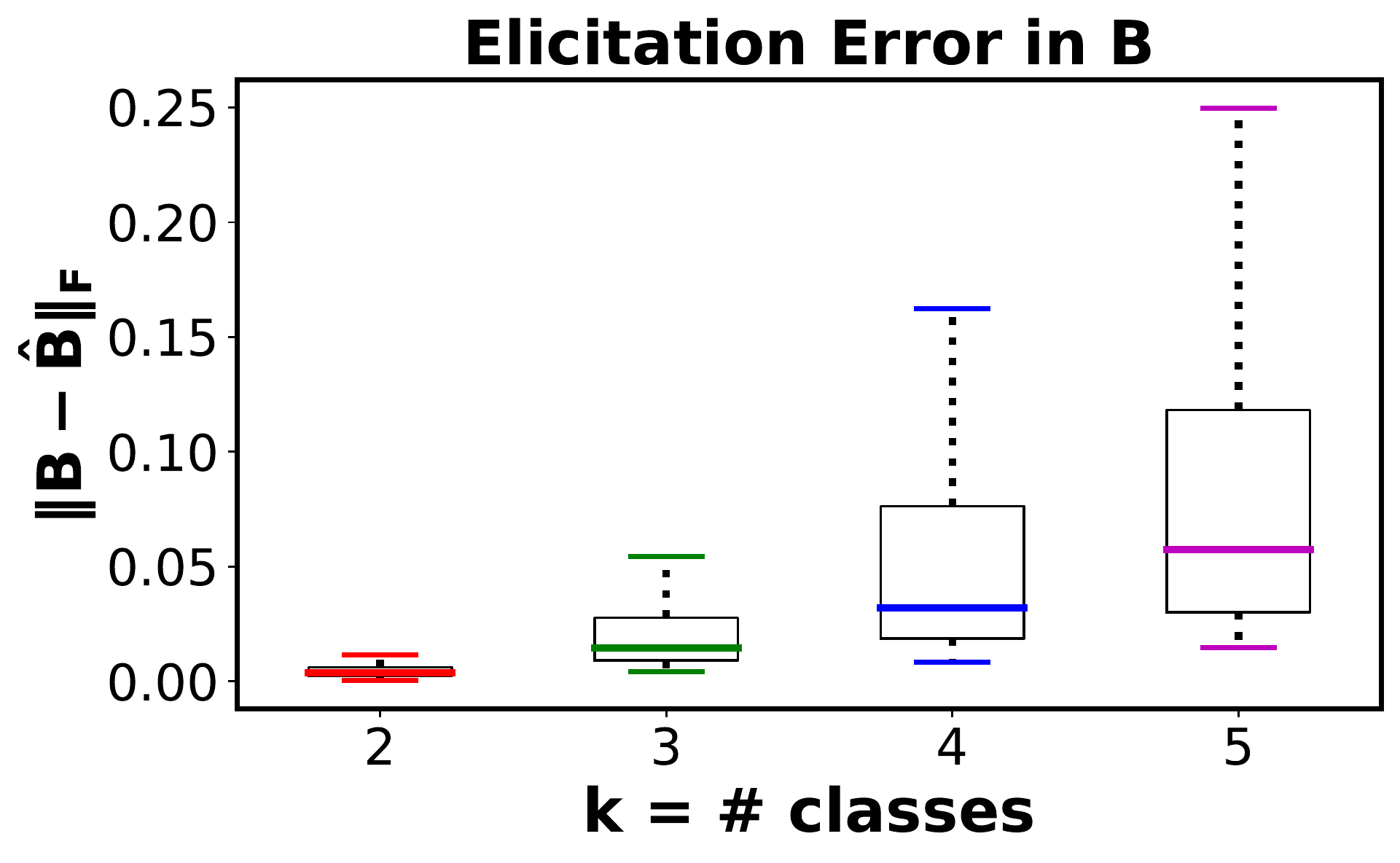}}
		\label{fig:q_rec_B}
	}
	\subfigure[]{
        \hspace{-0.3cm}
		{\includegraphics[width=4.5cm]{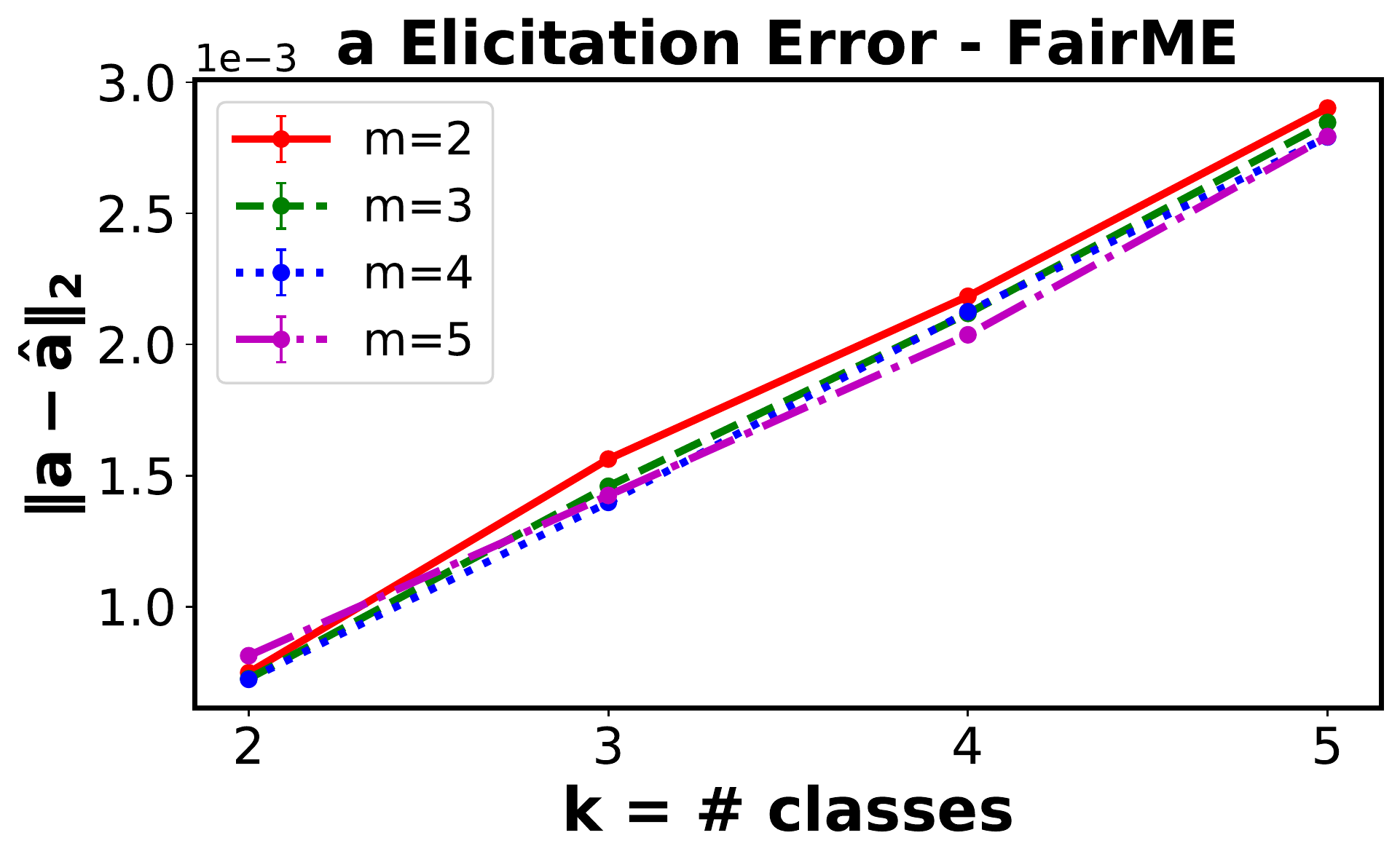}}
		\label{fig:f_rec_a}
	}
	\subfigure[]{
        \hspace{-0.3cm}
		{\includegraphics[width=4.5cm]{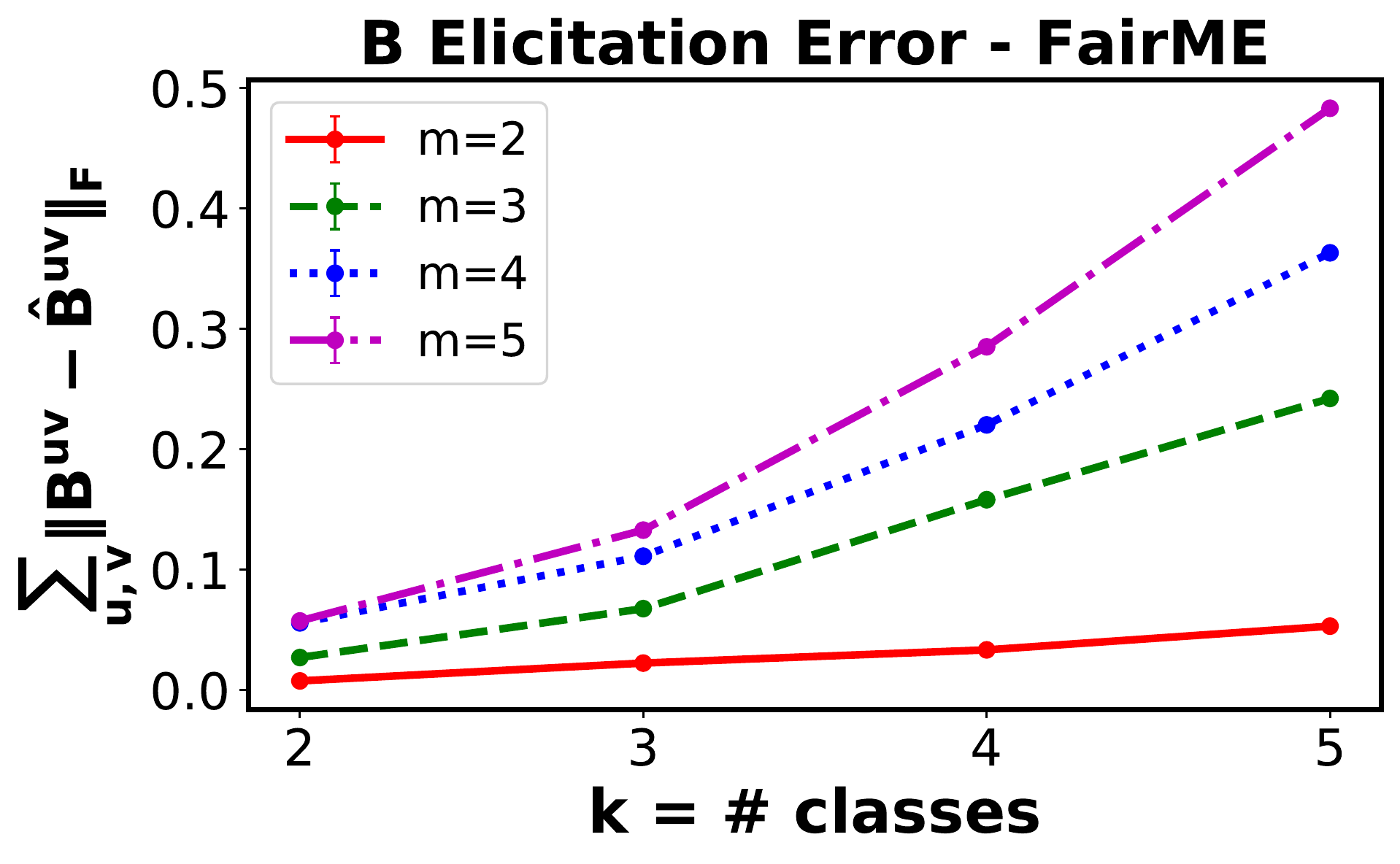}}
		\label{fig:f_rec_B}
	}
	\subfigure[]{
        \hspace{-0.3cm}
		{\includegraphics[width=4.5cm]{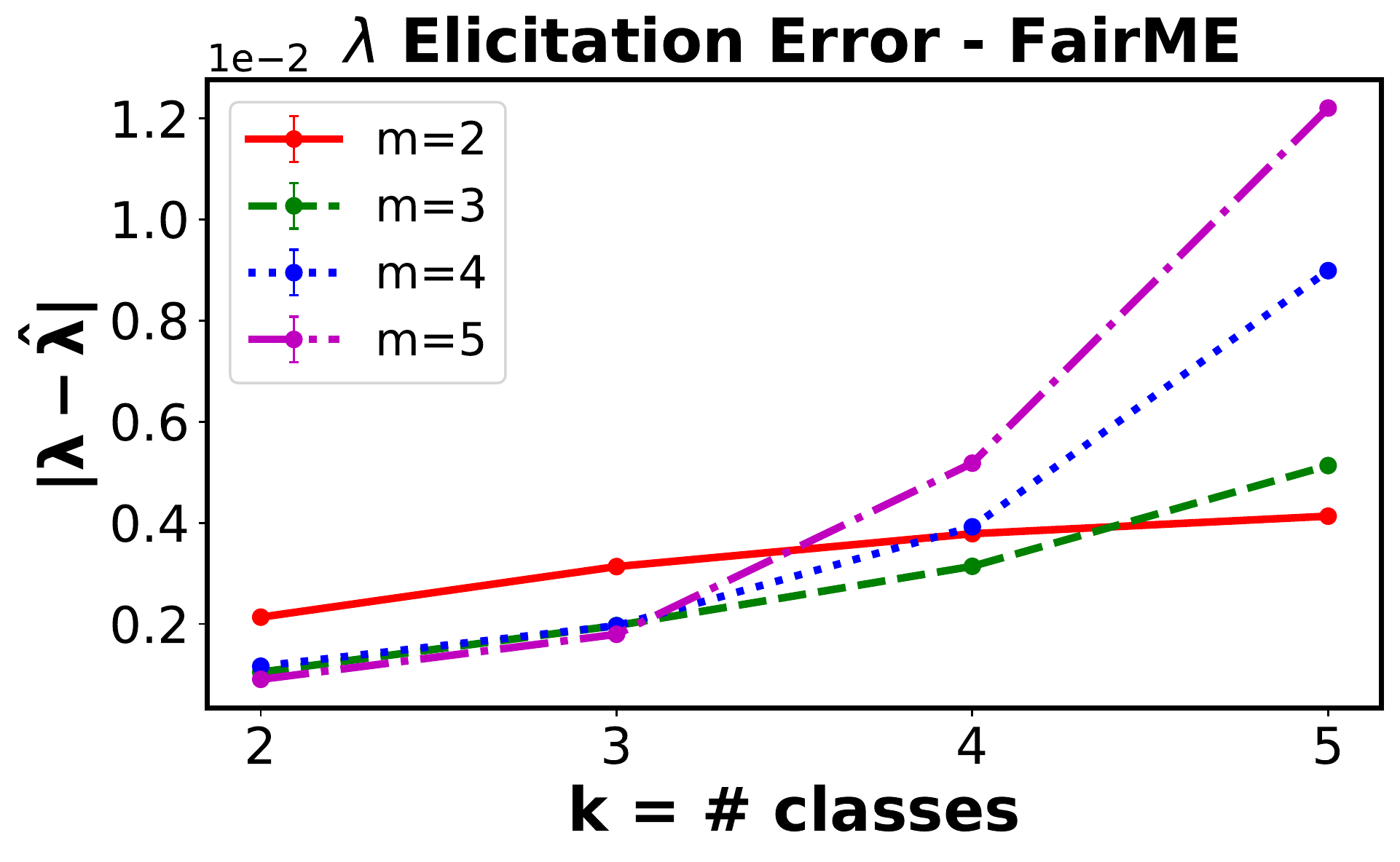}}
		\label{fig:f_rec_l}
	}
	\vskip -0.2cm
	\caption{Average elicitation error over 100 metrics as a function of number of classes $k$ and groups $m$ for quadratic metrics in Definition \ref{def:quadmet} (a--b) and fairness metrics in Definition \ref{def:f-linmetric} (c--e). See Table~\ref{tab:numqueries} in Appendix \ref{append:sec:extexp} for the number of queries needed.
	}
	\label{fig:recovery}
	\vskip -0.15cm
\end{figure*}

\section{Guarantees}
\label{sec:guarantees}

We discuss guarantees for the QPME procedure under the following practically relevant feedback model. 
The fair metric elicitation guarantees follow 
as a consequence.

\bdefinition[Oracle Feedback Noise: $\epsilon_\Omega \geq 0$] Given rates $\rmbf_1, \rmbf_2$, 
the oracle responds correctly iff $|\phi^{\quadr}(\rmbf_1) - \phi^{\quadr}(\rmbf_2)| > \epsilon_\Omega$ and may be incorrect otherwise.
\label{def:noise}
\edefinition

In words, the oracle may respond incorrectly if the rates are close 
 as measured by the metric. 
Since eliciting the metric involves offline computations 
of ratios, 
we make a 
regularity assumption
ensuring that all components are well defined. 
\bassumption
For the shifted quadratic metric $\bphi$ in~\eqref{eq:quadmetshift},  
the gradients at the rate profiles $\ombf$, $-\zmbf_1$, and $\{\zmbf_1, \dots, \zmbf_q\}$, are non-zero vectors. 
Additionally, $\rho > \varrho \gg \epsilon_\Omega$.
\label{as:regularity-q}
\eassumption

\btheorem
Given $\epsilon,\epsilon_\Omega\geq 0$, and a 1-Lipschitz metric $\phi^{\quadr}$ (Def.\ \ref{def:quadmet}) parametrized by $\ambf, \Bmbf$, under Assumptions~\ref{assump:distribution},  \ref{assump:smoothness},  and \ref{as:regularity-q}, after $O\left(k^2\log \tfrac 1 {\epsilon}\right)$ queries, Algorithm~1 returns a metric $\hphi^{\quadr} = (\ambfhat, \Bmbfhat)$ with
    $\Vert \ambf-\ambfhat \Vert_{2}\leq O\left(\sqrt{k}(\epsilon+\sqrt{\varrho+\epsilon_\Omega/\varrho})\right)$ 
    and 
    $\Vert \Bmbf -\Bmbfhat \Vert_{F}\leq O\left(k\sqrt{k}(\epsilon+\sqrt{\varrho + \epsilon_\Omega/\varrho})\right)$.
\label{thm:q-me}
\etheorem
The proof of Theorem~\ref{thm:q-me} uses the guarantee for LPME \emph{only} as an intermediate step, and substantially builds on it to take into account the smoothness of the non-linear metric, the multiplicative errors in the slopes, and the feedback noise. We also provide a \emph{finite sample version} of Theorem~\ref{thm:q-me} in Corollary~\ref{append:cor:finite} (Appendix~\ref{append:sec:guarantees}), which states that the above result holds with high probability as long as (i) the hypothesis class of classifiers has finite capacity, and (ii) the number of samples used to estimate the rates is large enough. 
\btheorem(\textbf{Lower Bound})
For any $\epsilon > 0$, at least $\Omega(k^2\log(1/(k\sqrt k\epsilon)))$ pairwise queries are needed to 
to elicit a quadratic metric (Def.\ \ref{def:quadmet})
to an error tolerance of $k\sqrt k\epsilon$. 
\label{thm:lb}
\etheorem
Theorem~\ref{thm:q-me} shows that the QPME procedure is robust to noise and its query complexity depends only \emph{linearly} in the number of unknowns. Theorem~\ref{thm:lb} shows that the inherent complexity of the problem depends on the \emph{number of unknowns}, thus our query complexity is optimal (barring the log term). So the $\tilde{O}(k^2)$ complexity is merely an artifact of our setup  in Definition~\ref{def:quadmet} being very general (with $O(k^2)$ unknowns).
Indeed, with added structural assumptions on the metric, our proposal can be modified to considerably reduce the query complexity. For example, if we know that the matrix $\Bmbf$ is diagonal, then each  LPME subroutine call  
needs to estimate only one parameter, which can be done with a constant number of queries, requiring a total of only {\small$\tilde O(k)$} queries. 
We also stress that despite eliciting a more complex (non-linear) metric, the query complexity is still \emph{linear in the number of unknowns}, which is same as prior linear elicitation methods ~\citep{hiranandani2018eliciting, hiranandani2019multiclass}. 

\section{Experiments}
\label{sec:experiments}

We evaluate our approach on simulated oracles. Here we present results on a synthetically generated query space and in Appendix \ref{append:ssec:ranking} include results on real-world datasets. 

\textbf{Eliciting quadratic metrics.}\ 
We first apply QPME (Algorithm~1) to elicit quadratic metrics in Definition \ref{def:quadmet}.  Like~\cite{hiranandani2020fair}, we assume access to a $k$-dimensional sphere $\Scal$ centered at rate $\ombf$ with radius $\rho = 0.2$, from which we query rate vectors $\rmbf$. The trends that we will discuss are robust to the sphere radius parameter $\rho$.
Recall that in practice,
Remark \ref{rem:sphere} guarantees the existence of such a sphere within the feasible region $\Rcal$. We  randomly generate quadratic metrics $\phi^\quadr$ parametrized by $(\ambf, \Bmbf)$ and repeat the experiment over 100 trials for varying  numbers of classes $k \in \{2,3,4,5\}$. 
We run the QPME procedure with tolerance $\epsilon = 10^{-2}$. In Figures~\ref{fig:q_rec_a}--\ref{fig:q_rec_B}, we show box plots 
of the $\ell_2$ (Frobenius) norm between the true and elicited linear (quadratic) coefficients. 
We  generally find that QPME is able to elicit metrics  close to the true ones.
This holds for varying $k$, showing the effectiveness of our approach in handling multiple classes. 
The average number of queries we needed for elicitation over the 100 trials is provided in Table~\ref{tab:numqueries} in Appendix \ref{append:sec:extexp}. Note that the number of queries is $\tilde O(d)$ for eliciting a quadratic metric with $d = k^2$ unknowns,
which clearly matches the lower bound in Theorem~\ref{thm:lb}. See Appendix \ref{append:practicality} for a discussion on the practicality of posing the requisite number of queries.

\textbf{Eliciting fairness metrics.}\
We next apply the elicitation procedure in Figure \ref{fig:fairness-workflow} with tolerance $\epsilon= 10^{-2}$ to elicit the fairness metrics in Definition \ref{def:f-linmetric}. We randomly generate oracle metrics $\phi^\fair$ parametrized by $(\ambf, \Bmbb, \lambda)$ and repeat the experiment over 100 trials and with varied  number of classes and groups $k, m \in  \{2,3,4,5\}$. Figures~\ref{fig:f_rec_a}--\ref{fig:f_rec_l} show the mean elicitation errors for the the three parameters. For the linear predictive performance, the error {\small$\Vert \ambf - \ambfhat\Vert_2$} increases only with the number of classes $k$ and not groups $m$, as it is independent of the number of groups. For the quadratic violation term, the error {\small$\sum_{u,v}\Vert \Bmbf^{uv} - \Bmbfhat^{uv} \Vert_F$} increases with both $k$ and $m$. This is because the QPME procedure is run {\small$m\choose 2$} times for eliciting {\small $m \choose 2$} matrices {\small $\{\Bmbf^{uv}\}_{v > u}$}, and so the elicitation error accumulates with increasing $q$. Lastly,  the elicited trade-off {\small $\hat \lambda$} is seen to be close to the true $\lambda$ as well. 

\textbf{Real-world datasets.}\ In App.~\ref{append:ssec:ranking}, we evaluate how well the elicited metric from QPME ranks a set of candidate classifiers trained on real-world datasets. 
We find that despite incurring elicitation errors, QPME 
achieves near-perfect ranking; 
whereas, the  baseline metrics fail to do so. 

\vspace{-0.1cm}
\section{Related Work}
\label{sec:relatedwork}
\cite{hiranandani2018eliciting} formalized the problem of ME for binary classification with (quasi-)linear metrics and later extended it to the multiclass setting~\citep{hiranandani2019multiclass}. Unlike them, we elicit more complex quadratic metrics, and also provide an information-theoretic lower bound on the query complexity (Theorem~\ref{thm:lb}). 
Prior works on ME 
offer no such lower bound guarantees. 
Learning linear functions passively using pairwise comparisons is a mature field 
\citep{joachims2002optimizing, peyrard2017learning}, 
but unlike their active learning counter-parts ~\citep{settles2009active, kane2017active}, these methods are not query-efficient. 
Studies such as~\cite{qian2015learning} provide  active linear elicitation strategies 
but with no guarantees and also work with a different query space. We are unaware of prior work that \emph{provably} elicit a quadratic function, either passively 
or
actively using pairwise comparisons. Our work is thus a significant first step towards active, nonlinear metric elicitation.

The use of metric elicitation for fairness is relatively new, with some work on eliciting \textit{individual} fairness metrics~\citep{ilvento2019metric, mukherjee2020two}. \cite{hiranandani2020fair} is the only work we are aware of that elicits \textit{group-fair} metrics, which we extend to handle more general 
metrics. 
\cite{zhang2020joint} 
elicit  the trade-off between accuracy and fairness using complex ratio queries. In contrast, we jointly  elicit the predictive performance, fairness violation, and trade-off 
using simpler pairwise queries. 
Lastly, there has been work on  learning fair classifiers under constraints 
\citep{zafar2017constraints,agarwal2018reductions}.
We take the regularization view of fairness, where the fairness violation is included in the objective 
\citep{kamishima2012fairness}.

Our work is also related to decision-theoretic \emph{preference elicitation}, however, with the following key  differences. We focus on estimating the utility function (metric) explicitly, whereas prior work such as~\citep{boutilier2006constraint, benabbou2017incremental} seek to find the optimal decision via minimizing the max-regret over a set of utilities. Studies that directly learn the utility~\citep{perny2016incremental} do not provide query complexity guarantees for pairwise comparisons. Formulations that consider a finite set of alternatives~\citep{boutilier2006constraint} are starkly different from ours, because the set of alternatives in our case (i.e. classifiers or rates) is infinite. 
Most of the papers in this literature focus on linear or bilinear~\citep{perny2016incremental} utilities except for~\citep{braziunas2012decision} (GAI utilities) and~\citep{benabbou2017incremental} (Choquet integral); whereas, we focus on quadratic metrics which are useful for classification tasks, especially, fairness. We are not aware of any decision-theory literature that \emph{provably} elicits quadratic (or polynomial) utility functions using pairwise comparisons.

Eliciting performance metrics bears similarities to \emph{learning reward functions} in the inverse reinforcement learning literature~\citep{wu2020efficient,abbeel2004apprenticeship,levine2011nonlinear,sadigh2017active} and the \emph{Bradley-Terry-Luce model with features} in the learning-to-rank literature~\citep{shah2015estimation, niranjan2017inductive}. However, in summary, these studies focus on either eliciting linear utilities or passively learning utility functions. Our work is substantially different from them as we are tied to the geometry of the space of classification error statistics, and elicit quadratic  utility functions using only pairwise comparisons, and particularly, in an active learning fashion. Moreover, we also provide query complexity bounds along with a lower bound. We further elaborate on the specific differences from these papers in  Appendix~\ref{append:sec:relwork}.

\section{Discussion}
\label{sec:discussion}

We have provided an efficient quadratic metric elicitation strategy with application to  fairness, and with a query complexity that has the same dependence on the number of unknowns as that for linear metrics. 

\textbf{Higher Order Polynomials:} We next show how our approach can be extended to elicit \emph{higher-order polynomial} metrics. Thus our work not only increases the use-cases for ME but also opens the door for non-linear metric elicitation in other fields such as \emph{active learning.}

Consider, e.g., a cubic polynomial:
\begin{align*}
    \phi^{\text{cubic}}(\rmbf)\coloneqq \sum_{i}a_ir_i + \frac{1}{2}\sum_{i,j}B_{ij}r_ir_j + \frac{1}{6}\sum_{i,j,l}C_{ijl}r_ir_jr_l,
\end{align*}
where $\Bmbf$ and $\Cmbf$ are symmetric, and $\sum_i a_i^2 +\sum_{ij} B_{ij}^2 + \sum_{ijl} C_{ijl}^2 = 1$ (w.l.o.g., due to scale invariance).  A quadratic approximation to this metric around a point $\zmbf$ is given by: 
\begin{align*}
&\frac{1}{2}\left(\sum_{i,j}B_{ij}r_ir_j + \sum_{i,j,l}C_{ijl}(r_i - z_i)(r_j - z_j)z_l\right) + \\
&\quad \; \sum_{i}a_ir_i + c,
\end{align*}
where $c$ is a constant not affecting the oracle responses. We can estimate the parameters of this  approximation by applying the QPME procedure from Algorithm~1 with the metric centered at an appropriate point, and its queries restricted to a small neighborhood around $\zmbf$. Running QPME once using a sphere around the point $\zmbf_l = \ombf + (\varrho - \varrho')\alphambf_l$, where $\varrho' < \varrho$ will elicit one face of the tensor $\Cmbf_{[:, :, l]}$ upto a scaling factor. Thus, it will require us to run the QPME procedure $k$ times around the basis points $\zmbf_l = \ombf + (\varrho - \varrho')\alphambf_l \; \; \forall l \in [k]$. Since we elicit scale-invariant quadratic approximation, we would need additional run of QPME procedure around the point $\Scal_{-\zmbf_1}$ to elicit all the coefficients. Thus, we can recover the metric $\hat{\phi}^{\text{cubic}} = (\hat{\ambf}, \hat{\Bmbf},\hat{\Cmbf})$ with as many queries as the number of unknowns, i.e, $\tilde O(k^3)$ in the cubic case. 

For a $d$-th order polynomial, one can recursively apply this procedure to estimate $(d-1)$-th order approximations at multiple points, and similarly derive the polynomial coefficients from the estimated local approximations.

\textbf{Handling large number of classes:} For applications where $k$ is very large, the parameterization discussed in Section \ref{sec:background} may not be  applicable in its current form. For example, when $k=1000$, the quadratic metric in \eqref{eq:quadmet} would use $O(1000^2)$ parameters, an exorbitantly high number to elicit in practice. Note that
the presence of $k^2$ unknowns is an  artifact of the problem formulation, and \emph{not} of our proposed procedure.
Moreover, as shown in Theorem \ref{thm:lb}, it is \emph{theoretically impossible} to estimate $k^2$ unknowns with fewer than $\tilde{O}(k^2)$ queries. While our QPME procedure does indeed match this lower bound, in practice, we do not expect it to be applied to estimate such an over-parameterized metric. Instead, for such large-scale settings, we  recommend making reasonable assumptions on the metric to reduce the number of unknowns, e.g., by having multiple classes share the same parameter, and the query complexity of QPME would then only depend \emph{linearly} on the reduced number of unknowns. For instance, in Table~\ref{append:tab:apxranking} (Appendix~\ref{append:ssec:ranking}), we show  that by simplifying the metric with structural assumptions, one can use fewer queries in practice to get comparable results. 

\textbf{Advantages:}\ Our proposal comes with many practical advantages: (a) \emph{Fairness:} we are aware of no prior work that can elicit fair quadratic metrics, particularly with provable guarantees; (b) \emph{Transportability:} our method is independent of the population $\Pmbb$, which allows any metric that is elicited using one dataset or model class to be applied to other applications, as long as the expert believes the tradeoffs to be the same; and (c) \emph{Feasibility:} we ensure that the rates are feasible throughout the elicitation (i.e., are achievable by classifiers), which allows  the flexibility to deploy systems that either compare classifiers or compare rates.

\textbf{Limitations:} Limitations of our work include the assumption that the metric has a parametric form, which can be restrictive in some cases, and not providing a concrete answer to who the oracles should be. One should also be cautious in applying ME to eliciting fairness metrics, as failures here could exacerbate the adverse effects on protected groups. 

\textbf{Future Work:} To thoroughly answer the above questions, we are actively conducting user studies on collecting preference feedback using intuitive visualizations of rates \citep{shen2020designing,beauxis2014visualization} or classifiers \citep{ribeiro2016should} . Please see Appendix~\ref{append:userstudy} to take a peek into the future work, where we discuss findings from a preliminary user study.

\section*{Acknowledgements} 
This research was funded by Google Research. The authors would like to thank Safinah Ali, Sohini Upadhyay, and Elena Glassman for helping with the pilot user study discussed in Appendix~\ref{append:userstudy}.

\bibliography{references}
\bibliographystyle{plainnat}

\clearpage

\appendix

\section{Linear Performance Metric Elicitation (LPME)}
\label{append:sec:slme}

In this section, we shed more light on the procedure from~\citep{hiranandani2019multiclass} that elicits a multiclass linear metric. We call it the Linear Performance Metric Elicitation (LPME) procedure.  As discussed in Algorithm~1, we use this as a subroutine to elicit metrics in the quadratic family. 

LPME exploits the enclosed sphere $\Scal \subset \Rcal$ for eliciting linear multiclass metrics. Let the sphere $\Scal$'s radius be $\rho>0$, and the oracle's scale invariant metric be $\phi^{\text{lin}}(\rmbf) \coloneqq \inner{\ambf}{\rmbf}$ such that $\Vert \ambf \Vert_2=1$. The oracle queries are $\tiny{\Omega\left( \rmbf_1, \rmbf_2 \,;\, \phi^{\text{lin}} \right) \coloneqq \1[\phi^{\text{lin}}(\rmbf_1) > \phi^{\text{lin}}(\rmbf_2)]}$. We first outline a trivial Lemma from~\citep{hiranandani2019multiclass}.

\blemma~\citep{hiranandani2019multiclass}
Let a normalized vector $\ambf$ with $\Vert \ambf \Vert_2 =1$ parametrize a linear metric $\phi^{\text{lin}} \coloneqq \inner{\ambf}{\rmbf}$, then the unique optimal rate $\rmbfbar$ over $\Scal$ is a rate on the boundary of $\Scal$ given by $\rmbfbar = \rho \ambf +\ombf$, where $\ombf$  is the center of  $\Scal$. 
\label{lem:spherebayes}
\elemma
\vskip -0.2cm

\begin{figure}[t]
\begin{minipage}[h]{\textwidth}
  \centering \hspace{-0.5em}
  \begin{minipage}[h]{.45\textwidth}
     \centering
\fbox{\parbox[t]{1\textwidth}{\vspace{0.0cm}\scriptsize{\underline{\bf Subroutine \emph{ShrinkInterval}}\normalsize}    \\
\scriptsize
\textbf{Input:} Oracle $\Omega$ and rate profiles $\rmbfbar^{(a)},\rmbfbar^{(c)},\rmbfbar^{(d)},\rmbfbar^{(e)},\rmbfbar^{(b)}$\\
Query $\Omega(\rmbfbar^{(a)}, \rmbfbar^{(c)}\,;\,\phi^{\text{lin}})$.\\
\textbf{If} \, ($\rmbfbar^{(a)} \succ \rmbfbar^{(c)}$) Set $\theta_j^{(b)} = \theta_j^{(d)}$.\\
\textbf{else} \, Query $\Omega(\rmbfbar^{(c)}, \rmbfbar^{(d)}\,;\,\phi^{\text{lin}})$.\\
\text{\,\,\,\,} \textbf{If} \, ($\rmbfbar^{(c)} \succ \rmbfbar^{(d)}$) Set $\theta_j^{(b)} = \theta_j^{(d)}$.\\
\text{\,\,\,\,} \textbf{else} \, Query $\Omega(\rmbfbar^{(d)}, \rmbfbar^{(e)}\,;\,\phi^{\text{lin}})$.\\
\text{\,\,\,\,}\text{\,\,\,\,} \textbf{If} \, ($\rmbfbar^{(d)} \succ \rmbfbar^{(e)}$) Set $\theta_j^{(a)} = \theta_j^{(c)}$ and $\theta_j^{(b)} = \theta_j^{(e)}$.\\
\text{\,\,\,\,}\text{\,\,\,\,} \textbf{else} Set $\theta_j^{(a)} = \theta_j^{(c)}$.\\
\textbf{Output:} $[\theta_j^{(a)}, \theta_j^{(b)}]$.  
\normalsize \vspace{-0.07cm}
}}
  \end{minipage} \hspace{0.3em}
  \begin{minipage}[h]{.47\textwidth}
     \centering
\vspace{-0.05cm}
\fbox{\parbox[t]{0.95\textwidth}{
\begin{tikzpicture}[scale = 3.15]
    

    	\begin{scope}[shift={(-5.0,0)},scale = 0.483]\scriptsize

\def\r{0.06};
	
    
    
    \draw[thick] (0,0) .. controls (1.8,0) and (2.6,1.6) .. (3.2,1.6) 
    ..controls (3.6,1.6) and (3.8,0) .. (4,0);
    
    \draw[-latex] (0,-.1)--(0,2.505); 
    \draw[-latex] (-0.1,0)--(4.4,0);
    \node[left] at (0,2.35) {$\phi^{\text{lin}}$};
    \node[below right] at (4.1,0) {$\theta_j$};
   
    
    \coordinate (C1) at (0,0.00);
    \coordinate (C2) at (1,0.18);
    \coordinate (C3) at (2,0.76);
    \coordinate (C4) at (3,1.56);
    \coordinate (C5) at (4,0.00);
    
    \node[below] at (0,0) {$\theta_j^{(a)}$};
    \node[below] at (1,0) {$\theta_j^{(c)}$};
    \node[below] at (2,0) {$\theta_j^{(d)}$};
    \node[below] at (3,0) {$\theta_j^{(e)}$};
    \node[below] at (4,0) {$\theta_j^{(b)}$};
    
    \foreach \x in {1,2,3,4} {
    	\draw (\x,-.1) -- (\x,.1);
        \draw[dotted] (\x,0) -- (\x,2);
    }
    \fill[color=black] 
    		(C1) circle (\r)
    		(C2) circle (\r)
            (C3) circle (\r)
            (C4) circle (\r)
            (C5) circle (\r);   
    
    
    \draw[very thick,-latex] (0.3,0.3) -- (0.7,0.7);
    \draw[very thick,-latex] (1.3,0.5) -- (1.7,0.9);
    \draw[very thick,-latex] (2.3,0.3) -- (2.7,0.7);
    \draw[very thick,-latex] (3.3,0.7) -- (3.7,0.3);
    
    
    \draw (2,1.8)--(2,2.2) (4,1.8)--(4,2.2);
    \draw[<->] (2,2)--(4,2);
    
    \end{scope}
\end{tikzpicture}
\vspace{-0.15cm}}
}
  \end{minipage}
\end{minipage}
\caption{(Left): The \emph{ShrinkInterval} subroutine used in line 16 of Algorithm~\ref{alg:slme} (Right): Visual illustration of the subroutine \emph{ShrinkInterval}~\citep{hiranandani2019multiclass}; \emph{ShrinkInterval} shrinks the current search interval to half based on oracle responses to at most three queries.}
\vskip -0.4cm
\label{append:fig:shrink1}
\end{figure}
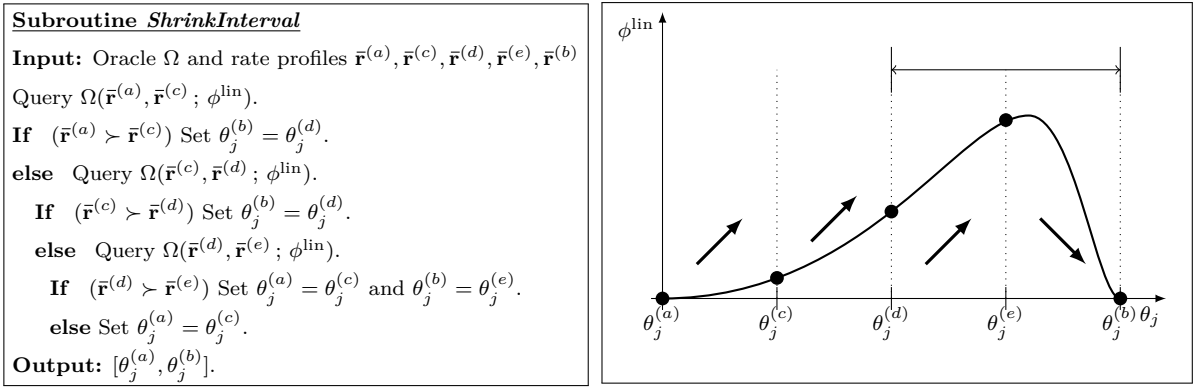

\addtocounter{algorithm}{1}
\balgorithm[H]
\caption{Linear Performance Metric Elicitation}
\label{alg:slme}
\small
\balgorithmic[1]
\STATE \textbf{Input:} Query space $\Scal \subset \Rcal$, binary-search tolerance $\epsilon > 0$, oracle $\Omega(\cdot, \cdot\,;\, \phi^{\text{lin}})$ with metric $\phi^{\text{lin}}$\\ \hfill\\
\FOR{$i = 1, 2, \cdots k$} 
\STATE Set $\ambf = \ambf' = (1/\sqrt{k}, \dots, 1/\sqrt{k})$.
\STATE Set $a'_i = -1/\sqrt{k}$.
\STATE Compute the optimal $\sbar^{(\ambf)}$ and $\sbar^{(\ambf')}$ over the sphere $\Scal$ using Lemma~\ref{lem:spherebayes}
\STATE Query $\Omega(\smbfbar^{(\ambf)}, \smbfbar^{(\ambf')} ; \phi^{\text{lin}})$\\
\ENDFOR
\COMMENT{These queries reveal the search orthant}\\ \hfill \\
\STATE Start with coordinate $j=1$.
\STATE\textbf{Initialize:} $\bm{\theta} = \bm{\theta}^{(1)}$ \hfill \COMMENT{$\bm{\theta}^{(1)}$ is a point in the search orthant.}
\FOR{$t=1, 2, \cdots, T=3(k-1)$}
\STATE Set $\bm{\theta}^{(a)} = \bm{\theta}^{(c)}=\bm{\theta}^{(d)}=\bm{\theta}^{(e)}=\bm{\theta}^{(b)} = \bm{\theta}^{(t)}$.\\
\STATE Set $\theta_j^{(a)}$ and $\theta_j^{(b)}$ to be the min and max angle, respectively, based on the search orthant
\WHILE{$\abs{\theta^{(b)}_j - \theta^{(a)}_j} > \epsilon$}
\STATE Set $\theta^{(c)}_j = \frac{3 \theta^{(a)}_j + \theta^{(b)}_j}{4}$, $\theta^{(d)}_j = \frac{\theta^{(a)}_j + \theta^{(b)}_j}{2}$, and $\theta^{(e)}_j = \frac{\theta^{(a)}_j + 3 \theta^{(b)}_j}{4}$.
\STATE Set $\rmbfbar^{(a)} = \mu(\bm{\theta}^{(a)})$ (i.e. parametrization of $\partial \Scal$). Similarly, set $\rmbfbar^{(c)}, \rmbfbar^{(d)}, \rmbfbar^{(e)}, \rmbfbar^{(b)}$
\STATE $[\theta^{(a)}_j, \theta^{(b)}_j] \leftarrow$ \emph{ShrinkInterval} ($\Omega, \rmbfbar^{(a)},\rmbfbar^{(c)},\rmbfbar^{(d)},\rmbfbar^{(e)},\rmbfbar^{(b)}$)\hfill \COMMENT{see Figure~\ref{append:fig:shrink1}}
\ENDWHILE
\STATE Set $\theta^{(d)}_j = \frac{1}{2}(\theta^{(a)}_j+\theta^{(b)}_j)$ \\
\STATE Set $\bm{\theta}^{(t)} = \bm{\theta}^{(d)}$.
\STATE Update coordinate $j \leftarrow j + 1$ cyclically. 
\ENDFOR
\STATE \textbf{Output:} $\hat a_i =\Pi_{j=1}^{i-1} \sin\theta_j^{(T)} \cos{\theta_i}^{(T)} \, \forall i \in [k-1],\;\hat a_k =\Pi_{j=1}^{k-1} \sin\theta_j^{(T)}$
\ealgorithmic
\ealgorithm

Lemma~\ref{lem:spherebayes} provides a way to define a one-to-one correspondence between a  linear performance metric and its optimal rate over the sphere. That is, given a linear performance metric, using Lemma~\ref{lem:spherebayes}, we may get a unique point in the query space lying on the boundary of the sphere $\partial\Scal$. Moreover, the converse is also true; i.e., given a feasible rate on the boundary of the sphere $\partial\Scal$, one may recover the linear metric for which the given rate is optimal. Thus, for eliciting a linear metric,~\cite{hiranandani2019multiclass} essentially search for the optimal rate (over the sphere $\Scal$) using pairwise queries to the oracle. The optimal rate by virtue of Lemma~$\ref{lem:spherebayes}$ reveals the true metric. The LPME subroutine is summarized in Algorithm~\ref{alg:slme}. Intuitively, Algorithm~\ref{alg:slme} minimizes a strongly convex function denoting distance of query points from a supporting hyperplane whose slope is the true metric (see Figure~2(c) in~\citep{hiranandani2019multiclass}). The procedure also uses the following standard paramterization for the surface of the sphere $\partial\Scal$. 

\textbf{Parameterizing the boundary of the enclosed sphere $\partial \Scal$.} 
Let $\thetambf$ be a ($k-1$)-dimensional vector of angles. In $\thetambf$, all the angles except the primary angle are in $[0, \pi]$, and the primary angle is in $[0, 2\pi]$. A scale invariant linear performance metric with $\Vert \ambf \Vert_2=1$ can be constructed by assigning $a_i = \Pi_{j=1}^{i-1} \sin\theta_j \cos{\theta_i}$ for $i \in [k-1]$ and $a_k = \Pi_{j=1}^{k-1} \sin\theta_j$. Since we can easily compute the metric's optimal rate over $\Scal$ using Lemma~\ref{lem:spherebayes}, by varying $\thetambf$ in this procedure, we parametrize the surface of the sphere $\partial\Scal$. We denote this parametrization by $\mu(\thetambf)$, where $\mu: [0, \pi]^{k-2} \times [0, 2\pi] \to \partial \Scal$.

\emph{Description of Algorithm~\ref{alg:slme}:} Let the oracle's metric be $\phi^{\text{lin}} = \inner{\ambf}{\rmbf}$ such that $\Vert \ambf \Vert_2=1$ (Section~\ref{ssec:mpme}). Using the parametrization $\mu(\thetambf)$ for the boundary of the sphere $\partial \Scal$,  Algorithm~\ref{alg:slme} returns an estimate $\ambfhat$ with $ \Vert \ambfhat \Vert_2=1$. Line 2-6 recover the search orthant of the optimal rate over the sphere by posing $k$ trivial queries. Once the search orthant of the optimal rate is known, the algorithm in each iteration of the for loop (line 9-18) updates one angle $\theta_j$ keeping other angles fixed 
by the \emph{ShrinkInterval} subroutine. The \emph{ShrinkInterval}  subroutine (illustrated in Figure~\ref{append:fig:shrink1}) is binary-search based routine that shrinks the interval $[\theta^a_j, \theta^b_j]$ by half based on the oracle responses to (at most) three queries.\footnote{The description of the binary search algorithm in~\cite{hiranandani2019multiclass} always assumes getting responses to four queries that essentially correspond to the four intervals. In practice, the binary search can be adaptive and may only require at most three queries as we have discussed in this paper. The order of the queries in LPME, however, remains the same.} Note that, depending on the oracle responses, one may reduce the search interval to half using less than three queries in some cases. Then the algorithm cyclically updates each angle until it converges to a metric sufficiently close to the true metric. We fix the number of cycles in coordinate-wise binary search to three. Therefore, in order to elicit a linear performance metric in $k$ dimensions, the LPME subroutine requires at most $3 \times 3 \times k \log(\pi/2\epsilon) $ queries, where three is the number of cycles in coordinate wise binary search, three is the (maximum) number of queries to shrink the search interval into half, and the initial search interval for the angles is $\pi/2$.

\section{Geometry Of The Feasible Space (Proofs of Section~\ref{sec:background} and Section~\ref{sec:fairme})}
\label{append:sec:confusion}
\vskip -0.3cm

\bproof[Proof of Proposition~\ref{prop:C} and Proposition~\ref{prop:f-C}]

We prove Proposition~\ref{prop:f-C}. The proof of Proposition~\ref{prop:C} is analogous where the probability measures (corresponding to classifiers and their rates) are not conditioned on any group. 

The group-specific set of rates $\Rcal^g$ for a group $g$ has the following properties~\citep{hiranandani2020fair}:
\vspace{-0.2cm}
\bitemize[leftmargin=1em]
\item \emph{Convex}: Consider two classifiers $h_1^g, h_2^g \in \Hcal^g$ that achieve the rates $\rmbf_1^g, \rmbf_2^g \in \Rcal^g$, respectively. 
Also, consider a classifier $h^g$ that predicts what classifier $h_1^g$  predicts with probability $\gamma$ and predicts what classifier $h_2^g$ predicts with probability $1-\gamma$. Then the rate vector of the classifier $h^g$ is: 

\vspace{-0.2cm}
\begin{align*}
R_{ij}^g(h) &= \Pmbb(h^g=i | Y=i)  \\ \nonumber
&= \Pmbb(h^g_1=i|h^g=h^g_1, Y=i)\Pmbb(h^g=h_1^g) + \Pmbb( h_2^g=i|h^g=h_2^g, Y=i)\Pmbb(h^g=h^g_2)   \\ \nonumber
&= \gamma \rmbf_{1}^g + (1-\gamma)\rmbf_{2}^g.
\end{align*}
The above equations show that the convex combination of any two rates is feasible as well, i.e., one can construct a randomized classifier which will achieve the convex combination of rates. Hence, $\Rcal^g \; \forall \; g \in [m]$ is convex. Since intersection of convex sets is convex, the intersection set $\Rcal^1\cap \dots \cap \Rcal^m$ is convex as well.  
\item \emph{Bounded:} Since $R^g_{ij}(h) = P[h=i|Y=i] \leq 1$ for all $i\in [k]$, $\Rcal^g \subseteq [0, 1]^k$.
\item \emph{The rates $\ombf$ and  $\embf_i$'s are always achievable:} A uniform random classifier, i.e, the classifier, which for any input, predicts all classes with probability $1/k$ achieves the rate profile $\ombf$. A classifier that always predicts class $i$ achieves the rate $\embf_i$. Thus, $\embf_i \in \Rcal^g \, \forall\, i \in [k], g \in [m]$ are always feasible. 
\item \emph{$\embf_i$'s are vertices:} Consider the supporting  hyperplanes with the following slope: $a_{i} > a_{j} > 0$ and $a_{l}=0$ for $l \in [k], l \neq i, j$. These hyperplanes will be supported by $\embf_i$. Thus, $\embf_i$'s  are vertices of the convex set $\Rcal^g$. From Assumption~\ref{as:sphere}, one can construct a ball around the trivial rate $\ombf$ and thus $\ombf$ lies in the interior.
\eitemize
The above points apply to space of overall rates $\Rcal$ as well; thus, proving Proposition~\ref{prop:C}. 
\eproof

\subsection{Finding the Sphere $\Scal\subset \Rcal$}
\label{append:ssec:sphere}

In this section, we provide details regarding how a sphere $\Scal$ with sufficiently large radius $\rho$ inside the feasible region $\Rcal$ may be found (see Figure~\ref{fig:geometry}(a)). The following discussion is borrowed from~\citep{hiranandani2019multiclass} and provided here for completeness. 
\balgorithm[t]
\caption{Obtaining the sphere $\Scal \subset \Rcal$ (Figure~\ref{fig:geometry}(a)) of radius $\rho$ centered at $\ombf$}
\label{alg:sphere}
\small
\balgorithmic[1]
\FOR{$j=1, 2, \cdots, k$}
\STATE Let $\mathbf \alphambf_j$ be the standard basis vector. 
\STATE Compute the maximum constant $c_j$ such that $\ombf + c_j \mathbf \alphambf_j$ is feasible by solving~\eqref{eq:op1}.
\ENDFOR
\STATE Let $CONV$ denote the convex hull of $\{\ombf\pm c_j\mathbf \alphambf_j\}_{j=1}^{k}$. It will be centered at $\ombf$.
\STATE Compute the radius $\rho$ of the largest ball that fits in $CONV$.
\STATE\textbf{Output:} Sphere $\Scal$ with radius $\rho$ centered at $\ombf$.
\ealgorithmic
\ealgorithm

The following optimization problem is a special case of OP2 in~\citep{narasimhan2018learning}. The problem is associated with a feasibility check problem.  Given a rate profile $\rmbf_0$, the optimization routine tries to construct a classifier that achieves the rate $\rmbf_0$ within small error $\epsilon >0$. 

\begin{align}
    \min_{\rmbf \in \Rcal} \; 0 \qquad s.t. \;\; \Vert \rmbf - \rmbf_0 \Vert_2 \leq \epsilon.
     \tag{OP1}
    \label{eq:op1}
\end{align}

The above optimization problem checks the feasibility, and if there exists a solution to the above problem, then Algorithm~1 of~\citep{narasimhan2018learning} returns it. 
Furthermore, Algorithm~\ref{alg:sphere} computes a value of $\rho\geq \tilde{p}/k$, where $\tilde{p}$ is the radius of the largest ball contained in the set $\Rcal$. Also, the approach in~\citep{narasimhan2018learning} is consistent, thus we should get a good estimate of the sphere, provided we have sufficiently large number of samples. The algorithm is completely offline and does not impact oracle query complexity.

\blemma\citep{hiranandani2019multiclass}
    Let $\tilde{p}$ denote the radius of the largest ball in $\Rcal$ centered at $\ombf$. Then Algorithm~\ref{alg:sphere} returns a sphere with radius $\rho\geq \tilde{p}/k$, where $k$ is the number of classes. 
\elemma

The idea in Algorithm~\ref{alg:sphere} can be trivially extended to finding a sphere $\Sbar \subset \Rcal^1\cap\dots\cap\Rcal^m$ corresponding to Remark~\ref{as:f-sphere}.

\section{{Quadratic Performance Metric Elicitation Procedure}}
\label{append:sec:qpme}

In this section, we describe how the subroutine calls to LPME in Algorithm~1 elicit a quadratic metric in Definition~\ref{def:quadmet}. We start with the shifted metric  of Equation~\eqref{eq:loclinapx}. Also, as explained in the main paper, we may assume $d_1 \neq 0$ due to Assumption~\ref{assump:smoothness}. We can derive the following solution using any non-zero coordinate of $\dmbf$, instead of $d_1$. We can identify a non-zero coordinate using $k$ trivial queries of the form $(\varrho\alphambf_i + \ombf, \ombf), \forall i \in [k]$. 

\begin{enumerate}
    \item From line 1 of Algorithm~1, we get local linear approximation at $\ombf$. Using Remark~\ref{rm:ratio}, we have~\eqref{eq:0col} which is
    \begin{equation}
    d_i = \frac{f_{i0}}{f_{10}}d_1 \qquad \forall \; i \in \{2, \dots, k\}.
    \label{append:eq:0col}
\end{equation}
\item Similarly, if we apply LPME on small balls around rate profiles $\zmbf_j$, Remark~\ref{rm:ratio} gives us:
\begin{equation}
\frac{d_i + (\rho-\varrho)B_{ij}}{d_1 + (\rho-\varrho)B_{1j}} = \frac{f_{ij}}{f_{1j}} \quad \forall \; i \in \{2, \ldots, k\},\; j \leq i.
\label{append:eq:jcol}
\end{equation}

\begin{align*}
    &\implies d_i + (\rho-\varrho)B_{ij} = \frac{f_{ij}}{f_{1j}}(d_1 + (\rho-\varrho)B_{1j})\\
    &\implies (\rho-\varrho)B_{ij} = \frac{f_{ij}}{f_{1j}}(d_1 + (\rho-\varrho)B_{j1}) - d_i \\
    &\implies (\rho-\varrho)B_{ij} = \frac{f_{ij}}{f_{1j}}(d_1 +  \frac{f_{j1}}{f_{11}} (d_1 + (\rho - \varrho)B_{11}) - d_j ) - \frac{f_{i0}}{f_{10}}d_1\\
    &\implies (\rho-\varrho)B_{ij} = \left(\frac{f_{ij}}{f_{1j}} - \frac{f_{i0}}{f_{10}} + \frac{f_{ij}}{f_{1j}} \left(\frac{f_{j1}}{f_{11}} - \frac{f_{j0}}{f_{10}}\right)\right)  d_1 + (\rho-\varrho)  \frac{f_{j1}}{f_{11}}B_{11}, \numberthis \label{append:eq:solvemidssystem}
\end{align*}
where we have used that the matrix $\Bmbf$ is symmetric in the second step, and~\eqref{append:eq:0col} in the last two steps. We can represent each element in terms of $B_{11}$ and $d_1$. So, a relation between $B_{11}$ and $d_1$ may allow us to represent each element of $\ambf$ and $\Bmbf$ in terms of $d_1$.

\item Therefore, by applying LPME on small balls around rate profiles $-\zmbf_1$, Remark~\ref{rm:ratio} gives us~\eqref{eq:negativegrad}:

\begin{equation}
    \frac{d_2-(\rho - \varrho)B_{21}}{d_1-(\rho - \varrho)B_{11}} = \frac{f_{21}^-}{f_{11}^-}.
    \label{append:eq:negativegrad}
\end{equation}

\item Using~\eqref{append:eq:jcol} and~\eqref{append:eq:negativegrad}, we have:

\begin{align*}
    (\rho - \varrho)B_{11} = \frac{ \frac{f_{21}^-}{f_{11}^-} + \frac{f_{21}}{f_{11}} - 2\frac{f_{20}}{f_{10}}  }{ \frac{f_{21}^-}{f_{11}^-} - \frac{f_{21}}{f_{11}} }d_{1}.
    \numberthis \label{append:eq:firsttermB}
\end{align*}
Putting~\eqref{append:eq:firsttermB} in~\eqref{append:eq:solvemidssystem}, we get:
\begin{align*}
    B_{ij} &=  \left[\frac{f_{ij}}{f_{1j}}\left(1 + \frac{f_{j1}}{f_{11}} \right) - \frac{f_{ij}}{f_{1j}}\frac{f_{j0}}{f_{10}} - \frac{f_{i0}}{f_{10}} +  \frac{f_{ij}}{f_{1j}}\frac{f_{j1}}{f_{11}} \frac{ \frac{f_{21}^-}{f_{11}^-} + \frac{f_{21}}{f_{11}} - 2\frac{f_{20}}{f_{10}}  }{ \frac{f_{21}^-}{f_{11}^-} - \frac{f_{21}}{f_{11}}  }\right]d_1 \\
    &= \left(F_{i,1,j} (1 + F_{j,1,1}) - F_{i,1,j} F_{j,1,0}  - F_{i,1,0} + F_{i,1,j}\frac{F^-_{2,1,1} + F_{2,1,1} - 2F_{2,1,0}}{F^-_{2,1,1} - F_{2,1,1}}\right)d_1,
    \numberthis \label{append:eq:poly2elicitamatfinal}
\end{align*}
where
$F_{i,j,l} = \frac{f_{il}}{f_{jl}}$ and $F^-_{i,j,l} = \frac{f^-_{il}}{f^-_{jl}}$. As $\ambf = \dmbf + \Bmbf \ombf$, we can represent each element of $\ambf$ and $\Bmbf$ using~using~\eqref{append:eq:0col}  and \eqref{append:eq:poly2elicitamatfinal} in terms of $d_1$. We can then use the normalization condition $\Vert \ambf\Vert_2^2 + \Vert \Bmbf \Vert_F^2 = 1$ to get estimates of $\ambf, \Bmbf$ which are independent of $d_1$. 
\end{enumerate}

This completes the derivation of solution from QPME (section~\ref{sec:quadme}).

\section{{Fair (Quadratic) Performance Metric Elicitation Procedure}}
\label{append:sec:fpme}

\begin{figure}[H]
\centering
\fbox{\parbox[t]{0.60\textwidth}{\small{\underline{\bf Algorithm~4: FPM Elicitation}\normalsize}    \\
\textbf{Input:} Query set $\Scal'$, search tolerance $\epsilon > 0$, oracle $\Omega'$ \\
1. \text{ \ }Let $\Lcal \leftarrow \varnothing$ \\
2: \text{ \ }\textbf{For} \, $\sigma \in \Mcal$ \textbf{do}\\
3: \text{ \ \ \ } $\bm{\beta}^{\sigma}\leftarrow$ QPME$(\Scal', \epsilon, \Omega')$\\
4: \text{ \ \ \ } Let $\ell^\sigma$ be Eq.~\eqref{append:eq:fairBij}, extend $\Lcal \leftarrow \Lcal \cup \{\ell^\sigma\}$\\
5:  \text{ \ }$\hat{\Bmbb} \leftarrow $ normalized solution from~\eqref{append:eq:fairbsol} using $\Lcal$\\
6: \text{ \ }$\hat \lambda \leftarrow$ trace back normalized solution from~\eqref{append:eq:fairBij} for any $\sigma$\\
\textbf{Output:} $\ambfhat, \hat{\Bmbb}, \hat \lambda$ 
\normalsize \vspace{-0.25em}
}}
\label{alg:f-linear}
\end{figure}

We first discuss eliciting the fair (quadratic) metric in Definition~\ref{def:f-linmetric}, where all the parameters are unknown. We then provide an alternate procedure for eliciting just the trade-off parameter $\lambda$ when the predictive performance and fairness violation coefficients are known. The latter is a separate application as discussed in~\citep{zhang2020joint}. However, unlike~\cite{zhang2020joint}, instead of ratio queries, we use simpler pairwise comparison queries.

In this section, we work with any number of groups $m\geq 2$. The idea, however, remains the same as described in the main paper for number of groups $m=2$. We specifically select queries from the sphere $\overline{\Scal} \subset \Rcal^1 \cap \dots \cap\Rcal^m$, which is common to all the group-specific feasible region of rates, so to reduce the problem into multiple instances of the proposed QPME procedure of Section~\ref{sec:quadme}. 

Suppose that the oracle's fair performance metric is $\phi^{\text{fair}}$ parametrized by $(\ambf, \Bmbb, \lambda)$  as in Definition~\ref{def:f-linmetric}. The overall fair metric elicitation procedure framework is summarized in Algorithm~4. The framework exploits the sphere $\overline{\Scal} \subset \Rcal^1 \cap \dots\cap\Rcal^m$ and uses the QPME procedure (Algorithm~1) as a subroutine multiple times. 

Let us consider a non-empty set of sets $\Mcal \subset 2^{[m]} \setminus \{\varnothing, [m]\}$. We will later discuss how to choose such a set $\Mcal$. 
We partition the set of groups $[m]$ into two sets of groups. Let $\sigma \in \Mcal$ and $[m] \setminus \sigma$ be one such partition of the $m$ groups defined by the set of groups $\sigma$. For example, when $m=3$, one may choose the set of groups $\sigma = \{1, 2\}$. 

Now, consider a sphere $\Scal'$ whose elements $\rmbf^{1:m} \in \Scal'$ are given by:
\vspace{-0.15cm}
\begin{equation}
    \rmbf^g = \begin{cases}
    \smbf & \text{if } g \in \sigma\\
    \ombf & \text{o.w. }
    \end{cases}
\label{eq:parvarphi}
\end{equation}
\vskip -0.25cm
This is an extension of the sphere $\Scal'$ defined in the main paper for the $m>2$ case. Elements in $\Scal'$ have rate profiles $\smbf \in \overline{\Scal}$ to the groups in $\sigma$ and trivial rate profile $\ombf$ to the remaining groups in $[m] \setminus \sigma$. 
Analogously, the modified oracle is $\Omega'(\rmbf_1, \rmbf_2) = \Omega((\rmbf^{1:m}_1), (\rmbf^{1:m}_2))$, where $\rmbf^{1:m}_1, \rmbf^{1:m}_2$ are the elements of the spheres $\Scal'$ above. 
Thus, for elements in $\Scal'$, the metric in Definition~\ref{def:f-linmetric} reduces to:

\begin{align*}
\phi^{\text{fair}}(\rmbf^{1:m} \in \Scal' \,;\, \ambf, \Bmbb, \lambda) =  
(1-\lambda)\inner{-\ambf \odot \taumbf^\sigma}{\smbf - \ombf} + \lambda \frac{1}{2} (\smbf - \ombf)^T\Wmbf^\sigma(\smbf - \ombf) + c^\sigma 
\numberthis \label{eq:metricbrich}
\end{align*}

where $\taumbf^\sigma = \sum_{g\in \sigma}\taumbf^g$, $\Wmbf^\sigma = \sum_{u \in \sigma, v \in [m]\setminus\sigma} B^{uv}$, and $c^\sigma$ is a constant not affecting the oracle responses.

The above metric is a particular instance of $\bphi(\smbf; \dmbf, \Bmbf)$ in~\eqref{eq:quadmetshift} with $\dmbf \coloneqq -(1-\lambda)\ambf\odot\taumbf^\sigma$ and $\Bmbf \coloneqq \lambda \Wmbf^\sigma$; thus, we apply QPME procedure as a subroutine in  Algorithm~4 to elicit the metric in~\eqref{eq:metricbrich}. 

The only change needed to be made to the algorithm is in line 5, where 
we need to take into account the changed relationship between $\dmbf$ and $\ambf$, and need to separately (not jointly) normalize the linear and quadratic coefficients. With this change, the output of the algorithm directly gives us the required estimates. 
Specifically, we have from step 1 of Algorithm~1 and \eqref{eq:0col} 
an estimate 
\begin{equation}
 \frac{{d}_{i}}{{d}_{1}} = \frac{\tau^\sigma_{i} {a}_i}{\tau^\sigma_{1} {a}_1} = \frac{f_{i0}}{f_{10}} \implies    {a}_i = \frac{f_{i0}}{f_{10}} \frac{\tau^\sigma_{1}}{\tau^\sigma_{i}} {a}_1.
 \label{append:eq:fair0col}
\end{equation}

Using the normalization condition (i.e., $\Vert \ambf \Vert_2 = 1$), we directly get an estimate $\ambfhat$ for the linear coefficients. Similarly, steps 2-4 of Algorithm~1 and \eqref{eq:poly2elicitamatfinal} gives us:$\hat{B}_{ij} = $
\begin{align*}
    \sum_{u \in \sigma, v \in [m]\setminus\sigma} \tilde B^{uv}_{ij} &= \Big(F_{i,1,j}^\sigma (1 + F_{j,1,1}^\sigma) - F_{i,1,j}^\sigma F_{j,1,0}^\sigma d_{1}
    - F_{i,1,0}^\sigma + F_{i,1,j}^\sigma\textstyle\frac{F^{-, \sigma}_{2,1,1} + F_{2,1,1}^\sigma - 2F_{2,1,0}^\sigma}{F^{-, \sigma}_{2,1,1} - F_{2,1,1}^\sigma}\Big)\tau^1_1\hat{a}_1 \\
    &= \beta^\sigma,  \numberthis \label{append:eq:fairBij}
\end{align*}
where the above solution is similar to the two group case, but here it is corresponding to a partition of groups defined by $\sigma$, and $\tilde \Bmbf^{uv} \coloneqq \lambda\Bmbf^{uv}/(1 - \lambda)$ is a scaled version of the true (unknown) $\Bmbf^{uv}$. Let equation~\eqref{append:eq:fairBij} be denoted by $\ell^\sigma$. Also, let the right hand side term of~\eqref{append:eq:fairBij} be denoted by $\beta^\sigma$. 

Since we want to elicit $m\choose 2$ fairness violation weight matrices in $\Bmbb$, we require $m\choose 2$ ways of partitioning the groups into 
two sets so that we construct $m\choose 2$ independent matrix equations similar to~\eqref{append:eq:fairBij}. 
Let $\Mcal$ be those set of sets. 
Thus, running over all the choices of sets of groups $\sigma \in \Mcal$ provides the system of equations $\Lcal \coloneqq \cup_{\sigma \in \Mcal} \ell^\sigma$ (line 4 in Algorithm~4), which is:
\begin{equation}
    \left[ \begin{array}{cccc} \Xi & 0 & \dots & 0\\
    0 & \Xi & \dots & 0 \\
    \dots & \dots & \dots & \dots \\
    0 & 0 & \dots & \Xi 
    \end{array}\right] \left[ \begin{array}{c} \tilde \bmbf_{(11)} \\
    \tilde \bmbf_{(12)} \\
    \dots \\
    \tilde \bmbf_{(kk)}
    \end{array}\right] = \left[ \begin{array}{c} \bm\beta_{(11)} \\
    \bm\beta_{(12)} \\
    \dots \\
    \bm\beta_{(kk)}
    \end{array}\right],
    \label{append:btilde}
\end{equation}
where $\tilde \bmbf_{(ij)} = (\tilde b_{ij}^1,\tilde b_{ij}^2, \cdots, \tilde b_{ij}^{m\choose 2})$ and $\gammambf_{(ij)} = (\beta_{ij}^1, \beta_{ij}^2, \cdots, \beta_{ij}^{m\choose 2})$ are vectorized versions of the $ij$-th entry across groups for $i, j \in [k]$, and $\Xi \in \{0,1\}^{{m\choose 2}\times {m\choose 2}}$ is a binary full-rank matrix denoting membership of groups in the set $\sigma$. For example, when one chooses $\Mcal = \{ \{1,2\}, \{1,3\}, \{2,3\}\}$ for $m=3$, $\Xi$ is given by:
$$\Xi = \left[ \begin{array}{ccc} 0 & 1 & 1\\
    1 & 0 & 1\\
    1 & 1 & 0\\
\end{array}\right].$$
One may choose any set of sets $\Mcal$ that allows the resulting group membership matrix $\Xi$ to be  non-singular. The solution of the system of equations $\Lcal$ is:
\begin{equation}
    \left[ \begin{array}{c} \tilde \bmbf_{(11)} \\
    \tilde \bmbf_{(12)} \\
    \dots \\
    \tilde \bmbf_{(kk)}
    \end{array}\right] = \left[ \begin{array}{cccc} \Xi & 0 & \dots & 0\\
    0 & \Xi & \dots & 0 \\
    \dots & \dots & \dots & \dots \\
    0 & 0 & \dots & \Xi
    \end{array}\right]^{(-1)} \left[ \begin{array}{c} \bm\beta_{(11)} \\
    \bm\beta_{(12)} \\
    \dots \\
    \bm\beta_{(kk)}
    \end{array}\right].
    \label{append:eq:sol-b}
\end{equation}
When all $\tilde \Bmbf^{uv}$'s are normalized, we have the estimated fairness violation weight matrices as:
\begin{equation}
\Bmbfhat^{uv} = \frac{\tilde \Bmbf^{uv}}{\frac{1}{2}\sum_{u,v=1, v > u}^m \Vert \tilde \Bmbf^{uv} \Vert_F} \quad \text{for} \quad u,v \in [m], v>u.
 \label{append:eq:fairbsol}
\end{equation} 
Due to the above normalization, the solution is again independent of the true trade-off $\lambda$.

Given estimates $\hat{B}^{uv}_{ij}$ and $\ahat_1$,  we can now additionally estimate the trade-off parameter $\hat{\lambda}$ from  $\ell^\sigma$~\eqref{append:eq:fairBij} for any $\sigma \in \Mcal$. This completes the fair (quadratic) metric elicitation procedure. 

\subsection{Eliciting Trade-off $\lambda$ when (linear) predictive performance and (quadratic) fairness violation coefficients are known}
\label{append:ssec:lambda}

We  now provide an alternate binary search based method similar to~\cite{hiranandani2020fair} for eliciting the trade-off parameter $\lambda$ when the linear predictive and quadratic fairness coefficients are already known. This is along similar lines to the application considered by~\cite{zhang2020joint}, but unlike them, instead of ratio queries, we require simpler pairwise queries. 

Here, the key insight is to approximate the non-linearity posed by the fairness violation in Definition~\ref{def:f-linmetric}, which then reduces the problem to a  one-dimensional binary search. We have:
\begin{align*}
&\phi^{\text{fair}}(\tupr \,;\, \ambf, \Bmbb, \lambda) \,\coloneqq\,  (1-\lambda)\inner{\ambf}{\bm{1} - \rmbf} + \lambda \frac{1}{2} \left(\sum\nolimits_{u,v=1,v>u}^{m} (\rmbf^u - \rmbf^v)^T\mathbbm{\Bmbf}^{uv}(\rmbf^{u} - \rmbf^v)\right). \numberthis \label{append:eq:fairmetshifted}
\end{align*}
To this end, we define a new sphere $\Scal' = \{ (\smbf,\ombf, \dots, \ombf)  | \smbf \in \overline{\Scal}\}$. The elements in $\Scal'$ is the set of rate profiles whose first group achieves rates $\smbf \in \overline{\Scal}$ and rest of the groups achieve trivial rate $\ombf$ (corresponding to uniform random classifier). For any element in $\Scal'$, the associated discrepancy terms $(\rmbf^u - \rmbf^v) = 0$ for $u,v \neq 1$. 
Thus for elements in $\Scal'$, the metric in Definition~\ref{def:f-linmetric} reduces to:
\vspace{-0.2cm}
\begin{align*}
        \phi^{\text{fair}}((\smbf, \ombf, \dots, \ombf) \,;\,\ambf, \Bmbb, \lambda) =& (1-\lambda)\inner{-\taumbf^1\odot\ambf}{\smbf - \ombf} + 
        \lambda\frac{1}{2} (\smbf - \ombf)^T\sum_{v=2}^m \Bmbf^{1v} (\smbf - \ombf) + c.
         \numberthis \label{append:eq:metriclambda}
\end{align*}
\vskip -0.3cm
Additionally, we consider a small sphere $\overline{\Scal}'_{\zmbf_1}$, where $\zmbf_1 \coloneqq (\rho - \varrho)\bm{\alpha}_1 + \ombf$, similar to what is shown in Figure~\ref{fig:geometry}(a). We may approximate the quadratic term on the right hand side above by its first order Taylor approximation as follows:
\begin{align*}
        \phi^{\text{fair}}( (\smbf, \ombf, \dots, \ombf) ;\ambf, \Bmbb, \lambda) &\approx  \phi^{\text{fair, apx}}( (\smbf, \ombf, \dots, \ombf) ;\ambf, \Bmbb, \lambda) \\ &= \inner{-(1-\lambda)\taumbf^1\odot\ambf + \lambda \sum_{v=2}^m \Bmbf^{1v}(\zmbf_1 - \ombf)}{\smbf}
         \numberthis \label{eq:metriclambdalinear}
\end{align*}
for $\smbf$ in a small neighbourhood around the rate profile $\zmbf_1$. Since the metric is essentially linear in $\smbf$, the following lemma from~\citep{hiranandani2020fair} shows that the metric in~\eqref{eq:metriclambdalinear} is quasiconcave in $\lambda$. 

\vspace{-0.1cm}
\blemma
Under the regularity assumption that $\inner{-\taumbf^1\odot\ambf}{\sum_{v=2}^m \Bmbf^{1v}(\zmbf_1 - \ombf)}\neq 1$, the function
\begin{equation}
\vartheta(\lambda) \coloneqq \max_{\smbf \in \overline{\Scal}'_{\zmbf_1}} \phi^{\text{fair, apx}}( (\smbf, \ombf, \dots, \ombf) ;\ambf, \Bmbb, \lambda)
\label{append:eq:vartheta}
\end{equation}
is strictly quasiconcave (and therefore unimodal) in $\lambda$.
\label{lm:lambda}
\elemma
\vskip -0.2cm
The unimodality of $\vartheta(\lambda)$ allows us to perform the one-dimensional binary search in Algorithm~\ref{alg:lambda} using the query space $\overline{\Scal}'_{\zmbf_1}$, tolerance $\epsilon$, and the oracle $\Omega$. The binary search algorithm is same as Algorithm~4 in~\citep{hiranandani2020fair} and provided here for completeness. 

\addtocounter{algorithm}{1}
\balgorithm[t]
\caption{Eliciting the trade-off $\lambda$ when predictive performance and fairness violation are known}
\label{alg:lambda}
\small
\balgorithmic[1]
\STATE \textbf{Input:} Query space $\overline{\Scal}'_{\zmbf_1}$, binary-search tolerance $\epsilon > 0$, oracle $\Omega$
\STATE \textbf{Initialize:} $\lambda^{(a)} = 0$, $\lambda^{(b)} = 1$.
\WHILE{$\abs{\lambda^{(b)} - \lambda^{(a)}} > \epsilon$} 
\STATE Set $\lambda^{(c)} = \frac{3 \lambda^{(a)} + \lambda^{(b)}}{4}$, $\lambda^{(d)} = \frac{\lambda^{(a)} + \lambda^{(b)}}{2}$, $\lambda^{(e)} = \frac{\lambda^{(a)} + 3 \lambda^{(b)}}{4}$
\STATE Set $\smbf^{(a)} = \displaystyle\argmax_{\smbf \in\overline{\Scal}'_{\zmbf_1}} \inner{-(1-\lambda^{(a)})\taumbf^1\odot\ambfhat + \lambda^{(a)} \sum_{v=2}^m \Bmbfhat^{1v}(\zmbf_1 - \ombf)}{\smbf}$ using Lemma~\ref{lem:spherebayes}
\STATE Similarly, set $\smbf^{(c)}$, $\smbf^{(d)}$, $\smbf^{(e)}$, $\smbf^{(b)}$.
\STATE $[\lambda^{(a)}, \lambda^{(b)}] \leftarrow$ \emph{ShrinkInterval} $(\Omega, \smbf^{(a)}),\smbf^{(c)}),\smbf^{(d)}),\smbf^{(e)}),\smbf^{(b)}) )$ using a subroutine analogous to the routine shown in  Figure~\ref{append:fig:shrink1}.
\ENDWHILE
\STATE \textbf{Output:} $\hat\lambda = \frac{\lambda^{(a)}+\lambda^{(b)}}{2}$. 
\ealgorithmic
\ealgorithm

\section{Extension to Eliciting General Quadratic Metrics}
\label{append:generalquad}

In this section, we discuss how the entire setup including the proposed procedure and the guarantees of the main paper described in terms of the \emph{diagonal} entries of the predictive rate matrix extends to a setup where the metric is defined in all the terms of the rate matrix. For this section, we use an additional notation. For a matrix $\Ambf$, let $\offdiag(\Ambf)$ returns a vector of off-diagonal elements of $\Ambf$.  

Just like the main paper, we consider a $k$-class classification setting with $X \in \Xcal$ and $Y \in [k]$ denoting the input and output random variables, respectively. We assume access to an $n$-sized sample $\{(\xmbf, y)_i\}_{i=1}^n$ generated \emph{iid} from a distribution $ \Pmbb(X, Y)$. We work with randomized classifiers $h : \Xcal \rightarrow \Delta_k$ that for any $\xmbf$ gives a distribution $h(\xmbf)$ over the $k$ classes and use 
 $\Hcal = \{h : \Xcal \rightarrow \Delta_k\}$
 to denote the set of all classifiers. 

\emph{General Predictive rates:} 
We define the predictive rate matrix for a classifier $h$ by $\Rmbf(h, \Pmbb) \in \Rmbb^{k \times k}$, where the $ij$-th entry is the fraction of label-$i$ examples for which the randomized classifier $h$ predicts $j$:
\vspace{0.15cm}
\begin{align}
	R_{ij}(h, \Pmbb) \coloneqq \Pmbb(h(X) = j | Y = i)  \quad \text{for} \; i, j \in [k],
	\label{eq:generalcomponents}
\end{align}
\vskip 0.15cm
where the probability is over draw of $(X, Y) \sim \P$ and the randomness in $h$. 

Notice that each diagonal entry
of $\Rmbf$
can be written in terms of its off-diagonal elements: 
\vspace{0.15cm}
    $${R_{ii}(h, \Pmbb) = 1 - \sum\nolimits_{j=1,j\neq i}^k R_{ij}(h, \Pmbb).}$$
\vskip 0.15cm
Thus, we can represent a rate matrix with its $q \coloneqq (k^2 - k)$ off-diagonal elements, write it as a vector $\rmbf(h, \Pmbb) = \offdiag(\Rmbf(h, \Pmbb))$, and interchangeably refer to it as the \emph{`vector of general rates'} or \emph{`off-diagonal rates'}. To distinguish from rates considered in the main paper, we will call the rates entries corresponding to the diagonals of the rate matrix, i.e., $\Pmbb(h(X)=i|Y=i)$ as discussed in Equation\eqref{eq:components}, as the \emph{`diagonal rates'}. 

\emph{Feasible general rates:} The set of all feasible general rates is given by: \vspace{0.15cm}
$$\Rcal = \{\rmbf(h, \Pmbb) \in [0,1]^q\,:\, h \in \Hcal \}.$$ 
\vspace{-0.15cm}

The quadratic metric in general rates is defined in the same way as Definition~\ref{def:quadmet} as follows:

\bdefinition[Quadratic Metric in General Rates] For a vector $ \ambf \in \Rmbb^q$ and a positive semi-definite symmetric matrix $\Bmbf \in \Rmbb^{q \times q}$ with $\Vert \ambf \Vert_2^2  + \Vert \Bmbf \Vert_F^2 = 1$ (w.l.o.g.\ due to scale invariance):
\begin{equation}
    \phi^\quadr(\rmbf \,;\, \ambf, \Bmbf) = \inner{\ambf}{\rmbf} + \frac{1}{2} \rmbf^T \Bmbf \rmbf.
    \label{eq:generalquadmet}
\end{equation}
\vspace{-0.4cm}
\label{def:generalquadmet}
\edefinition

\bexample[Distribution matching]
\emph{
We can extend Example~\ref{ex:distmatchbin} in the multiclass case as follows. In certain applications, one needs the proportion of predictions 
for each class (i.e., the coverage) to match a target distribution $\boldsymbol{\pi} \in \Delta_k$ 
\citep{goh2016satisfying,narasimhan2018learning}. 
A 
measure often used for this task is the squared difference between the per-class coverage and the target distribution: 
{\small$\phi^{\cov}(\rmbf) \,=\, \sum_{i=1}^k \left(\cov_i(\rmbf) - \pi_i\right)^2$}, where 
{\small$\cov_i(\rmbf) = 1 - \sum_{j=1}^{k-1}r_{(i-1)(k-1) + j} + \sum_{j> i}r_{(j-1)(k-1)+i}+ \sum_{j<i}r_{(j-1)(k-1)+i-1}$}. 
Similar metrics can be found in the quantification literature where the target is set to the class prior $\Pmbb(Y=i)$ \citep{Fab1, 
Kar16}. 
We capture more general quadratic distance measures for distributions, e.g.\ {\small$(\bf{\cov}(\rmbf) - \boldsymbol{\pi})^{T}\Qmbf (\bf{\cov}(\rmbf)-\boldsymbol{\pi})$} for a positive semi-definite matrix $\Qmbf \in PSD_k$ \citep{Lindsay08}.
}
\eexample

The definition of metric elicitation and oracle query remain the same except that the vector $\rmbf$ now represents the vector of general rates, and not just the diagonal rates.

Consider Appendix~\ref{append:sec:confusion}, where we discuss proof of Proposition~\ref{prop:C}, Proposition~\ref{prop:f-C}, and a procedure to construct a feasible sphere of appropriate radius in the convex set of diagonal rates. The entire methodology applies to general set of rates by \emph{just} replacing diagonal rates in the proofs with the general rates. Thus, all the geometrical properties discussed in Proposition~\ref{prop:C} for the set of diagonal rates applies to the set of general rates.
The exact geometry of the set of diagonal rates, as shown in Figures~\ref{fig:geometry}(a) and \ref{fig:geometry}(b)may differ from the geometry of the set of general rates; however, the geometric properties including $\embf_i$ being the vertices remains the same. 
Therefore, under the same Assumption~\ref{assump:distribution}, we can guarantee an existence of a sphere in the set of general rates similar to Remark~\ref{as:sphere}. 

Once we guarantee a sphere in the set of general rates, we can follow LPME for eliciting linear metrics in general rates or QPME for eliciting linear metrics in general rates. The computational and query complexity will depend on the number of unknowns, which in the case of general rates, will be $\tilde O(q)$ for LPME and $\tilde O(q^2)$ for QPME.

\section{Practicality of Querying Oracle}
\label{append:practicality}
\vskip -0.25cm

Recall that in our setup, any query posed to the oracle needs to feasible, i.e., should be achievable by some classifier (see definition of feasible rates in Section~\ref{sec:background}). Therefore the oracle we query can be a human expert or a group of experts who compare intuitive visualizations of  rates, or can be an entire  population of users (as would be the case with A/B testing). 

An important practical concern in employing the proposed QPME procedure is the number of queries needed to be posed to the oracle. We  note that (i) the number of queries needed by our proposal is optimal (i.e.\ matches the lower bound for the problem in Theorem \ref{thm:lb}), (ii) has only a \emph{linear} dependence on the number of unknowns (Theorem \ref{thm:q-me}), and (iii) can be considerably reduced by making reasonable practical structural assumptions about the metric to reduce the number of unknowns. While our procedure's query complexity for the most general setup (with $O(k^2)$ unknowns) is $\tilde{O}(k^2)$, the quadratic dependence on $k$ is merely an artifact of there being $O(k^2)$ unknowns in this setup. For example, when the number of classes is large, one may just cluster the classes from error perspective. For example, one may assume same error costs for similar classes. This will reduce the number of unknowns to $O(c^2)$, where $c<<k$ is the number of cluster of classes.

We also stress that in many internet-based settings, one can deploy A/B tests to obtain preferences by aggregating feedback from a large group of participants. In this case, the entire user population serves as an oracle. Most internet-based companies run thousands of A/B tests daily making it practical to get preferences for our metric elicitation procedure. Moreover, one can employ practical improvements such as running A/B tests with fewer participants in the initial rounds (when the rates are far apart) and switch to running A/B tests with more precision in later rounds. Note that because the queries posed by our method always corresponds to a feasible classifier (see Section~\ref{sec:quadme}), one can easily run comparisons between classifiers as a part of an A/B test. 

If needed, our algorithms can also work with queries that compare classification statistics directly, instead of classifiers. There has been growing work on visualization of confusion matrices (predictive rates) for non-expert users. For example, see~\citep{beauxis2014visualization} and~\citep{shen2020designing}. With the aid of such intuitive visualizations, it is reasonable to expect human practitioners to comprehend the queries posed to them and provide us with pairwise comparisons. Moreover, we have shown that our approach is resilient to noisy responses, which enhances our confidence in their ability to handle human feedback. In practice, the most viable option will depend on the target population. 

Finally, we would like to emphasize that because our query complexity  has only a linear dependence on the number of unknowns, and the number of unknowns can be reduced with practical structural assumptions, our proposal is as practical as the prior methods \citep{hiranandani2018eliciting, hiranandani2019multiclass} for linear metric elicitation. In fact, despite eliciting from a more flexible class, our proposal has the same dependence on the number of unknowns as those prior methods.

\section{Elicitation Guarantee for the QPME Procedure}
\label{append:sec:guarantees}
\vskip -0.2cm

We now discuss guarantees of the QPME procedure and proofs of the theoretical results. 

\subsection{Sample complexity bounds} 
\vskip -0.1cm
Recall from Definition~\ref{def:noise} that the oracle responds correctly as long as $|\phi(\rmbf_1) - \phi(\rmbf_2)| > \epsilon_\Omega$. For simplicity, we assume that our algorithm 
has  access to the population rates $\rmbf$ defined in Eq.~(1). 
In practice, we expect  to estimate the rates using a sample $D\coloneqq \{\xmbf, y\}_{i=1}^n$ drawn from the distribution $\Pmbb$, and to query classifiers from a hypothesis class $\mathcal{H}$ with finite capacity. Standard generalization bounds (e.g.~\cite{Daniely:2015}) give us that with high probability over draw of $D$, the estimates  $\hat{\rmbf}$ are close to the population rates $\rmbf$, up to the desired  tolerance $\epsilon_\Omega$, 
as long as we have sufficient samples. Further, since the metrics $\phi$ are Lipschitz w.r.t.\ rates, with high probability, 
we thus gather correct oracle feedback from querying with finite sample estimates $\Omega(\hat{\rmbf}_1, \hat{\rmbf}_2)$.

More formally, for $\delta \in (0,1)$, as long as the  sample size $n$ is greater than ${O\big(\sfrac{\log(|\mathcal{H}|/\delta)}{\epsilon_\Omega^2}\big)}$, the guarantee in Theorem 1 hold with probability at least $1 - \delta$ (over draw of $D$), where $|\mathcal{H}|$ can in turn be replaced by a measure of capacity of the hypothesis class $\mathcal{H}$. For example, one can show the following corollary to Theorem \ref{thm:q-me} for a
hypothesis class $\mathcal{H}$ in which each classifier is a randomized combination of a finite number of deterministic classifiers chosen from a set $\bar{\mathcal{H}}$, and whose capacity is measured in terms of the Natarajan dimension~\citep{Natarajan:1989} of $\bar{\mathcal{H}}$.
\begin{corollary}
Suppose the hypothesis class $\mathcal{H}$ of randomized classifiers used to choose queries to the oracle is  of the form:
$$\mathcal{H} =\bigg\{x \mapsto \sum_{t=1}^T\alpha_t h_t(x) \,\bigg|\, T \in \mathbb{Z}_+, \alpha \in \Delta_T, h_1, \ldots, h_T \in \bar{\mathcal{H}}\bigg\},$$ for some class $\bar{\mathcal{H}}$ of deterministic multiclass classifiers $h: \mathcal{X} \rightarrow \{0,1\}^k$. Suppose the deterministic hypothesis class $\bar{\mathcal{H}}$  has   Natarajan dimension $d > 0$, and $\phi$ is $1$-Lipschitz. Then for any $\delta \in (0,1)$,
as long as the  sample size $n 
\geq O\Big(\frac{d\log(k) + \log(1/\delta)}{\epsilon_\Omega^2}\Big)$, the guarantee in Theorem 1 hold with probability at least $1 - \delta$ (over draw of $D = \{\xmbf_i, y_i\}_{i=1}^n$ from $\Pmbb$).
\label{append:cor:finite}
\end{corollary}
The proof adapts generalization bounds from~\cite{Daniely:2015}, and uses the fact that the predictive rate for any randomized classifier in $\mathcal{H}$ is a convex combination of rates for deterministic classifiers in $\bar{\mathcal{H}}$ (due to linearity of expectation). 

\subsection{Proofs}
Before presenting the proof of Theorem \ref{thm:q-me}, we re-write the LPME guarantees from~\citep{hiranandani2019multiclass} for linear metrics in the presence of an oracle noise parameter $\epsilon_\Omega$ from Definition~\ref{def:noise}. 

\blemma[LPME guarantees with oracle noise~\citep{hiranandani2019multiclass}]
\label{lem:LPMEwnoise}
Let the oracle $\Omega$'s metric be $\phi^{\text{lin}} = \inner{\ambf}{\rmbf}$ and its feedback noise parameter from Definition~\ref{def:noise} be $\epsilon_\Omega$. Then, if the LPME procedure (Algorithm~\ref{alg:slme}) is run using a 
sphere $\Scal \subset \Rcal$ of radius $\varrho$ and the  binary-search tolerance $\epsilon$, then by posing $O(k\log(1/\epsilon))$
queries it recovers coefficients $\ambfhat$ with $\Vert \ambf - \ambfhat \Vert_2 \leq O\left(\sqrt{k}(\epsilon + \sqrt{\epsilon_\Omega/\varrho})\right)$.
\elemma

\bproof[Proof of Theorem~\ref{thm:q-me}] 

We first find the smoothness coefficient of the metric in Definition~\ref{def:quadmet}.

A function $\phi$ is said to be $L$-smooth if for some bounded constant $L$, we have:

$$
\Vert \nabla \phi(x) - \nabla \phi(y) \Vert_2 \leq L\Vert x - y \Vert_2.
$$

For the metric in Definition~\ref{def:quadmet}, we have:
\begin{align*}
\Vert \nabla \phi^{\text{quad}}(x) - \nabla \phi^{\text{quad}}(y) \Vert_2 &= \Vert \ambf + \Bmbf\xmbf - (\ambf + \Bmbf\ymbf) \Vert_2 \\
&\leq \Vert \Bmbf \Vert_2 \Vert x - y \Vert_2\\
&\leq \Vert \Bmbf \Vert_F \Vert x - y \Vert_2,\\
&\leq 1\cdot\Vert x - y \Vert_2,
\end{align*}
where in the last step, we have used the scale invariance condition from Definition~\ref{def:quadmet}, i.e., $\Vert \ambf \Vert^2_2 + \Vert \Bmbf \Vert^2_F = 1$, which implies that  $\Vert \Bmbf \Vert^2_F = 1 - \Vert \ambf \Vert^2_2 \leq 1$. 
Hence, the metrics in Definition~\ref{def:quadmet} are $1$-smooth. 

Now, we look at the error in Taylor series approximation when we approximate the metric $\phi^{\text{quad}}$ in  Definition~\ref{eq:quadmet} with a linear approximation. Our metric is 

$$
\phi^{\text{quad}}(\rmbf) = \inner{\ambf}{\rmbf} + \frac{1}{2}\rmbf^T\Bmbf\rmbf.
$$

We approximate it with the first order Taylor polynomial around a point $\zmbf$, which we define as follows:

$$
T_1(\rmbf) = \inner{\ambf}{\zmbf} + \frac{1}{2}\zmbf^T\Bmbf\zmbf + \inner{\ambf + \Bmbf\zmbf}{\rmbf}
$$
The bound on the error  in this approximation is:
\begin{align*}
\vert E(\rmbf) \vert &= \vert \phi^{\text{quad}}(\rmbf) - T_1(\rmbf) \vert   \\
&= \frac{1}{2} \vert (\rmbf -\zmbf)^T \Delta\phi^{\text{quad}}|_\cmbf  (\rmbf -\zmbf) \vert \qquad\qquad  \text{(First-order Taylor approximation error)} \\
&= \frac{1}{2} \vert (\rmbf -\zmbf)^T \Bmbf  (\rmbf -\zmbf) \vert \qquad\qquad\qquad \;\;\; \text{(Hessian at any point $\cmbf$ is the matrix $\Bmbf$)}\\
&\leq \frac{1}{2}\Vert \Bmbf \Vert_2 \Vert \rmbf - \zmbf \Vert _2^2 \\
&\leq \frac{1}{2}\Vert \Bmbf \Vert_F \varrho^2 \\
&\leq  \frac{1}{2} \varrho^2 \qquad\qquad\qquad\qquad\qquad\qquad\qquad \text{(Due to the scale invariance condition)}
\end{align*}

So when the oracle is asked $\Omega(\rmbf_1, \rmbf_2) = \1[\phi^{\text{quad}}(\rmbf_1) > \phi^{\text{quad}}(\rmbf_2)]$, the approximation error can be treated as feedback error from the oracle with feedback noise 
$2\times \frac{1}{2} \varrho^2$. 
Thus, the overall feedback noise by the oracle is 
$\epsilon_\Omega + \varrho^2$ for the purposes of using Lemma~\ref{lem:LPMEwnoise} later. 

We first prove guarantees for the matrix $\Bmbf$ and then for the vector $\ambf$. We write Equation~\eqref{eq:poly2elicitamatfinal} in the following form assuming $d_1 = 1$ (since we normalize the coefficients at the end due to scale invariance): 

\begin{align*}
B_{ij} &= F_{ij} =  \left[\frac{f_{ij}}{f_{1j}}\left(1 + \frac{f_{j1}}{f_{11}} \right) - \frac{f_{ij}}{f_{1j}}\frac{f_{j0}}{f_{10}} - \frac{f_{i0}}{f_{10}} + \frac{f_{ij}}{f_{1j}}\frac{f_{j1}}{f_{11}} \frac{ \frac{f_{21}^-}{f_{11}^-} + \frac{f_{21}}{f_{11}} - 2\frac{f_{20}}{f_{10}}  }{ \frac{f_{21}^-}{f_{11}^-} - \frac{f_{21}}{f_{11}}  }\right]. \\
\implies \Bmbf[:, j] &= \fmbf_j\left( \frac{1}{f_{1j}} + \frac{f_{j1}}{f_{1j}f_{11}} + \frac{f_{j0}}{f_{1j}f_{10}} + \frac{f_{j1}}{f_{1j}f_{11}}\left(  \frac{ \frac{f_{21}^-}{f_{11}^-} + \frac{f_{21}}{f_{11}} - 2\frac{f_{20}}{f_{10}}  }{ \frac{f_{21}^-}{f_{11}^-} - \frac{f_{21}}{f_{11}}  } \right) \right) + \fmbf_0\frac{1}{f_{10}} \\
&= c_j\fmbf_j + c_0\fmbf_0, \numberthis \label{eq:Bj}
\end{align*} 
where $\Bmbf[:, j]$ is the $j$-th column of the matrix $\Bmbf$, and the constants $c_j$ and $c_0$ are well-defined due to the regularity Assumption~\ref{as:regularity-q}. Notice that,
$$
\frac{\partial \Bmbf[:, j]}{\partial \fmbf_j} = \diag(\cmbf'_j)\odot\Imbf \quad, \text{and} \quad 
\frac{\partial \Bmbf[:, j]}{\partial \fmbf_0} = \diag(\cmbf'_0)\odot\Imbf,
$$
where $\cmbf'_j, \cmbf'_0$ are vector of Lipschitz constants (bounded due to Assumption~\ref{as:regularity-q}). This implies

\begin{align*}
\Vert \Bmbfbar[:, j] - \Bmbfhat[:, j]\Vert_2 &\leq c'_j \Vert \fmbfbar_j - \fmbfhat_j \Vert_2 + c'_0 \Vert \fmbfbar_0 - \fmbfhat_0 \Vert_2\\
&\leq c'_j\sqrt{k}\left(\epsilon + \sqrt{\varrho + \epsilon_\Omega/\varrho}\right) + c'_0\sqrt{k}\left(\epsilon + \sqrt{\varrho + \epsilon_\Omega/\varrho}\right) \\
&= O\left(\sqrt{k}\left(\epsilon + \sqrt{\varrho + \epsilon_\Omega/\varrho}\right)\right),
\end{align*}
where we have used LPME guarantees from Lemma~\ref{lem:LPMEwnoise} under the oracle-feedback noise parameter $\epsilon_\Omega + \varrho^2$. 

The above inequality provides bounds on each column of $\Bmbf$. Since $\Vert \xmbf \Vert_\infty \leq \Vert \xmbf \Vert_2$, we have $\max_{ij}\vert B_{ij} - \hat{B}_{ij} \vert \leq O\left(\sqrt{k}\left(\epsilon + \sqrt{\varrho + \epsilon_\Omega/\varrho}\right)\right)$, and consequentially, $$\Vert \Bmbf - \Bmbfhat \Vert_F \leq O\left(k\sqrt{k}\left(\epsilon + \sqrt{\varrho + \epsilon_\Omega/\varrho}\right)\right)$$. 

Now let us look at guarantees for $\ambf$. Since $\ambf = \dmbf - \Bmbf\ombf$ from~\eqref{eq:quadmetshift}, we can write 

$$
\ambf = c_0\fmbf_0 - \sum_{j=1}^k o_j\Bmbf[:, j],
$$
where $c_0 = 1/f_{10}$. Since $\ombf$ is the rate achieved by random classifier, $o_j = 1/k \; \forall j \in [k]$, and thus we have
$$
\frac{\partial \ambf}{\partial \fmbf_0} = c_0\Imbf \quad \text{and} \quad \frac{\partial \ambf}{\partial \Bmbf[:, j]} = \frac{1}{k}\Imbf.
$$
Thus,
\begin{align*}
\Vert \ambf - \ambfhat \Vert_2 &\leq c'_0 \sqrt{k}\left(\epsilon + \sqrt{\varrho + \epsilon_\Omega/\varrho}\right) + \frac{1}{k}\sum_{j=1}^k \sqrt{k}\left(\epsilon + \sqrt{\varrho + \epsilon_\Omega/\varrho}\right) \\
& =c'_0 \sqrt{k}\left(\epsilon + \sqrt{\varrho + \epsilon_\Omega/\varrho}\right) + \frac{1}{k}\sum_{j=1}^k c'_j\sqrt{k}\left(\epsilon + \sqrt{\varrho + \epsilon_\Omega/\varrho}\right) \\
&= O\left(\sqrt{k}\left(\epsilon + \sqrt{\varrho + \epsilon_\Omega/\varrho}\right)\right),
\end{align*}
where $c'_0, c'_j$'s are some Lipschitz constants (bounded due to Assumption~\ref{as:regularity-q}).
\eproof

Notice the trade-off in the elicitation error that depends on the size of the sphere. As expected, when the radius of the sphere $\varrho$ increases, the error due to approximation increases, but at the same time, error due to feedback reduces because we get better responses from the oracle. In contrast, when the radius of the sphere $\varrho$ decreases, the error due to approximation decreases, but the error due to feedback increases.

The following corollary translates our guarantees on the elicited metric to the guarantees on the optimal rate of the elicited metric. This is useful in practice, because the optimal classifier (rate) obtained by optimizing a certain metric is often the key entity for many applications. 

\bcorollary
Let $\phi^{\quadr}$ be the original quadratic metric of the oracle and $\hat\phi^{\quadr}$ be its estimate obtained by the QPME procedure (Algorithm~1). Moreover, let $\rmbf^*$ and $\hat\rmbf^*$ be the mazximizers of $\phi^{\quadr}$ and $\hat\phi^{\quadr}$, respectively. Then,  $\phi^{\quadr}(\rmbf^*) \leq \phi^{\quadr}(\hat\rmbf^*) + O\left(k^2\sqrt{k}\left(\epsilon + \sqrt{\varrho + \epsilon_\Omega/\varrho}\right)\right).$
\ecorollary

\bproof
We first show that if $\vert \phi^{\quadr}(r) - \hat\phi^{\quadr}(r)\vert \leq \epsilon$ for all rates $r$  and some slack $\epsilon$, then it follows that $\phi^{\quadr}(\hat\rmbf^*) \geq \phi^{\quadr}(\rmbf^*) - 2\epsilon.$ This is because:

\begin{align*}
    \phi^{\quadr}(\hat\rmbf^*) &\geq \hat\phi^{\quadr}(\hat\rmbf^*) - \epsilon \qquad\qquad \left(\text{as $\hat\phi^{\quadr}$ approximates $\phi^{\quadr}$}\right)\\
    &\geq \hat\phi^{\quadr}(\rmbf^*) - \epsilon \qquad\qquad \left(\text{as $\hat\rmbf^*$ maximizes $\hat\phi^{\quadr}$}\right) \\
    &\geq \phi^{\quadr}(\rmbf^*) - 2\epsilon \qquad\quad\;\; \left(\text{as $\hat\phi^{\quadr}$ approximates $\phi^{\quadr}$}\right) \numberthis \label{eq:metapx}
\end{align*}

Now, let us derive the trivial bound $\vert \phi^{\quadr}(r) - \hat\phi^{\quadr}(r)\vert$ for any rate $\rmbf$. 

\begin{align*}
    \vert \phi^{\quadr}(r) - \hat\phi^{\quadr}(r)\vert &= \vert \inner{\ambf - \hat \ambf}{\rmbf} + \frac{1}{2}\rmbf^T (\Bmbf - \hat\Bmbf)\rmbf \vert \\
    &\leq \vert \inner{\ambf - \hat \ambf}{\rmbf} \vert + \frac{1}{2}\vert \rmbf^T (\Bmbf - \hat\Bmbf)\rmbf \vert\\
    &\leq \Vert \ambf - \ambf \Vert_2 \Vert \rmbf\Vert_2 + \frac{1}{2}\Vert \Bmbf - \Bmbf \Vert_2 \Vert \rmbf\Vert_2^2\\
    &\leq \Vert \ambf - \ambf \Vert_2 \sqrt k + \frac{1}{2}\Vert \Bmbf - \Bmbf \Vert_F k\\
    &\leq O\left(k^2\sqrt{k}\left(\epsilon + \sqrt{\varrho + \epsilon_\Omega/\varrho}\right)\right), \numberthis \label{eq:metapx2}
\end{align*}
where in the fourth step, we have used the fact that the rates are bounded in $[0, 1]$; hence $\Vert \rmbf \Vert_2 \leq \sqrt{k}$, and in the fifth step, we have used the guarantees from Theorem~\ref{thm:q-me}. Combining\eqref{eq:metapx} and~\eqref{eq:metapx2} gives us the desired result. 
\eproof


\bproof[Proof of Theorem~\ref{thm:lb}]
For the purpose of this proof, let us replace $\left(\epsilon + \sqrt{\varrho + \epsilon_\Omega/\varrho}\right)$ by some slack $\epsilon$. Theorem 1 guarantees that after running the QPME procedure for $O(k^2\log(1/\epsilon)$ queries, we have 

\begin{itemize}
\item $\norm {a - \hat a}_2 \leq O(\sqrt k \epsilon)$
\item $\norm {B - \hat B}_F \leq O(k\sqrt k\epsilon).$
\end{itemize}

If we vectorize the tuple $(\ambf, \Bmbf)$ and denote it by $w$, we have $\norm{w - \hat w}_2 \leq O(k\sqrt k\epsilon)$, where both $\Vert w\Vert_2, \Vert \hat w\Vert_2=1$, due to the scale invariance condition from Definition~\ref{def:quadmet}. Note that $w$ is $\frac{k^2 + 3k}{2}$-dimensional vector and defines the scale-invariant quadratic metric elicitation problem. 
Now, we have to count the minimum number of $\hat w$ that are possible such that $\norm{w - \hat w}_2 \leq O(k\sqrt k\epsilon)$.

This translates to finding the covering number of a ball in $\Vert \cdot \Vert_2$ norm with radius 1, where the covering balls have radius $k\sqrt k\epsilon$. Let us denote the cover by $\{u_i\}_{i=1}^N$ and the ball with radius 1 as $\Bmbb$. We then have:

\begin{align*}
Vol(\Bmbb) &= \leq \sum_{i=1}^N Vol(k\sqrt k\epsilon \Bmbb + u_i) \\
&= NVol(k\sqrt k\epsilon \Bmbb) \\
&= (k\sqrt k\epsilon)^{\frac{k^2 + 3k}{2} - 1}.
\end{align*}

Thus the number of $\hat w$ that are possible are at least 
$$
c\left(\frac{1}{k\sqrt k\epsilon}\right)^{\frac{k^2 + 3k}{2} - 1} \leq N,
$$
where $c$ is a constant. Since each pairwise comparison provides at most one bit, at least $O(k^2)\log(\frac{1}{k\sqrt k\epsilon})$ bits are required to get a possible $\hat w$. We require $O(k^2)\log(\frac{1}{\epsilon})$ queries, which is near-optimal barring log terms. 
\eproof

\section{\textsc{Extended Experiments}}
\label{append:sec:extexp}

The source code is provided along with the supplementary material. The experiments in this paper were conducted on a machine with the following configuration: \emph{2.6 GHz 6 code Intel i7 processor with 16GB RAM.} 

\subsection{More Details on Simulated Experiments on Quadratic Metric Elicitation} 
\label{append:ssec:details}

\textbf{Number of queries.} First, we look at the number of queries that were actually required to elicit quadratic and fair (quadratic) metrics in Section~\ref{sec:experiments}. Recall that the QPME procedure (Algorithm~1) requires running the LPME subrotuine (Algorithm~\ref{alg:slme}) $k+2$ times. As discussed in Appendix~\ref{append:sec:slme}, each run of LPME requires at most $3 \times 3 \times k \log (\pi/2\epsilon)$ queries. So, the maximum number of queries required in eliciting a quadratic metric is $(k+2)\times 3 \times 3 \times k \times 8$ for a binary search tolerance $\epsilon=10^{-2}$, where we vary $k \in \{2, 3, 4, 5\}$ in the experiments. 

However, note that, the elicitation error shown in Figure~\ref{fig:recovery} is averaged over 100 simulated oracles, each one with its own simulated quadratic metric. Due to the nature of the binary search involved in the LPME subroutine (see Algorithm~\ref{alg:slme} and Figure~\ref{append:fig:shrink1}), not every reduction of the search interval requires three queries. Many times the interval can be shrunk in less than three queries. The actual number of queries may vary across the oracles. The number of queries averaged over the 100 oracles corresponding to experiments in Section~\ref{sec:experiments} is shown in Table~\ref{tab:numqueries}. 

Theorem~\ref{thm:lb} shows that our query complexity (which is linear in the number of unknowns) matches the lower bound for the problem, which means that it is theoretically impossible to obtain a better complexity order for our problem setup. In practice, it can be considerably reduced by making reasonable assumptions on the metric. 
For example, when the number of classes is large, one may just cluster the classes from error perspective. For example, one may assume same error costs for similar classes. This will reduce the number of unknowns to $O(c^2)$, where $c<<k$ is the number of cluster of classes.

\begin{table}[t]
    \centering
    \begin{tabular}{|c|c|}
    \hline
        \multicolumn{2}{|c|}{Number of queries for QPME}\\\hline
         \multirow{2}{*}{$k$} &  \\ 
         &  \\
         \hline
         2 & 265.43\\
         3 & 669.29\\
         4 & 1205.91\\
         5 & 1879.74\\
         \hline
    \end{tabular}
    \text{\;}
    \begin{tabular}{|c|c|c|c|c|}
    \hline
        \multicolumn{5}{|c|}{Number of queries for Fair-QPME}\\\hline
         \diagbox{$k$}{$m$}  & 2 & 3 & 4 & 5  \\ \hline
         2 & 332.10 & 867.65 & 1663.73 & 2738.59 \\
         3 & 796.37 & 2127.44 & 4094.31 & 6734.15\\
         4 & 1398.14 & 3808.67 & 7363.82 & 12180.84\\
         5 & 2130.92 & 5887.99 & 11454.18 & 18999.71\\
         \hline
    \end{tabular}
    \caption{Number of queries required for eliciting regular quadratic metrics (Def.~\ref{def:quadmet}) and fairness quadratic metrics (Def.~\ref{def:f-linmetric}) in Section~\ref{sec:experiments}. The number of predictive rates and sensitive groups are denoted by $k$ and $m$, respectively. Recall that a quadratic metric has $O(k^2)$ unknowns. We see that the number of queries is of order $O(k^2)$ for the quadratic metric in rates, and additionally, $O(m^2k^2)$ for the fair (quadratic) metrics. Theorem~\ref{thm:lb} shows that one cannot improve on this query complexity. Nonetheless, one may make more structural assumptions on the metric to bring down the number of queries in practice.}
    \label{tab:numqueries}
\end{table}

\textbf{Comparison to a baseline.} In Figures~\ref{fig:q_rec_a}--\ref{fig:q_rec_B}, we show box plots of the $\ell_2$ (Frobenius) norm between the true and elicited linear (quadratic) coefficients. 
We  generally find that QPME is able to elicit metrics  close to the true ones.

\begin{figure*}[h]
	\centering 
	\subfigure{
		{\includegraphics[width=5cm]{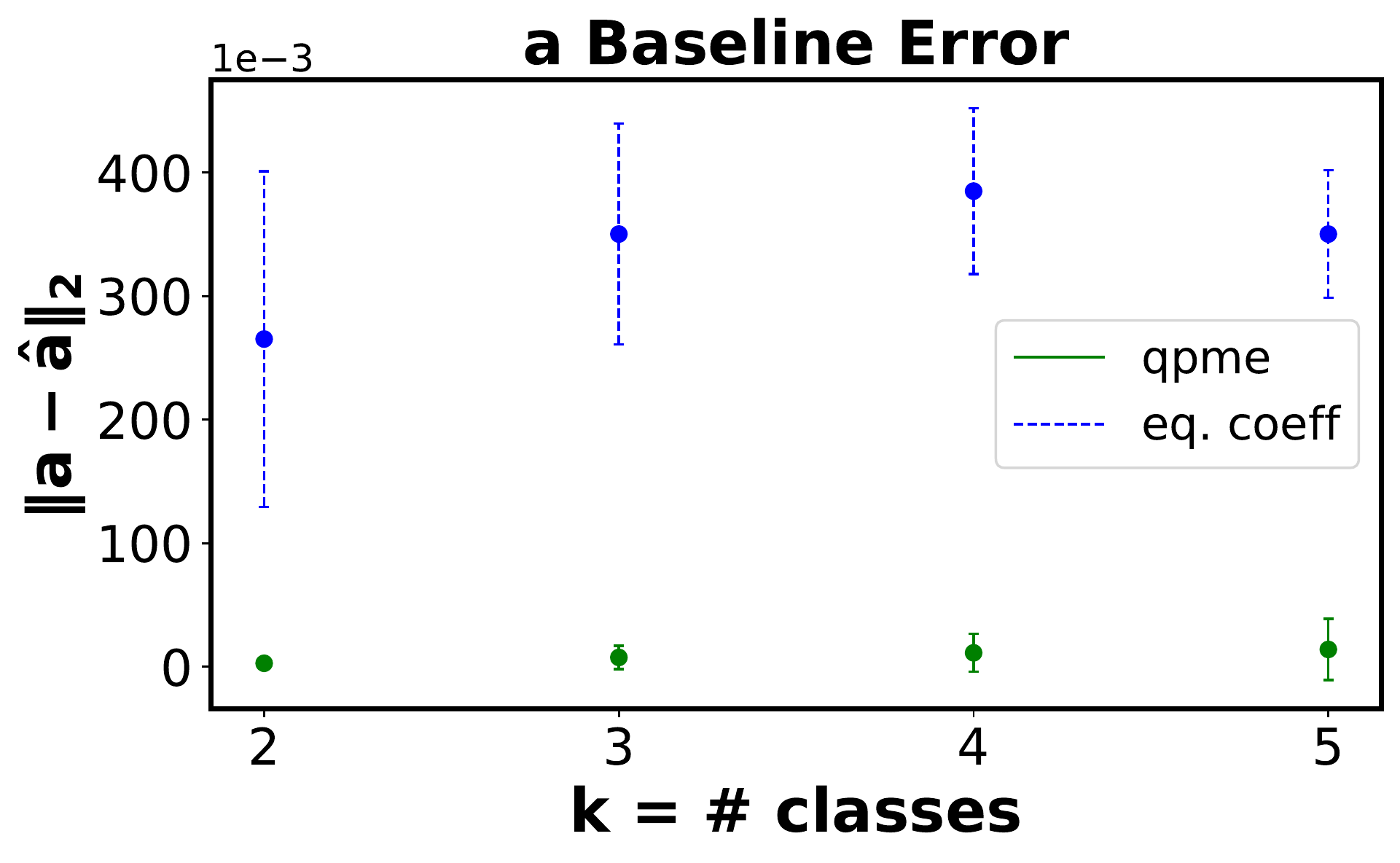}}
		\label{fig:rec_a}
	}\quad\quad
	\subfigure{
		{\includegraphics[width=5cm]{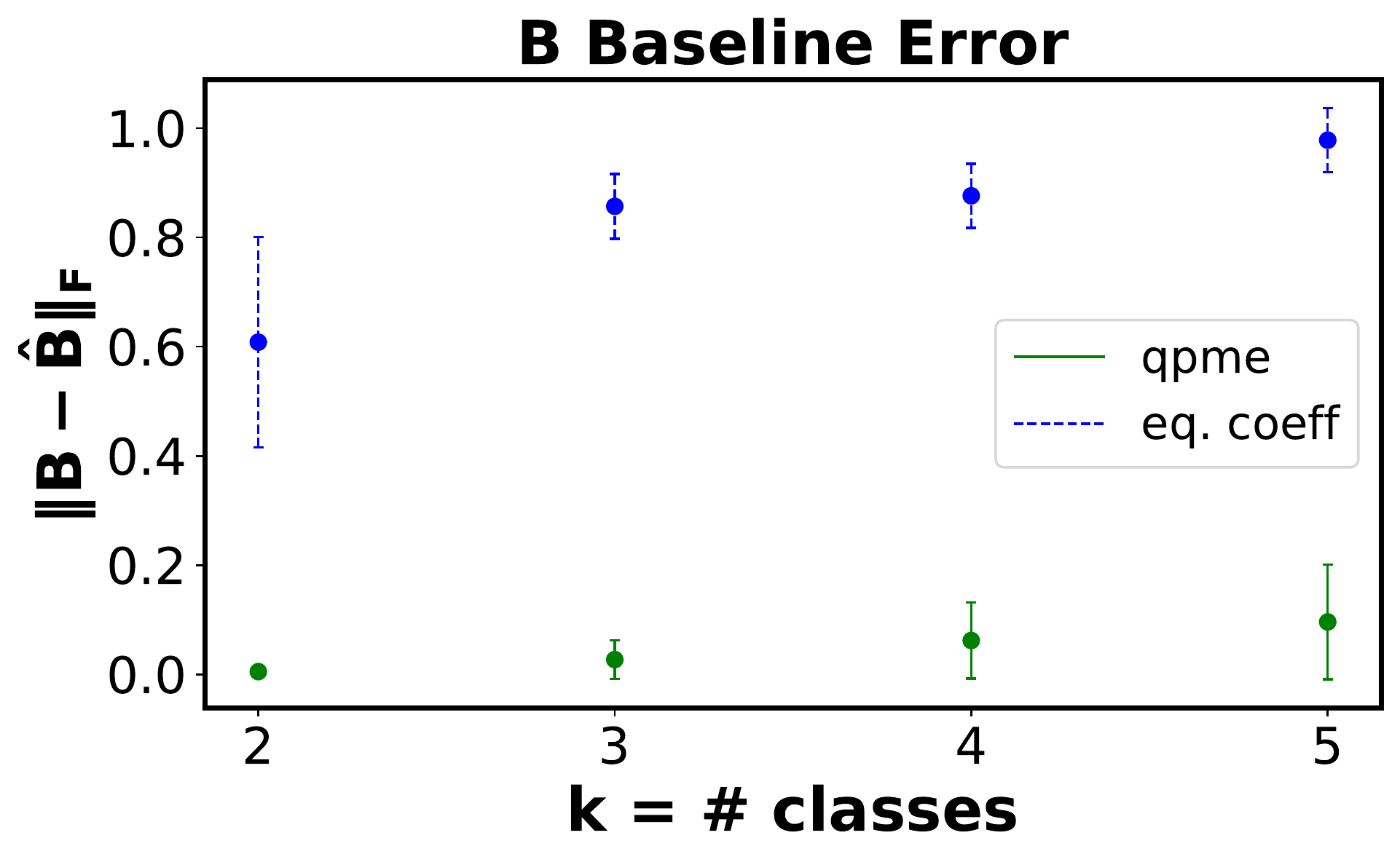}}
		\label{fig:rec_l}
	}
	\caption{Elicitation error in comparison to a baseline which assigns equal coefficients.}
	\label{append:fig:baseline}
\end{figure*}

To reinforce this point, we also compare the elicitation error of the QPME procedure and the elicitation error of a baseline which assigns equal coefficients to $\ambf$ and $\Bmbf$ in Figure~\ref{append:fig:baseline}. We see that the elicitation error of the baseline is order of magnitude higher than the elicitation error of the QPME procedure. This holds for varying $k$ showing that the QPME procedure is able to elicit oracle's multiclass quadratic metrics very well. 

\textbf{Effect of Assumption~\ref{as:regularity-q}}. 
The larger standard deviation for $k=5$ in Figure~\ref{fig:recovery} is due to Assumption~\ref{as:regularity-q} failing to hold with sufficiently large constants $c_{0},c_{-1}, c_1 \ldots, c_q$ in a small number of trials and the resulting estimates not being as accurate. 
We now analyze in greater detail the effect of this regularity assumption in eliciting quadratic metrics and understand how the lower bounding constants 
impact the elicitation error. Assumption~\ref{as:regularity-q} effectively ensures that the ratios computed in~\eqref{eq:poly2elicitamatfinal} are well-defined. To this end, we generate two sets of 100 quadratic metrics. One set is generated following Assumption~\ref{as:regularity-q} with one coordinate in the gradient being greater than $10^{-2}$, and the other is generated randomly without any regularity condition. For both sets, we run QPME and elicit the corresponding metrics. 

\begin{figure*}[h]
	\centering 
	\subfigure{
		{\includegraphics[width=5cm]{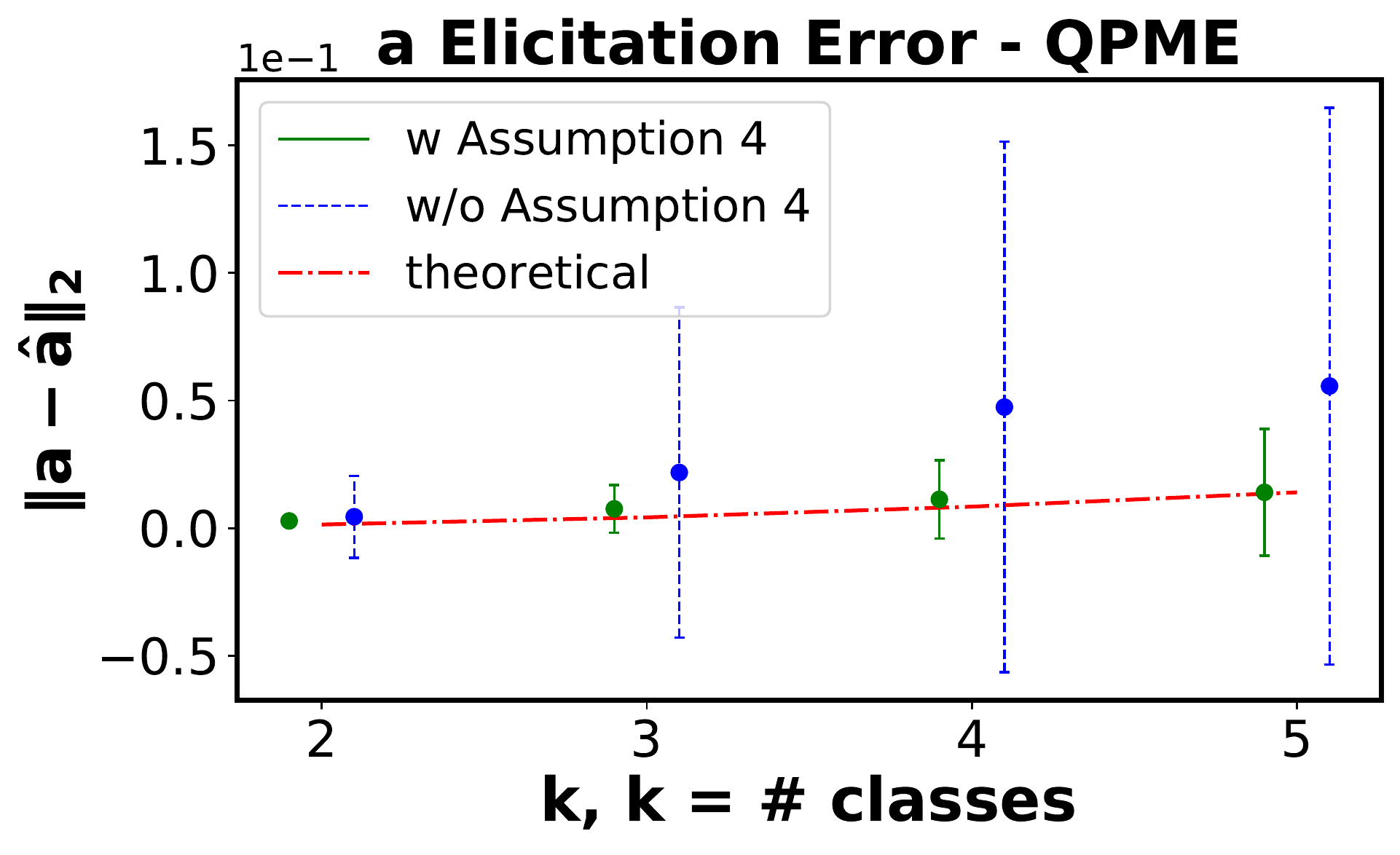}}
		\label{fig:rec_a}
	}\quad\quad
	\subfigure{
		{\includegraphics[width=5cm]{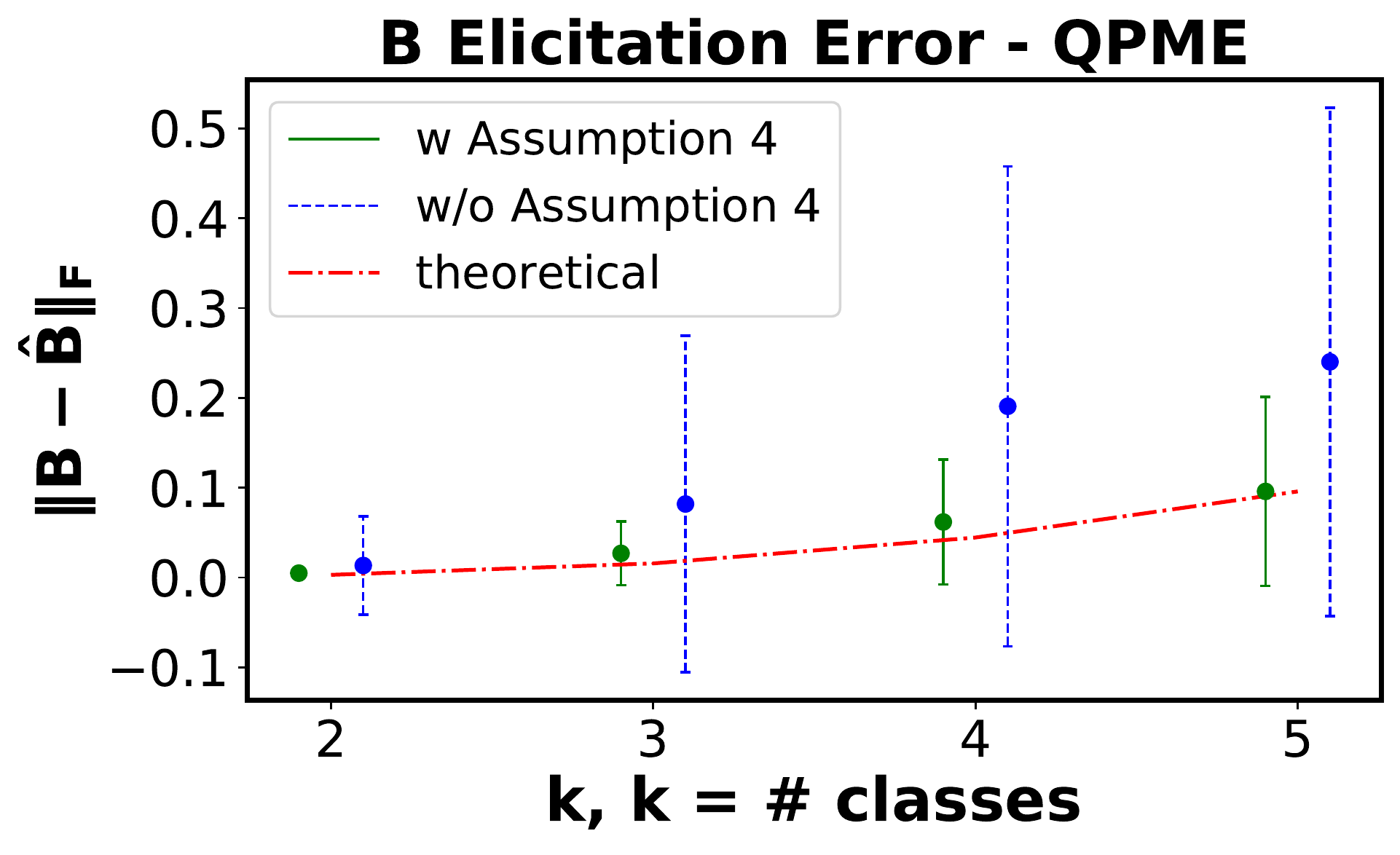}}
		\label{fig:rec_l}
	}
	\caption{Elicitation error for metrics following Assumption~\ref{as:regularity-q} vs elicitation error for completely random metrics.}
	\label{append:fig:regassump}
\end{figure*}

In Figure~\ref{append:fig:regassump}, we see that the elicitation error is much higher when the regularity Assumption~\ref{as:regularity-q} is not followed, owing to the fact that the ratio computation in~\eqref{eq:poly2elicitamatfinal} is more susceptible to errors when gradient coordinates approach zero in some cases of randomly generated metrics. The dash-dotted curve (in red color) shows the trajectory of the theoretical bounds with increasing $k$ (within a constant factor). In Figure~\ref{append:fig:regassump}, we see that  the mean of $\ell_2$ (analogously, Frobenius) norm better follow the theoretical bound trajectory in the case when regularity Assumption~\ref{as:regularity-q}
 is followed by the metrics.

\begin{figure*}[t]
	\centering 
	\subfigure{
		{\includegraphics[width=4cm]{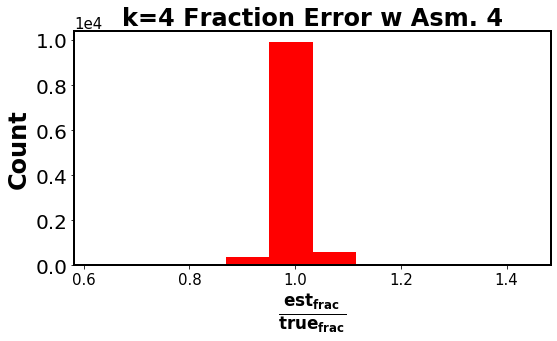}}
		\label{fig:rec_l}
	} 
	\subfigure{
		{\includegraphics[width=4cm]{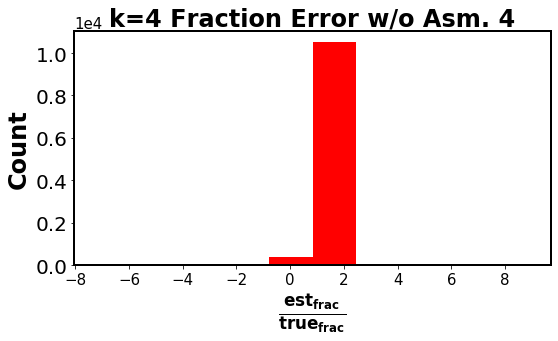}}
		\label{fig:rec_a}
	} \\
	\subfigure{
		{\includegraphics[width=4cm]{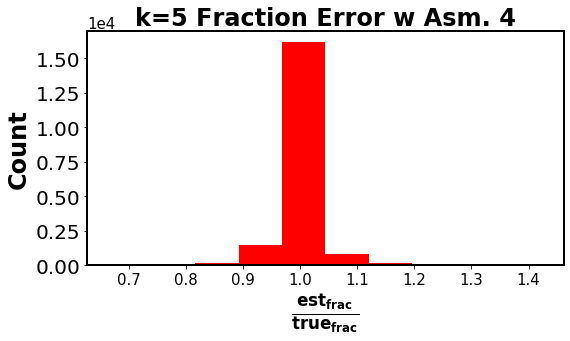}}
		\label{fig:rec_l}
	}
	\subfigure{
		{\includegraphics[width=4cm]{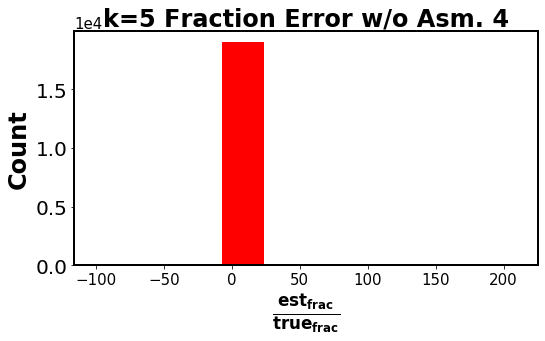}}
		\label{fig:rec_a}
	}
	\caption{Ratio of estimated to true fractions over 1000 simulated runs with and without Assumption~\ref{as:regularity-q}.}
	\label{append:fig:ratio}
\end{figure*}

We next analyze the ratio of estimated fractions to the true fractions used in~\eqref{eq:poly2elicitamatfinal} over 1000 simulated runs. Ideally, this ratio should be 1, but as we see in Figure~\ref{append:fig:ratio}, these estimated ratios can be off by a significant amount for a few trials when the metrics are generated randomly. The estimated ratios, however, are more stable under Assumption~\ref{as:regularity-q}. Since we multiply fractions in~\eqref{eq:poly2elicitamatfinal}, even then we may observe the compounding effect of fraction estimation errors in the final estimates. Hence, we see for $k=5$ in Figure~\ref{fig:q_rec_a}-\ref{fig:q_rec_B}, the standard deviation is high due to few trials where the lower bound of $10^{-2}$ on the constants in Assumption~\ref{as:regularity-q}  may not be enough. However, majority of the trials as shown in Figure~\ref{fig:q_rec_a}-\ref{fig:q_rec_B} and Figure~\ref{append:fig:baseline} incur low elicitation error. 

\subsection{Ranking of Real-World Classifiers}
\label{append:ssec:ranking}

Performance metrics provide quantifiable scores to classifiers. This score is then often used to rank classifiers and select the best set of classifiers in practice. In this section, we discuss the benefits of elicited metrics in comparison to some default metrics while ranking real-world classifiers. 

\begin{table}[t]
\centering
\caption{Dataset statistics}
\begin{tabular}{|c|ccc|}
\hline
\textbf{Dataset} & $k$ & \textbf{\#samples} & \textbf{\#features} \\ 
\hline
default        & 2   & 30000           &    33    \\
adult        &  2  &    43156       &    74   \\
sensIT Vehicle        &  3  & 98528          &     50   \\
covtype        &  7 &    581012       &     54 \\ 
\hline
\end{tabular}
\label{append:tab:stats}
\end{table}

\textbf{Ranking in case of quadratic metrics:} For this experiment, we work with four real world datasets with varying number of classes $k\in \{2,3, 7\}$. See Table~\ref{append:tab:stats} for details of the datasets. We use 60\% of each dataset to train classifiers. The rest of the data is used to compute (testing) predictive rates. For each dataset, we create a pool of 80 classifiers by tweaking hyper-parameters in some famous machine learning models that are routinely used in practice. Specifically, we create 20 classifiers each from logistic regression models~\citep{kleinbaum2002logistic}, multi-layer perceptron models~\citep{pal1992multilayer},  LightGBM models~\citep{ke2017lightgbm}, and support vector machines~\citep{joachims1999svmlight}. 
We compare ranking of these 80 classifiers provided by competing baseline metrics with respect to the ground truth ranking, which is provided by the oracle's true metric. 

We generate a random quadratic metric $\phi^{\text{quad}}$ following Definition~\ref{def:quadmet}. We treat the true $\phi^{\text{quad}}$ as oracle's metric. It provides us the ground truth ranking of the classifiers in the pool. We then use our proposed procedure QPME (Algorithm~1) to recover the oracle's metric. For comparison in ranking of real-world classifiers, we choose two linear metrics that are routinely employed by practitioners as baselines. The first is accuracy $\phi^{acc} = 1/\sqrt{k}\inner{\bm{1}}{\rmbf}$, and the second is weighted accuracy, where we just use the linear part  $\inner{\ambf}{\rmbf}$ of the oracle's true quadratic metric $\inner{\ambf}{\rmbf} + \frac{1}{2}\rmbf^T\Bmbf\rmbf$. We repeat this experiment over 100 trials. 

\begin{figure*}[h]
	\centering 
	\subfigure{
		{\includegraphics[width=5cm]{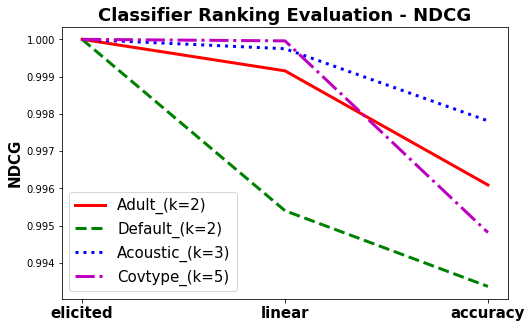}}
		\label{fig:rec_B}
	}\quad\quad
	\subfigure{
		{\includegraphics[width=5cm]{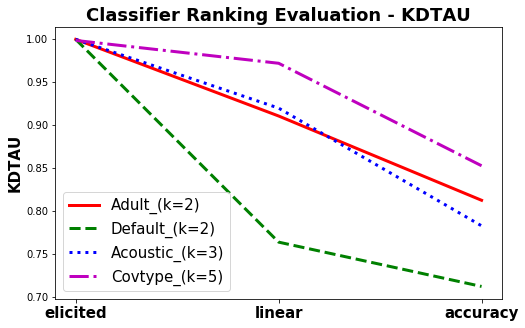}}
		\label{fig:rec_l}
	}
	\caption{Performance of competing metrics while ranking real-world classifiers. `elicited' is the metric elicited by QPME, `linear' is the metric that comprises only the linear part of the oracle's true quadratic metric, and `accuracy' is the linear metric which weigh all classification errors equally (often used in practice).}
	\label{append:fig:ranking}
\end{figure*}

We report NDCG (with exponential gain)~\citep{valizadegan2009learning} and Kendall-tau coefficient~\citep{shieh1998weighted} averaged over the 100 trials in Figure~\ref{append:fig:ranking}. We observe consistently for all the datasets that the elicited metrics using the QPME procedure achieve the highest possible NDCG and Kendall-tau coefficient of 1. As we saw in Section~\ref{sec:guarantees}, QPME may incur elicitation error, and thus the elicited metrics may not be very accurate; however, Figure~\ref{append:fig:ranking} shows that the elicited metrics may still achieve near-optimal ranking results. This implies that when given a set of classifiers, ranking based on elicited metric scores align most closely to true ranking in comparison to ranking based on default metric scores. Consequentially, the elicited metrics may allow us to select or discard classifiers for a given task. This is advantageous in practice. 
For the \emph{covtype} dataset, we see that the \emph{linear} metric also achieves high NDCG values, so perhaps ranking at the top is quite accurate; however Kendall-tau coefficient is low suggesting that the overall ranking of classifiers is poor. We also observe that, in general, the weighted version (\emph{linear} metric) is better than \emph{accuracy} while ranking classifiers.

\textbf{Ranking in case of fair (quadratic) metrics:} With regards to fairness, we performed a similar experiment as above for comparing fair-classifiers' ranking on Adult and Default datasets with gender as the protected group. There are two genders provided in the datasets, i.e., $m=2$. We simulate fairness metrics as given in Definition~\ref{def:f-linmetric} that gives ground-truth ranking of classifiers and evaluate the ranking by the elicited (fair-quadratic) metric using the procedure described in Section 4 (also depicted in Figure~\ref{fig:fairness-workflow}). In Table~\ref{append:tab:rankingfpme}, we show the NDCG and KD-Tau values for our method and for three baselines: (a) `Linear with no fairness', which is the metric that comprises only the linear part of the oracle's true quadratic fair metric from Definition~\ref{def:f-linmetric} without the fairness violation, (b) `Accuracy with eq. odds' is the metric which weigh all classification errors and fairness violations equally, and (c) Fair Performance Metric Elicitation (FPME) procedure from~\citep{hiranandani2020fair}.\footnote{While FPME~\citep{hiranandani2020fair} does not elicit a quadratic metric, one can still compare the elicited metrics based on how they rank candidate classifiers on real-world data.} We again see that the ranking by the metric elicited using the proposed fair-QPME procedure (Section~\ref{sec:fairme}) is closest to the ground-truth ranking.  The metric elicited by FPME~\citep{hiranandani2020fair} ranks classifiers better than ‘Linear with no fairness’ and ‘Accuracy with equalized odds’; however, it is beaten by the proposed fair-QPME procedure.

\begin{table}[h]
    \centering
    \begin{tabular}{|c|c|c|c|c|}
    \hline
    \textbf{Dataset}\;\;$\rightarrow$  & \multicolumn{2}{|c|}{\textbf{Adult}} &\multicolumn{2}{|c|}{\textbf{Default}} \\
    \hline
    \textbf{Method}$\;\;\downarrow$, \textbf{Ranking Measures}\;\;$\rightarrow$ & \textbf{NDCG} & \textbf{KD-TAU} & \textbf{NDCG} & \textbf{KD-TAU}\\
    \hline
         Linear with no fairness & 0.9875 & 0.5918 & 0.9994 & 0.9057 \\
         Accuracy with equalized odds & 0.9857 & 0.3763 & 0.9889 & 0.4953\\
         Elicited via FPME~\citep{hiranandani2020fair} & 0.9989  & 0.9611 & 0.9974 & 0.9650\\
         Elicited via Fair-QPME (Proposed) & \textbf{1.0000} & \textbf{0.9972} & \textbf{1.0000} & \textbf{0.9981}\\
         \hline
    \end{tabular}
    \caption{Performance of competing metrics while ranking real-world classifiers for fairness. `Linear with no fairness' is the metric that comprises only the linear part of the oracle's true quadratic fair metric from Definition~\ref{def:f-linmetric} without the fairness violation, `Accuracy with eq. odds' is the metric which weigh all classification errors and fairness violations equally (often used in practice), `Elicited via FPME~\citep{hiranandani2020fair}' is the metric elicited using the procedure from~\cite{hiranandani2020fair},  `Elicited via Fair-QPME' is the metric elicited by the proposed (quadratic) fairness metric elicitation procedure from Section 4 (also depicted in Figure~\ref{fig:fairness-workflow}),}
    \label{append:tab:rankingfpme}
\end{table}

\textbf{Ranking in case of added structural assumptions on the metrics:} Lastly, we discuss an experiment where we show how one may make structural assumptions on the metric when the \emph{actual} number of unknowns is large and still get comparable results in practical settings. For this experiment, we assume that the oracle's true metric is quadratic in general rate entries as explained in Appendix~\ref{append:generalquad}. Thus, the number of unknowns is $O(q^2)$, where $q = k^2 - k$ and is the number of off-diagonal entries of the rate matrix. We can apply the QPME procedure as it is and elicit a quadratic metric in general rates with $O(q^2)$ queries (see Appendix~\ref{append:generalquad}), since there are $O(q^2)$ unknowns. 

Note that, even if the oracle's original metric is a quadratic metric in off-diagonal entries, as a heuristic, we could still use our procedure to elicit a quadratic metric in diagonal rate entries. Moreover, we can use LPME procedure (Appendix~\ref{append:sec:slme}), too, to elicit a linear metric in off-diagonals and diagonal rate entries depending on the assumption we make on the metric.

Thus, for the ranking based experiments explained in Figure~\ref{append:fig:ranking},  we additionally ran (a) linear elicitation with diagonal rates, (b) linear elicitation with general rates, and (c) quadratic elicitation with diagonal rates, and compare their ranking with the elicited quadratic metric in general rates.
As seen in Table~\ref{append:tab:apxranking}, the quadratic approximation in the diagonal rates performs significantly better than eliciting a linear approximation in the general rates, while requiring the same query complexity ($\tilde{O}(k^2)$), and is close to the elicited quadratic metric in general rates, which require ($\tilde{O}(k^4)$) queries. Hence, one can make structural assumptions on the metric to reduce the query complexity and still get comparable results in practice. 

\begin{table}[h]
    \centering
    \scriptsize{
    \setlength\tabcolsep{3pt}
    \begin{tabular}{|c|c|c|c|c|}
    \hline
    \textbf{Dataset}\;\;$\rightarrow$  & \multicolumn{2}{|c|}{\textbf{Adult dataset}} &\multicolumn{2}{|c|}{\textbf{Default dataset}} \\
    \hline
    \textbf{Elicited Metric}$\downarrow$, \textbf{Rank- Measure}$\rightarrow$ & \textbf{NDCG} & \textbf{KD-TAU} & \textbf{NDCG} & \textbf{KD-TAU}\\
    \hline
         Linear-diagonal ($\tilde{O}(k)$) & 0.9783 & 0.6053 & 0.9790 & 0.4536 \\
         \highlight{Linear-general ($\tilde{O}(k^2)$)} & \highlight{0.9908} & \highlight{0.7713} & \highlight{0.9863} & \highlight{0.6216}\\
         \highlight{Quadratic-diagonal ($\tilde{O}(k^2)$)} & \highlight{0.9968} & \highlight{0.9611} & \highlight{1.0000} & \highlight{0.9979}\\
         Quadratic-general ($\tilde{O}(k^4)$) & \textbf{1.0000} & \textbf{0.9986} & \textbf{1.0000} & \textbf{0.9979}\\
         \hline
    \end{tabular}
    }
    \vskip -0.2cm
    \caption{Performance on ranking real-world classifiers when the oracle's true metric is quadratic in general rate entries: the elicited \emph{quadratic metric in diagonal entries perform better than elicited linear metric} in general rate entries (while requiring same no.\ of queries), and close to the elicited quadratic metric in general rates.
    }
    \vskip -0.2cm
      \label{append:tab:apxranking}
\end{table}

\section{Extended Related Work}
\label{append:sec:relwork}

We discuss how the area of metric elicitation, in general, and our  quadratic elicitation proposal, in particular, differ from two related fields: (i) inverse reinforcement learning and (ii) ranking from pairwise comparisons using choice models. 

\subsection{Inverse reinforcement learning (RL)} 
The idea of learning a reward/cost function in the inverse RL problems is conceptually similar to metric elicitation. However, there are many key differences. Studies such as~\citep{ng2000algorithms, wu2020efficient, abbeel2004apprenticeship} try to learn a linear reward function either by knowing the optimal policy or expert demonstrations. Not only is the type of feedback in these studies different from the pairwise feedback we handle, but the studies are focused on linear rewards. In contrast, our goal is to use pairwise feedback to elicit quadratic metrics which are important for classification problems, especially, fairness. Indeed nonlinear reward estimation in inverse RL problems has been tackled before~\citep{levine2011nonlinear, fu2017learning}, but these are passive learning approaches and do not come with query complexity guarantees like we do. Because of the use of a complex function class, these methods are not easy to analyze. There has been some work on actively estimating the reward function in the inverse RL problems~\citep{lopes2009active}; this work involves discretizing the feature space and using maximum entropy based ideas to elicit a distribution over rewards, which clearly uses a different set of modeling assumptions than us. A recent work~\citep{sadigh2017active} elicits reward functions through active learning, but is again tied to eliciting linear functions and provides limited theoretical guarantees; whereas, we specifically focus on quadratic elicitation with rigorous guarantees.

In summary, our work is significantly different from inverse RL methods, in that unlike them, we are tied to a particular geometry of the query space (the space of error statistics achieved by feasible classifiers), and elicit quadratic (or polynomial) functions from pairwise comparisons, specifically, in an active learning manner. 

\subsection{Ranking from pairwise comparisons} 
Our work  is also quite different from the use of choice models such the Bradley-Terry-Luce (BTL)  model for rank aggregation. (i) Firstly, choice models such as BTL are commonly used to learn a aggregate global ranking of a finite set of  $N$ items from pairwise comparisons. The underlying problem involves estimating a $N$-dimensional quality score vector for the items~\citep{shah2015estimation}. In contrast, metric elicitation estimates an oracle’s classification metric, a function of a classifier’s error statistics. The applications for the two problems are very different: while ranking aggregation strategies using BTL are often prescribed for aggregating user opinions on a restaurant or product, metric elicitation seeks to find the right objective to optimize for a classification task. 
(ii) Secondly, the noise model in  BTL  is stochastic and depends on the distance between the quality of items; whereas, our noise model in Definition~\ref{def:noise} is not stochastic and is oblivious to the distance of rates unless the rates are very close.
(iii) Thirdly, while there is some work on the extended BTL model where items are represented by feature vectors~\citep{niranjan2017inductive}, and the goal is to learn weights on the features to complete the ranking of the items, most of the work in this area considers a passive setting, where pairwise comparisons are assumed to be iid. In contrast, our work involves actively learning nonlinear utilities with theoretical bounds on query complexity.
(iv) Lastly, the closest active learning work we could find with BTL models \citep{mohajer2017active} does not generalize to feature-dependent utilities and is proposed for finding the top-$k$ items, which is entirely different from metric elicitation.
~\\[-5pt]

Other fields that are less closely related to our work include learning scoring functions for supervised label ranking problems \citep{furnkranz2010preference}, and the more traditional metric learning literature, where the task is to learn a distance metric that captures similarities between data points, with the goal of using it for downstream learning tasks \citep{kulis2013metric}.

\section{Preliminary User Study}
\label{append:userstudy}

We are actively conducting user studies for eliciting performance metrics. In this section, we provide a peek into our future work. The goal of this preliminary study is to check workflow of the practical implementation of the metric elicitation framework with real data, and to a certain extent, support or reject the hypothesis that the implicit user preferences can be quantified using the pairwise comparison queries over confusion matrices or predictive rates. In addition, the goal includes testing certain assumptions regarding the noise in the subject's (oracle's) responses, work around with finite samples, eliciting actual performance metrics in real-life scenarios, and evaluating the quality of the recovered metric. 

The following user study works with the space of confusion matrices, i.e., entries of the type $\Pmbb(Y=i, h=j)$ for $i,j\in[k]$, instead of the predictive rates. In the future, we plan on incorporating rates, i.e., entries of the type $\Pmbb(h=j|Y=i)$ for $i,j\in[k]$, in the visualizations as well. Our contributions are summarized as follows:

\begin{itemize}
    \item We create a web UI that uses existing visualizations of confusion matrices (predictive rates) that are refined to capture preferences over pairwise comparisons. 
    \item The UI implements the binary-search procedure from Algorithm~\ref{alg:slme} at the back end that make use of the real-time responses over confusion matrices to elicit a linear performance metric for a binary classification task. 
    \item We perform a user study with ten subjects and elicit their linear performance metrics using the proposed web UI. We compare the quality of the recovered metric  by comparing their responses to the elicited metric's responses over a set of randomly chosen pairwise comparison queries. 
\end{itemize}

\subsection{Choice of Task and Dataset Used}
\label{pme-ssec:dataset}

Our choice of task is \emph{cancer diagnosis}~\citep{yang2014multiclass} for which we use the Breast Cancer Wisconsin (Original) dataset from the UCI repository.\footnote{The dataset can be downloaded from https://tinyurl.com/dn2esyvw.} The dataset has been extensively used in the literature for binary classification, where the label $1$ denotes \emph{malignant} cancer and label $0$ denotes \emph{benign} cancer. There are 699 samples in total, wherein each sample has 9 features. The task for any classifier is to take the 9 features of a patient as input and predict whether or not the patient has cancer. We divide this data into two equally sized parts -- the training and the test data. Using the training data, we learn a logistic regression model to obtain an estimate of the class-conditional probability, i.e., $\hat\eta(x) = \hat\Pmbb(y=1 | X)$. We then create a sphere using Algorithm~\ref{alg:sphere} inside the space of confusion matrices computed on the test data . 

\subsection{Choice of Visualization}
\label{pme-ssec:vis}

\begin{figure}[t]
    \centering
    \includegraphics[scale=0.5]{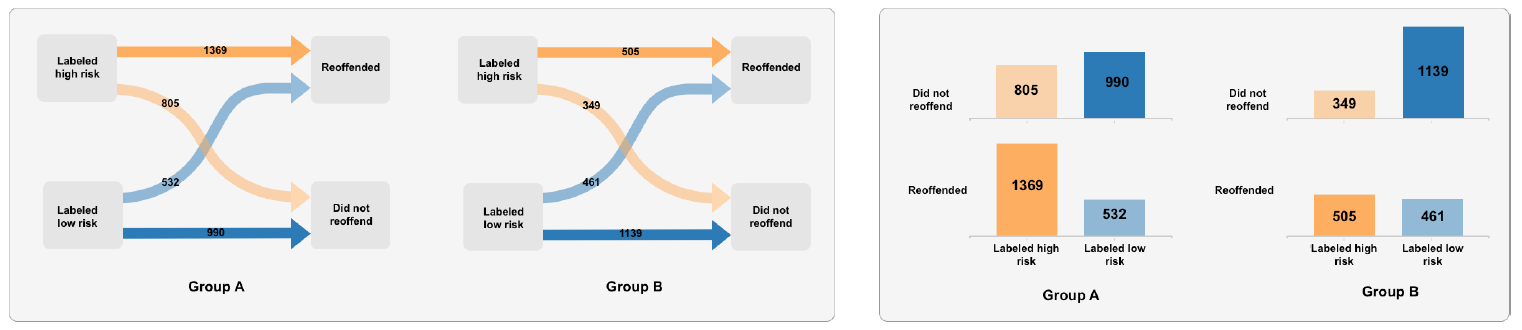}
    \caption{Flow-chart and bar-chart based visualizations for (binary classification) confusion matrices in the recidivism prediction task from~\cite{shen2020designing}.}
    \label{pme-fig:vis-prior}
    \vspace{-0.5cm}
\end{figure}

In modern times, ensuring effective public understanding of algorithmic decisions, especially, machine learning models has become an imperative task. With this view in mind, we borrow the visualizations of confusion matrices for the binary classifications setup from~\cite{shen2020designing}. The authors provide a concrete step towards the above goal by redesigning confusion matrices to support non-experts in understanding the performance of machine learning models. The final visualizations that we use from~\cite{shen2020designing} are created over multiple iterative user-studies. The visualizations are shown in Figure~\ref{pme-fig:vis-prior} in the context of a recidivism prediction task. One is the \emph{flow-chart}, which helps users in understanding the direction of the data, and the other is the \emph{bar chart}, which helps users in understanding the quantities involved.

However, in light of our preliminary discussions with Human-Computer Interaction (HCI) and machine learning researchers, we make/recommend the following changes in the visualization for pairwise comparison purposes in the metric elicitation framework.

\begin{figure}
    \centering
    \includegraphics[scale=0.5]{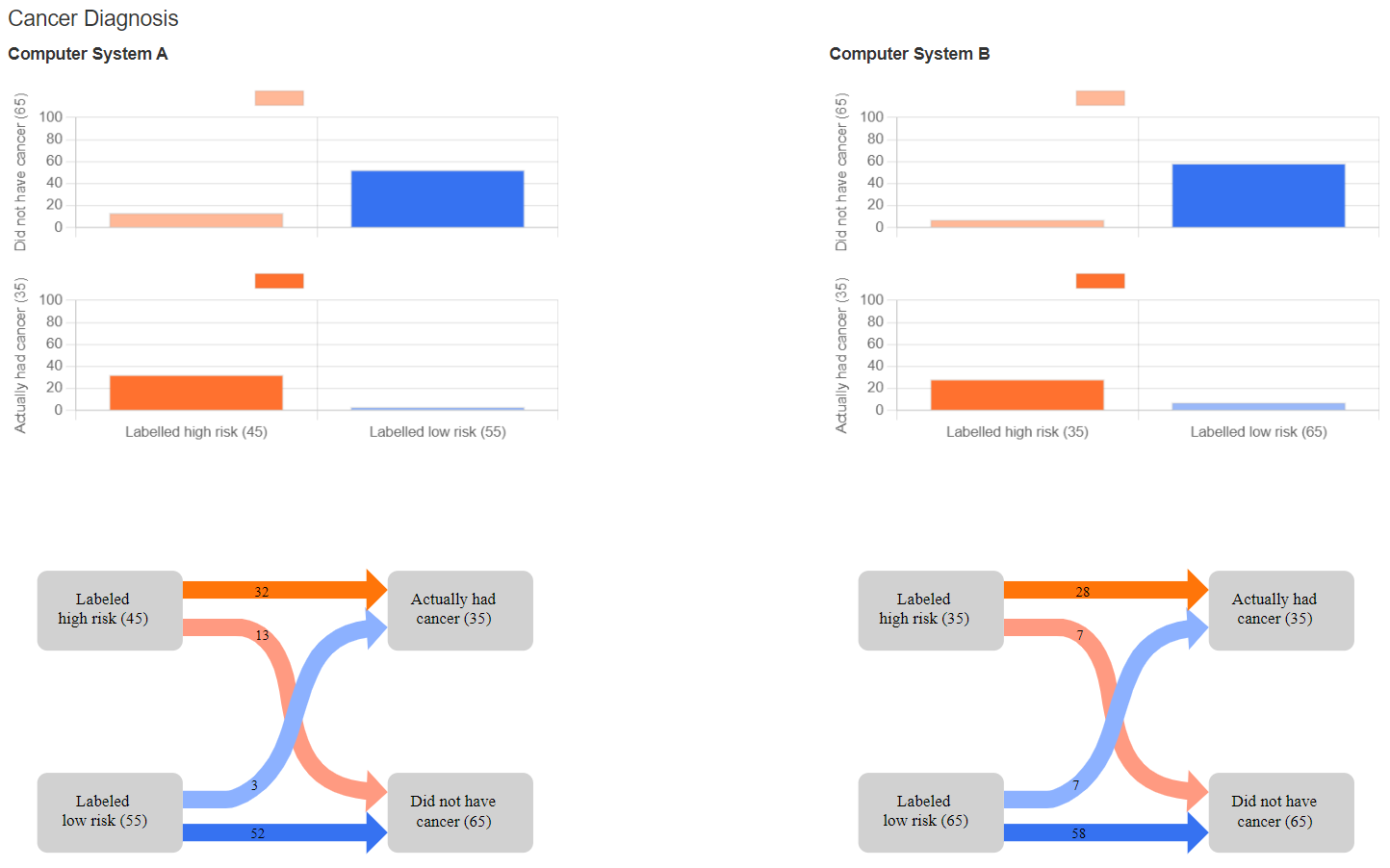}
    \includegraphics[scale=0.5]{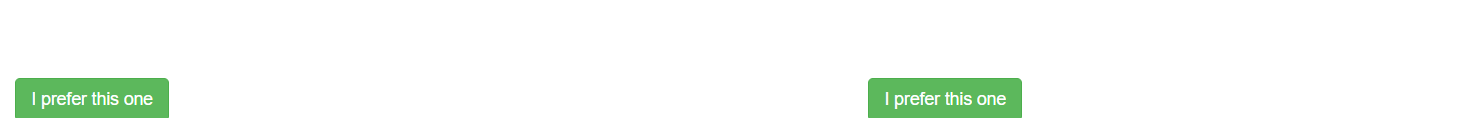}
    \caption{A sample of a pairwise comparison query from a run of the binary-search based procedure Algorithm~\ref{alg:slme}}
    \label{pme-fig:me}
\end{figure}

\begin{enumerate}
    \item Based on the observation that multiple visualizations of the information help in better user understanding, we choose to use both \emph{flow-chart} and \emph{bar-chart}, together to depict a confusion matrix. 
    \item We transform the data statistics so that the numbers denote out-of-100 samples. 
    \item We found that the total number of positive and negative labels along with total number of positive and negative predictions are very helpful in comparing two confusion matrices. Therefore, we add the total numbers in the flow-chart boxes and on axes in the bar-charts.
\end{enumerate}
A sample of a pairwise comparison query with modified visualizations incorporating the points above is shown in Figure~\ref{pme-fig:me}. We next discuss the web user interface. 

\subsection{User Interface}
\label{ssec:ui}

We discuss our proposed web User Interface (UI) in detail and discuss our rationale behind its components. The UI has two parts to it as explained below.

The first phase of the UI is where we actually ask subjects for pairwise preferences over confusion matrices, and implement our binary-search procedure from Algorithm~\ref{alg:slme}. The subjects have to make a choice reflecting on the trade-off between false positives and false negatives. Algorithm~\ref{alg:slme} takes in real-time preferences of the subjects, generates next set of queries based on the current responses, and converge to a linear performance metric at the back end. We save this (linear) performance metric for each subject. We stop the binary-search when the search interval becomes less than or equal to 0.05 ($\epsilon$ in line 1 of Algorithm~\ref{alg:slme}).

In order to evaluate the quality of the recovered metric, in the second phase, we ask the subjects fifteen pairwise comparison queries, each on a separate web page, right after the binary search algorithm  has converged, and we have elicited the metric. The subjects do not know this information and are shown evaluation queries in continuation to the previous phase (i.e., the binary search). The query comprises of two randomly selected confusion matrices that lie inside the feasible region.  This set of  queries is used to evaluate the effectiveness of the elicited metric. 

\subsection{Study Results}
\label{append:ssec:studyres}
We compute the fraction of times our elicited metric's preferences matches with the subject's preferences on the fifteen queries, i.e., 
\begin{equation}
\Mcal := \frac{\sum_{i=1}^{15} \1[\text{subject's prefer. for query } i == \text{metric's prefer. for query } i]}{15} \times 100.
    \label{pme-eq:fraction}
\end{equation}

We show the elicited metric for the fifteen subjects and the measure $\Mcal$ values in Table~\ref{pme-tab:metrics}. We see for nine out of ten subjects that more than 85\% of the time our elicited metric's preferences matches with the subject's preferences on the fifteen evaluation queries. For three subjects, our metric's preference matches exactly for all the evaluation queries. 

\begin{table}[t]
    \centering
    \caption{The elicited linear performance metrics for the ten subjects along with the fraction of times (in \%) the elicited metric's preferences matches with the subject's preferences over the fifteen evaluation queries.}
    \begin{tabular}{|c|c|c|}
    \hline
    \textbf{Subjects} & \textbf{Linear Performance Metric} & $\Mcal$ \\
    \hline
         S1 & 0.125 \text{TN} + 0.875 \text{TP}  & 87\\
         S2 & 0.141 \text{TN} + 0.859 \text{TP}  & 100\\
         S3 & 0.125 \text{TN} + 0.875 \text{TP}  & 93\\
         S4 & 0.141 \text{TN} + 0.859 \text{TP}  & 100\\
         S5 & 0.328 \text{TN} + 0.672 \text{TP}  & 73\\
         S6 & 0.031 \text{TN} + 0.969 \text{TP}  & 87\\
         S7 & 0.031 \text{TN} + 0.969 \text{TP}  & 100\\
         S8 & 0.359 \text{TN} + 0.641 \text{TP}  & 87\\
         S9 & 0.125 \text{TN} + 0.875 \text{TP}  & 93\\
         S10 & 0.141 \text{TN} + 0.859 \text{TP}  & 87\\
    \hline
    \end{tabular}
    \label{pme-tab:metrics}
\end{table}

The absolute numbers for the $\Mcal$ measure look good; however, how good they are is still a missing piece in this study because of the lack of a baseline. In future, we plan to devise ways to develop a baseline for the metric elicitation task and compare to that baseline on the measure $\Mcal$.

\end{document}